\documentclass{article}
\usepackage[margin=1in]{geometry}

\usepackage{microtype}
\usepackage{booktabs} % for professional tables

\usepackage{graphicx}
\usepackage{hyperref}       % hyperlinks
\usepackage{amsmath}
\usepackage{amssymb}
\usepackage{mathtools}
\usepackage{amsthm}
\usepackage[textsize=tiny]{todonotes}

\usepackage{amsmath,bm}
\usepackage{amsfonts}
\usepackage{amssymb}
\usepackage{amsthm}
\usepackage{nicefrac}

\usepackage{xcolor}
\usepackage{here}
\usepackage{natbib}
\usepackage{listings}
\usepackage{authblk}
\usepackage{subfig}
\usepackage{url}            % simple URL typesetting
\usepackage{arydshln}

\usepackage{accents}
\usepackage{wrapfig}

\usepackage{algpseudocode}
\usepackage{algorithm}
\algdef{SE}[PROCEDURE]{Procedure}{EndProcedure}%
   [2]{\algorithmicprocedure\ \textproc{#1}\ifthenelse{\equal{#2}{}}{}{(#2)}}%
   {\algorithmicend\ \algorithmicprocedure}%
\algdef{SE}[FUNCTION]{Function}{EndFunction}%
   [2]{\algorithmicfunction\ \textproc{#1}\ifthenelse{\equal{#2}{}}{}{(#2)}}%
   {\algorithmicend\ \algorithmicfunction}%
\def\*#1{\boldsymbol{#1}}

\newtheorem{theo}{Theorem}[section]

\newtheorem{lem}{Lemma}[section]

\newtheorem{rema}{Remarks}[section]

\newcommand{\tw}[0]{\textwidth}
\newcommand{\igr}[2]{\includegraphics[clip,width=#1\tw]{#2}}
\newcommand{\cigr}[2]{\centering \includegraphics[clip,width=#1\tw]{#2}}

\DeclareMathOperator*{\argmax}{arg\,max}
\newcommand{\eq}[1]{(\ref{#1})}

\newcommand{\lw}[1]{\smash{\lower2.ex\hbox{#1}}}

\newcommand{\RR}{\mathbb{R}}

\newcommand{\EE}{\mathbb{E}}
\newcommand{\VV}{\mathbb{V}}
\newcommand{\cA}{{\cal A}}
\newcommand{\cB}{{\cal B}}

\newcommand{\cD}{{\cal D}}

\newcommand{\cF}{{\cal F}}
\newcommand{\cG}{{\cal G}}
\newcommand{\cH}{{\cal H}}

\newcommand{\cJ}{{\cal J}}

\newcommand{\cN}{{\cal N}}

\newcommand{\cP}{{\cal P}}

\newcommand{\cX}{{\cal X}}

\title{Sequential- and Parallel- Constrained Max-value Entropy Search \\ via Information Lower Bound}
\author[1,2]{Shion~Takeno}
\author[1]{Tomoyuki~Tamura}
\author[3,4]{Kazuki~Shitara}
\author[1]{Masayuki~Karasuyama}

\affil[1]{Nagoya Institute of Technology}
\affil[2]{RIKEN Center for Advanced Intelligence Project}
\affil[3]{Osaka University}
\affil[4]{Japan Fine Ceramics Center}

\affil[ ]{\textit{takeno.s.mllab.nit@gmail.com, tomoyuki.tamura@nitech.ac.jp,}}
\affil[ ]{\textit{shitara@jwri.osaka-u.ac.jp, karasuyama@nitech.ac.jp}}
\date{}

\begin{document}
\maketitle

%%%%%%%%%%%%%%%%%%%%%%%%%%%%%%%%%%%%%%%%%%%%%%%%%%%%%%%%%%%%%%%%%%%%%%%%%%%%%%%%%%%%%%%%%%%%%%
% abstract
%%%%%%%%%%%%%%%%%%%%%%%%%%%%%%%%%%%%%%%%%%%%%%%%%%%%%%%%%%%%%%%\begin{abstract}

\begin{abstract}
    Max-value entropy search (MES) is one of the state-of-the-art approaches in Bayesian optimization (BO).
    In this paper, we propose a novel variant of MES for constrained problems, called Constrained MES via Information lower BOund (CMES-IBO), that is based on a Monte Carlo (MC) estimator of a lower bound of a mutual information (MI).
    Unlike existing studies, our MI is defined so that uncertainty with respect to feasibility can be incorporated.
    We derive a lower bound of the MI that guarantees non-negativity, while a constrained counterpart of conventional MES can be negative.
    We further provide theoretical analysis that assures the low-variability of our estimator
    which has never been investigated for any existing information-theoretic BO.
    Moreover, using the conditional MI, we extend CMES-IBO to the parallel setting while maintaining the desirable properties.
    We demonstrate the effectiveness of CMES-IBO by several benchmark functions and real-world problems.
%
% Max-value entropy search (MES) is one of the state-of-the-art approaches in Bayesian optimization (BO). In this paper, we propose a novel variant of MES for constrained problems, called Constrained MES via Information lower BOund (CMES-IBO), that is based on a Monte Carlo (MC) estimator of a lower bound of a mutual information (MI). Unlike existing studies, our MI is defined so that uncertainty with respect to feasibility can be incorporated. We derive a lower bound of the MI that guarantees non-negativity, while a constrained counterpart of conventional MES can be negative. We further provide theoretical analysis that assures the low-variability of our estimator which has never been investigated for any existing information-theoretic BO. Moreover, using the conditional MI, we extend CMES-IBO to the parallel setting while maintaining the desirable properties. We demonstrate the effectiveness of CMES-IBO by several benchmark functions and real-world problems.
\end{abstract}

%%%%%%%%%%%%%%%%%%%%%%%%%%%%%%%%%%%%%%%%%%%%%%%%%%%%%%%%%%%%%%%%%%%%%%%%%%%%%%%%%%%%%%%%%%%%%%
\section{Introduction}
\label{sec:intro}
%%%%%%%%%%%%%%%%%%%%%%%%%%%%%%%%%%%%%%%%%%%%%%%%%%%%%%%%%%%%%%%%%%%%%%%%%%%%%%%%%%%%%%%%%%%%%%

%%%%%%%%%%%%%%%%%%%%%%%%%%%%%%%%%%%%%%%%%%%%%%%%%%%%%%%%%%%%%%%%%%%%%%%%%%%%%%%%%%%%%%%%%%%%%%
Bayesian optimization (BO) has been widely studied as an effective framework for expensive black-box optimization problems.
%
% On the other hand, various real-world problems have additional unknown \emph{constraints}.
On the other hand, additional unknown \emph{constraints} often exist in real-world problems of a variety of fields such as scientific experiments, industrial product designs, and automatic machine learning (AutoML).
For example, materials discovery can be seen as an optimization of a physical property of materials, such as conductivity, under constraints derived from other physical properties, such as stability.
In the case of AutoML, for example, constraints are useful to guarantee the classification accuracy of minority classes for class-imbalanced datasets.
However, constrained black-box optimization is often quite difficult because, in many practical problems, a measurement comes at a high cost, and further, functional forms of both the objective and the constraints are unknown.
To incorporate unknown constraint functions, \emph{constrained BO} (CBO) has also been studied \citep{Gardner2014-Bayesian,Gelbart2014-Bayesian,Schonlau1998-Global,Snoek2013-Bayesian}.
CBO tries to achieve the sample-efficient optimization by repeatedly observing objective and constraint functions that are estimated to be beneficial to identify the optimal solution.

Max-value entropy search (MES) \citep{Wang2017-Max} is one of the state-of-the-art approaches in BO.
The basic idea is to maximize the mutual information (MI) between a querying point and the optimal objective value.
For constrained problems,
% \cite{Perrone2019-Constrained} have considered a straightforward extension of MES.
% to CBO called constrained MES (cMES)
\citet{Perrone2019-Constrained} have proposed an extension of MES, but it is restricted to only one constraint (note that although their main focus is on a setting in which only a binary indicator of feasibility is observed for constraints, they also show the `real-valued feedback' case in the appendix).
% \blue{(takeno:固有名詞でないから小文字?)}
%
% \blue{
Although it is possible to consider a multiple constraint extension of this approach, which we call constrained MES (CMES), the resulting MI approximation can have a negative value (though the MI should be non-negative) when the number of constraints is more than 5 as we will show in Section~\ref{sec:relation_MES}.
A more comprehensive review of related studies is shown in Section~\ref{sec:related_work}.

In this paper, we propose a novel information-theoretic CBO method called {\it Constrained Max-value Entropy Search via Information lower BOund} (CMES-IBO), which is based on a Monte Carlo (MC) estimator of a lower bound of an MI.
We first define the random `max-value' of a constrained problem so that uncertainty with respect to feasibility (whether the problem has a feasible region or not) can be incorporated.
%
% This issue has not been considered by existing information-theoretic CBO studies, though for constrained problems in general, to identify the feasibility of the problem is a key issue (called the feasibility problem \citep{Chinneck2007-feasibility}).
Uncertainty of feasibility has not been considered by existing information-theoretic CBO studies, though for constrained problems in general, to identify the feasibility of the problem is a key issue known as the feasibility problem \citep{Chinneck2007-feasibility}.
Unlike CMES, our acquisition function guarantees non-negativity.
% Then, we derive an acquisition function based on a lower bound of the MI that can guarantee non-negativity.
%
% Unlike CMES, the MI approximation of CMES-IBO guarantees non-negativity by which
% We empirically observe that CMES-IBO can provide reasonable acquisition function values even when CMES provides negative values.
We theoretically and empirically verify that CMES-IBO can provide reasonable acquisition function values even when CMES provides negative values.
% Because of its \blue{non-negativity}, CMES-IBO improves the \blue{worst-case error} of the MI estimation compared with a constrained counterpart of conventional MES, i.e., CMES.
%
Further, we provide theoretical analysis that assures the low-variability of our estimator, such as a concentration bound achieving an exponentially fast convergence to the true lower bound in the number of MC samples, which has never been investigated for any existing information-theoretic BO.
%
% Another notable difference of our study from other information-theoretic CBO is that our MI can be defined even when there is a possibility that the feasible region becomes empty.
%
% This occurs particularly when we still do not have any feasible solution, though it has not been considered by existing information-theoretic CBO.
%
% \red{This enables us to incorporate uncertainty about feasibility (whether the problem has a feasible region or not) into the decision-making.}
%
Moreover, using the conditional MI, we extend CMES-IBO to the parallel setting while maintaining the desirable properties.

Our main contributions are summarized as follows:
\vspace{-0.5em}
\begin{enumerate}
 \setlength{\parskip}{0em}
 \setlength{\itemsep}{0em}
 \item We develop an MI lower bound based CBO called CMES-IBO.
       Our MI is well-defined even when a feasible region can be empty.
       The resulting MI approximation is quite simple and guarantees non-negativity.
       %
%  \item We consider differences of CMES-IBO from the more direct MES extension, called CMES, that has the possible negativity issue, and we also provide analysis on the estimation variability of our estimator.
       % We derive the desirable properties of CMES-IBO (a)-(d) listed in the previous paragraph, and further, relations with the na{\"i}ve extension of MES is also revealed.
 \item We reveal advantageous properties of CMES-IBO compared with the direct MES extension (CMES), such as the smaller worst-case error of the MI estimation.
 We also derive an estimation variability of CMES-IBO that suggests robustness with respect to the number of MC samples.
 \item We further develop a parallel extension of CMES-IBO, in which multiple queries can be issued simultaneously.
       The extension is based on a combination of the conditional MI and the greedy selection.
\end{enumerate}
\vspace{-0.5em}
We demonstrate the effectiveness of CMES-IBO by benchmark and real-world functions.

%%%%%%%%%%%%%%%%%%%%%%%%%%%%%%%%%%%%%%%%%%%%%%%%%%%%%%%%%%%%%%%%%%%%%%%%%%%%%%%%%%%%%%%%%%%%%
\section{Preliminary}
\label{sec:preliminary}
%%%%%%%%%%%%%%%%%%%%%%%%%%%%%%%%%%%%%%%%%%%%%%%%%%%%%%%%%%%%%%%%%%%%%%%%%%%%%%%%%%%%%%%%%%%%%%

We are interested in the maximum value of the objective function $f: \cX \mapsto \RR$ under $C$ constraint functions $g_c: \cX \mapsto \RR$ for $c=1,\dots, C$, where $\cX \subset \RR^{d}$ is an input space.
Let
$\cX_{\rm feasible} \coloneqq \{ \*x \mid g_c(\*x) \geq z_c, c = 1, \ldots, C \}$
be the feasible region, where $z_c \in \RR$ is a given constant.
The optimal solution of this constrained optimization problem is written as
$\*x_* \coloneqq \argmax_{\*x \in \cX_{\rm feasible}} f(\*x)$.
% \begin{align*}
%  \*x_* \coloneqq \argmax_{\*x \in \cX_{\rm feasible}} f(\*x).  % \*x_* \coloneqq \argmax_{\*x \in \cX} f(\*x),\ {\rm s.t. }\  g_c(\*x) \geq z_c,\  c=1, \dots, C, % \label{eq:goal}
% \end{align*}
% where $z_c \in \RR$ is an arbitrary constant.
%
We consider the setting that the functions $f$ and $g_1, \dots, g_C$ are unknown but can be evaluated simultaneously for any given $\*x \in \cX$ with high observation cost.
We assume that $f$ and $g_1, \dots, g_C$ follow independent \emph{Gaussian processes} (GPs).
%
% \red{We assume that each follows an independent \emph{Gaussian process} (GP).}
%
Suppose that we already queried $n$ points $\{ (f(\*x_i), g_1(\*x_i), \ldots, g_C(\*x_i)) \}_{i \in 1, \ldots, n}$ for each of $f(\*x_i)$ and $g_c(\*x_i)$, actual observations can be contaminated with additive noises generated by a Gaussian distribution.
We write predictive distributions of $f(\*x)$ and $g_c(\*x)$ as
% $\cN\bigl(\mu^{(f)}_{n}(\*x), {\sigma^{(f)}_{n}}^2(\*x)\bigl)$
$\cN\bigl(\mu^{(f)}(\*x), {\sigma^{(f)}}^2(\*x)\bigl)$
and
% $\cN\bigl(\mu^{(g_c)}_{n}(\*x), {\sigma^{(g_c)}_{n}}^2(\*x)\bigl)$,
$\cN\bigl(\mu^{(g_c)}(\*x), {\sigma^{(g_c)}}^2(\*x)\bigl)$,
respectively.
% %
% \red{Note that
% ${\sigma^{(f)}_{n}}^2(\*x) = {\sigma^{(g_c)}_{n}}^2(\*x)$
% always holds and we write simply $\sigma_{n}(\*x)$ as a common predictive variance.}
%
% Since we use standard GP models, detailed definitions of these posteriors are omitted here (see Appendix~\ref{sec:definition_GPs}).
%
% Since we use standard GP models, detailed definitions of these posteriors are shown in Appendix~\ref{sec:definition_GPs}.
More detailed definitions of these posteriors are shown in Appendix~\ref{sec:definition_GPs}.

%%%%%%%%%%%%%%%%%%%%%%%%%%%%%%%%%%%%%%%%%%%%%%%%%%%%%%%%%%%%%%%%%%%%%%%%%%%%%%%%%%%%%%%%%%%%%%
\section{Constrained Max-value Entropy Search via Information Lower Bound}
\label{sec:CMES-IBO}
%%%%%%%%%%%%%%%%%%%%%%%%%%%%%%%%%%%%%%%%%%%%%%%%%%%%%%%%%%%%%%%%%%%%%%%%%%%%%%%%%%%%%%%%%%%%%%

\begin{figure}[!t]
    % \centering
    \igr{.58}{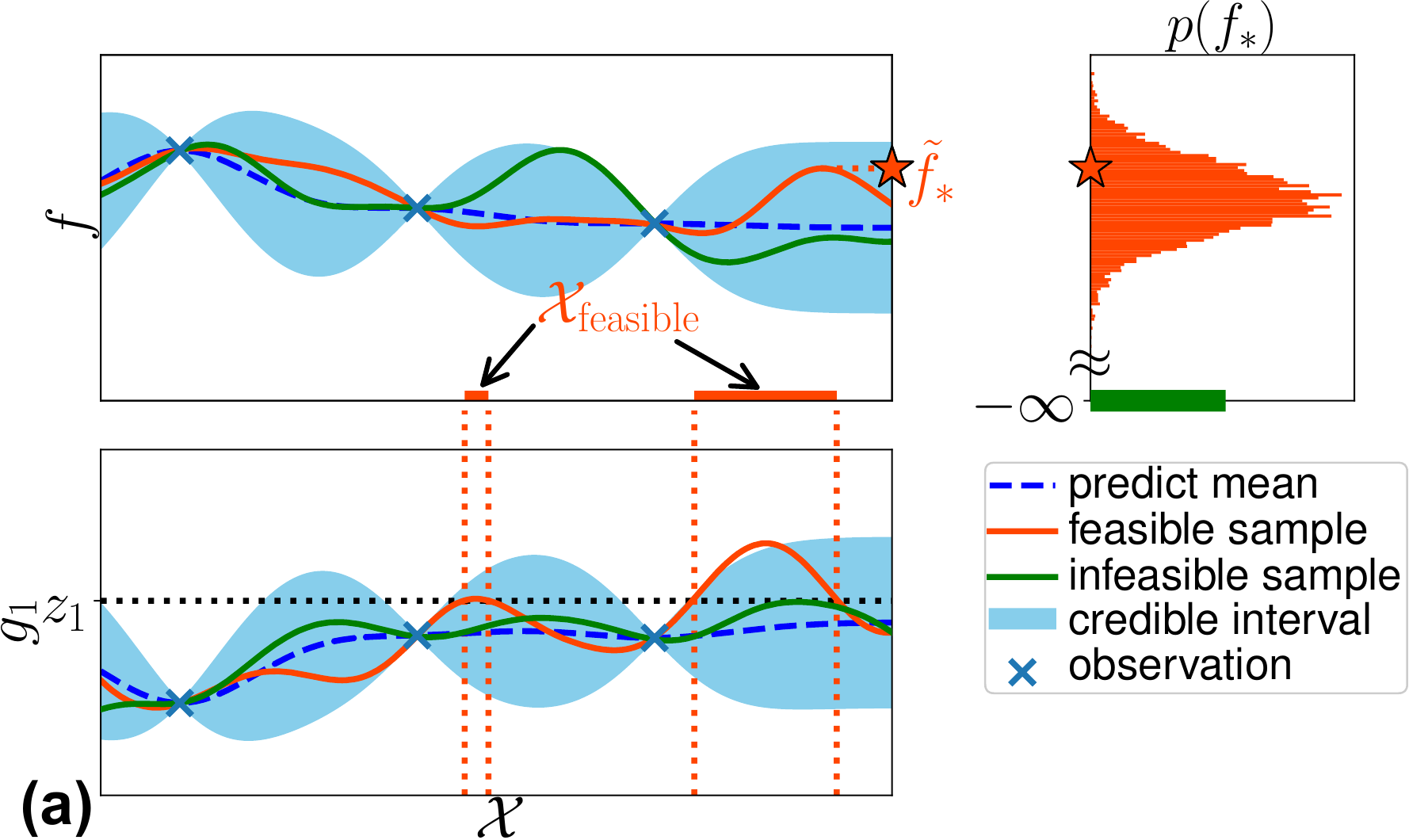}
    % \subfloat[][]{\igr{.47}{./CMES-explanation_a.pdf}}
    % \centering
    \igr{.38}{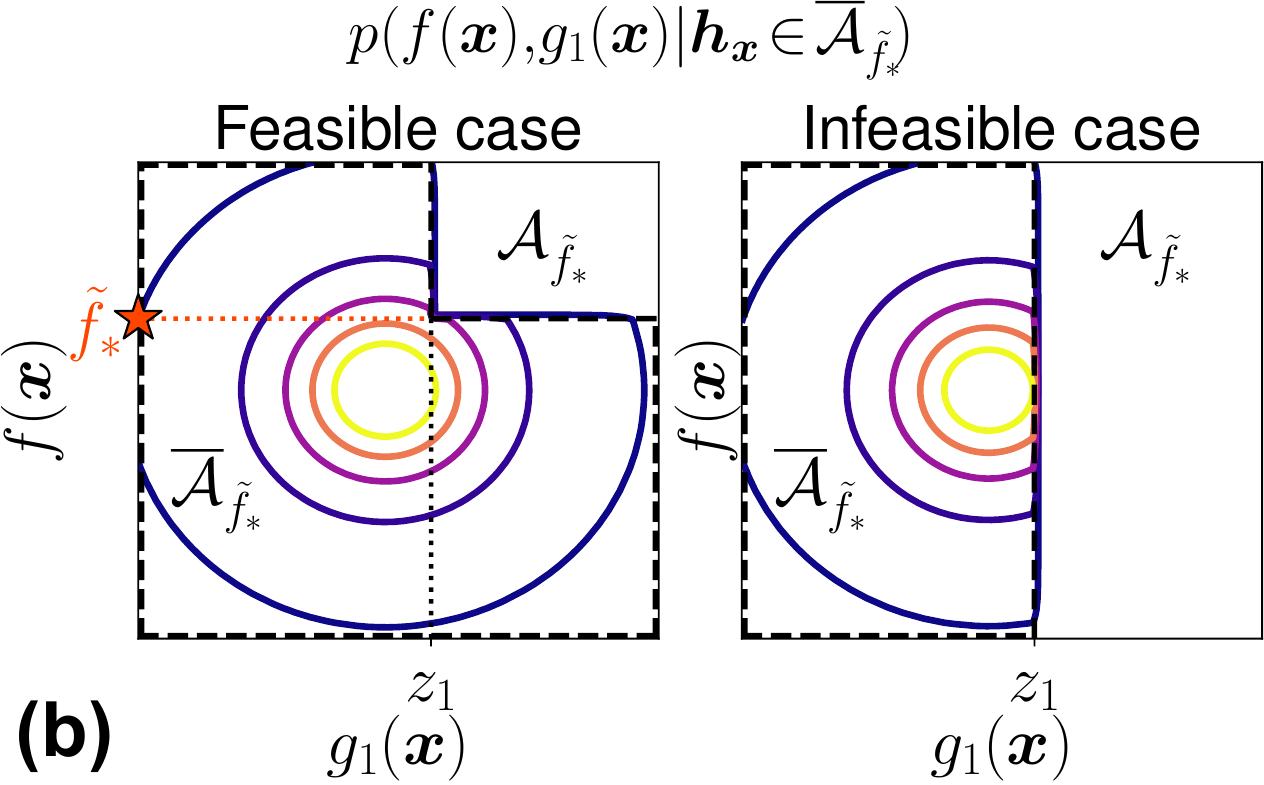}
    % \subfloat[][]{\igr{.35}{./CMES-explanation_b.pdf}}
   %  \subfloat[][]{
   %  \igr{.2}{./CMES-explanation_b-ver2-1.pdf}
   %  \igr{.2}{./CMES-explanation_b-ver2-2.pdf}
   %  }
    \caption{
    Generative process of the optimal value $f_*$ and truncation of the predictive distribution.
    (a) In the left two plots, two sets of sample paths (red and green) are generated from GPs.
    The red sample path of $g_1$ has the non-empty feasible region $\cX_{\rm feasible}$ while the green sample path has an empty feasible region.
    The red star $\tilde{f}_*$ represents the optimal value of the red sample path, and for the green sample path, the optimal value is defined as $-\infty$ as shown in \eqref{eq:f_star}.
    The distribution of the optimal value is plotted in the upper right figure.
    % \blue{
    % The optimal values determined by \eqref{eq:f_star} are is shown as stars, which can be seen as samples from $p(f_*)$.
    % }
    %
    % The distribution of the optimal value determined by \eqref{eq:f_star} is plotted in the upper right figure, and the optimal value of the red sample path is shown as a red star.
    %
    (b)
    % The horizontal and vertical axes express $g_1(\*x)$ and $f(\*x)$, respectively.
    %
    The two contour plots represent the density function {$p(\*h_{\*x} | \*h_{\*x} \in \overline{\cA}_{\tilde{f}_*}) = p(f(\*x), g_1(\*x) | \*h_{\*x} \in \overline{\cA}_{\tilde{f}_*})$} for the feasible and infeasible cases, respectively.
    For a given $\tilde{f}_*$, the density should be $0$ in $\cA_{\tilde{f}_*}$, because $f(\*x)$ cannot be larger than the given $\tilde{f}_*$ for $\*x$ such that $g_1(\*x) \geq z_1$.
    }
    \label{fig:truncation}
\end{figure}

\subsection{Max-value of Constrained Problem}
\label{sec:max-value}
First, we consider the definition of a random variable $f_*$, which is the max-value of a constraint optimization problem.
The straightforward definition is
$\max_{x \in \cX_{\rm feasible}} f(\*x)$,
employed in a prior work on the MES-based CBO \citep{Perrone2019-Constrained}.
%
% However, this approach cannot provide any value when $\cX_{\rm feasible}$ is empty, and thus, $\max_{x \in \cX_{\rm feasible}} f(\*x)$ is not well-defined as a proper random variable when $\Pr(\cX_{\rm feasible} = \emptyset) > 0$.
However, this definition is not well-defined as a proper random variable when $\cX_{\rm feasible}$ can be empty, i.e., when $\Pr(\cX_{\rm feasible} = \emptyset) > 0$.
Note that $\cX_{\rm feasible}$ is randomly determined through GPs $g_1, \dots, g_C$, and thus, $\Pr(\cX_{\rm feasible} = \emptyset) > 0$ can occur even when the underlying unknown true constraint functions have a non-empty feasible region.
Thus, we define $f_*$ as follows:
\begin{align}
    f_* \coloneqq
    \begin{cases}
        \max_{x \in \cX_{\rm feasible}} f(\*x), & \text{if } \cX_{\rm feasible} \neq \emptyset, \\
        % \min_{\*x \in \cX} f(\*x) - \zeta, & \text{otherwise},
        -\infty, & \text{otherwise}.
    \end{cases}
    \label{eq:f_star}
\end{align}
% \blue{where $\zeta \geq 0$ is an arbitrary pre-specified constant.}
%
Regarding the optimal value as $- \infty$ when no feasible solution exists is a common convention in classical optimization literature \citep{Boyd2004-convex}.
%
% \blue{When the feasible region is empty, $f_*$ is set as a value that must be worse than any other possible $f(\*x)$ in the domain $\cX$.}
%
% The schematic illustration of $f_*$ is shown in Figure~\ref{fig:truncation}~(a).
The schematic illustration of $f_*$ is shown in Fig.~\ref{fig:truncation}~(a).
%
% \blue{Note that $p(f_*)$ can be seen as a mixture distribution of the discrete and continuous variables (red and green histograms in the figure).}
%
The figure indicates that our definition of $p(f_*)$ can incorporate the uncertainty about whether the problem is feasible or not.

\subsection{Acquisition Function of CMES-IBO}
\label{sec:CMES-IBO_acq}

Let
% \begin{align*}
\[
 \*h_{\*x} \coloneqq \bigl(f(\*x), g_1(\*x), \dots, g_C(\*x) \bigl)^\top \in \RR^{C+1},
\]
% \end{align*}
% \red{$\*h_{\*x} \coloneqq \bigl(f(\*x), g_1(\*x), \dots, g_C(\*x) \bigl)^\top \in \RR^{C+1}$}
%
%
be a vector concatenating the objective and all the constraint function values.
Suppose that we already have the dataset $\cD_{t-1}$, which contains observations for the past $t-1$ queries.
We consider selecting a next query $\*x_{t}$ by maximizing the MI
\footnote{
    Note that although $p(f_*)$ can be seen as a mixture distribution in which one of the components (the green distribution in Fig.~\ref{fig:truncation}) consists of a single constant value ($-\infty$), MI still can be defined in this case. Details are shown in Appendix~\ref{sec:IBO}.
}
% Since $p(f_*)$ is a mixture distribution of degenerate and continuous distributions, Since $p(f_*)$ is a mixture distribution of degenerate and continuous distributions, we use the definition of MI in \cite{Nair2007-On,Beknazaryan2019-On} (details are shown in Appendix~\ref{sec:IBO}).
between $\*h_{\*x}$ and $f_*$ given $\cD_{t-1}$:
\begin{align}
 % I(\*h_{\*x} ; f_* | \cD_{t-1}). \label{eq:MI}
 {\rm{MI}}(\*h_{\*x} ; f_* | \cD_{t-1}). \label{eq:MI}
\end{align}
Hereafter, we omit the conditioning by $\cD_{t-1}$ in MI, entropy, density, and probability when it is obvious from the context.
For example,
$p(\*h_{\*x} \mid f_*)$ indicates
$p(\*h_{\*x} \mid \cD_{t-1}, f_*)$.

Since directly evaluating the MI \eqref{eq:MI} is computationally intractable, we consider a lower bound derived as follows:
\begin{align}
 % I(\*h_{\*x} ; f_*)
 & {\rm{MI}}(\*h_{\*x} ; f_*)
    % \eqref{eq:MI}
    =
 \EE_{f_*} \biggl[ \EE_{\*h_{\*x} | f_*} \left[  \log \frac{p(\*h_{\*x} | f_*)}{p(\*h_{\*x})} \right] \biggl]
 % \EE_{f_*} \biggl[ \int p(\*h_{\*x} | f_*) \log \frac{p(\*h_{\*x} | f_*)}{p(\*h_{\*x})} {\rm d} \*h_{\*x}  \biggl]
    % \label{eq:MIforLB} \\
    \nonumber \\
 &
 % \! = \! \EE_{f_*} \! \biggl[ \int \! p(\*h_{\*x} | f_*) \log \frac{q(\*h_{\*x} | f_*)}{p(\*h_{\*x})} {\rm d} \*h_{\*x}
 \! = \! \EE_{f_*} \! \biggl[ \EE_{\*h_{\*x} | f_*} \! \left[ \log \frac{q(\*h_{\*x} | f_*)}{p(\*h_{\*x})} \! \right]
 % \nonumber \\ & \qquad \quad
   \! + \!  D_{\rm KL} \bigl( p(\*h_{\*x} | f_*) || q(\*h_{\*x} | f_*) \bigl) \! \biggl] \nonumber \\
%  & \! = \! \EE_{f_*} \! \biggl[ \int \! p(\*h_{\*x} | f_*) \log \biggl( \frac{q(\*h_{\*x} | f_*)}{p(\*h_{\*x})} \frac{p(\*h_{\*x} | f_*)}{q(\*h_{\*x} | f_*)} \biggr) {\rm d} \*h_{\*x} \nonumber \\
%  \nonumber \\ & \qquad \qquad
%    \! + \!  D_{\rm KL} \bigl( p(\*h_{\*x} | f_*) || q(\*h_{\*x} | f_*) \bigl) \! \biggl] \nonumber \\
 & \! \geq \! \EE_{f_*} \! \biggl[ \int p(\*h_{\*x} | f_*) \log \frac{q(\*h_{\*x} | f_*)}{p(\*h_{\*x})} {\rm d} \*h_{\*x}  \biggl], \label{eq:LB_first}
    % &= - \EE_{f_*} [ \log ( 1 - Z ) ], \label{eq:IBO}
\end{align}
where $D_{\rm KL}(\cdot)$ is Kullback-Leibler (KL) divergence and $q(\*h_{\*x} | f_*)$ is an arbitrary probability density function (PDF) that has same support as $p(\*h_{\*x} | f_*)$.
%
% See Appendix~\ref{sec:IBO} for the derivation.
See Appendix~\ref{sec:IBO} for the derivation.
% \blue{(note that our derivation takes into account that $f_*$ is defined through the mixture distribution)}
%
The difference between the lower bound and the MI is given as
$\EE_{f_*} \bigl[D_{\rm KL}\bigl( p(\*h_{\*x} | f_*) || q(\*h_{\*x} | f_*) \bigl) \bigl]$,
and thus, the equality holds if
$p(\*h_{\*x} | f_*)$ is equal to $q(\*h_{\*x} | f_*)$ for $\forall f_*$.
%
% where $q(\*h_{\*x} | f_*)$ is an arbitrary probability density function (PDF) that has same support as $p(\*h_{\*x} | f_*)$.
% %
% See Appendix~\ref{sec:IBO} for the derivation.
% %
% The difference between the lower bound and the MI is given as
% $\EE_{f_*} \bigl[D_{\rm KL}\bigl( p(\*h_{\*x} | f_*) || q(\*h_{\*x} | f_*) \bigl) \bigl]$,
% where $D_{\rm KL}(\cdot)$ is Kullback-Leibler (KL) divergence.
% %
% Thus, the equality holds if
% $p(\*h_{\*x} | f_*)$ is equal to $q(\*h_{\*x} | f_*)$ for $\forall f_*$.

% To use the lower bound \eqref{eq:LB_first}, we have to specify the distribution $q(\*h_{\*x} | f_*)$.
To use \eqref{eq:LB_first}, we have to specify the distribution $q(\*h_{\*x} | f_*)$.
For this type of KL-based lower bounds, the variational approximation is a well-known approach by which the lower bound is maximized with respect to $q(\*h_{\*x} | f_*)$.
However, the expectation and the integral with respect to $f_*$ and $\*h_{\*x}$ in \eqref{eq:LB_first} make this optimization computationally too complicated.
Instead, we consider setting $q(\*h_{\*x} | f_*)$ as a specific distribution that can be easily computed.
%
% \blue{(k1009:ちょっと誤魔化してる状態. 計算が大変だからやらないという最初の記述がよくわからなかった．$\EE_{f_*}$だってcomputationally intractableだけどやってる)}

The optimal selection of $q(\*h_{\*x} | f_*)$ is $p(\*h_{\*x} | f_*)$, i.e., predictive distributions of GPs given the optimal value $f_*$ which we call a GP$_{\mid f_*}$.
For all existing MES-based approaches, the GP$_{\mid f_*}$ is required in a slightly different context
% \red{form} \blue{reason}
(see Section~\ref{sec:relation_MES} for the case of the existing MES for constrained problems).
Since the GP$_{\mid f_*}$ is difficult to evaluate, it is approximated by a distribution defined by truncating the original GP predictive distribution at $f_*$ in almost all the existing methods \citep[e.g.,][]{Wang2017-Max,Takeno2020-Multifidelity,Perrone2019-Constrained,Suzuki2020-multi}.
For example, in the case of the original MES, $p(f(\*x) \mid f_*)$ is replaced with
$p(f(\*x) \mid f(\*x) \leq f_*)$,
which is the normal distribution truncated at $f_*$.
This replacement makes the resulting acquisition function simple and practical performance of MES-based approaches with this replacement has been repeatedly shown.
Therefore, we follow this truncation-based approach.

In the case of constrained problems, when the optimal value $f_*$ is fixed, $f(\*x) > f_*$ does not hold if $g_c(\*x) \geq z_c$ for $\forall c$ (i.e., $\*x$ is feasible).
This means that $f(\*x)$ and $g_c(\*x)$ are jointly truncated so that their densities become $0$ in
$\cA_{f_*} \coloneqq (f_*, \infty) \times (z_1, \infty) \times \dots \times (z_C, \infty) \subset \RR^{C+1}$.
%
% In other words, $\*h_{\*x}$ should be in $\overline{\cA}_{f_*} \subset \RR^{C+1}$, which is a complementary set of $\cA_{f_*}$, as illustrated in Figure~\ref{fig:truncation}~(b).
In other words, $\*h_{\*x}$ should be in $\overline{\cA}_{f_*} \subset \RR^{C+1}$, which is a complementary set of $\cA_{f_*}$, as illustrated in Fig.~\ref{fig:truncation}~(b).
Thus, we set $q(\*h_{\*x} | f_*)$ as the following {\it truncated multivariate normal} (TMN) distribution:
\begin{align}
 q(\*h_{\*x} | f_*)
 &=
 p(\*h_{\*x} \mid \*h_{\*x} \in \overline{\cA}_{f_*}) \nonumber \\
 &=
 \begin{cases}
  % \frac{1}{1 - Z} p(\*h_{\*x}) & \text{ for } \*h_{\*x} \in \overline{\cA}_{f_*}, \\
  p(\*h_{\*x}) / \overline{Z}_{\*x}(f_*) & \text{ for } \*h_{\*x} \in \overline{\cA}_{f_*}, \\
%   p(\*h_{\*x}) / \Pr(\*h_{\*x} \in \overline{\cA}_{f_*}) & \text{ for } \*h_{\*x} \in \overline{\cA}_{f_*}, \\
  0 & \text{ for }  \*h_{\*x} \in \cA_{f_*},
 \end{cases}
 \label{eq:TMN}
\end{align}
where
$\overline{Z}_{\*x}(f_*) = \Pr(\*h_{\*x} \in \overline{\cA}_{f_*})$
is the normalization constant.
% $\overline{Z}_{\*x}(f_*) = \Pr(\*h_{\*x} \in \overline{\cA}_{f_*})$.

% By substituting TMN \eqref{eq:TMN} and $\overline{Z}_{\*x}(f_*) = 1 - Z_{\*x}(f_*)$ into \eqref{eq:LB_first}, we obtain a lower bound $L(\*x)$ as
By substituting TMN \eqref{eq:TMN} into \eqref{eq:LB_first}, we obtain a lower bound $L(\*x)$ as
% in which the integral inside the expectation can be canceled out as follows:
%
\begin{align}
 L(\*x)
 & \!
 \coloneqq  % =
 % \EE_{f_*} \biggl[ \int_{\overline{\cA}_{f_*}} \! \! p(\*h_{\*x} | f_*) \log \frac{p(\*h_{\*x}) /  (1 - Z_{\*x}(f_*))}{p(\*h_{\*x})} {\rm d} \*h_{\*x} \! \biggr] \nonumber \\
 \EE_{f_*}
 \biggl[
 \int_{\overline{\cA}_{f_*}} \! \! p(\*h_{\*x} | f_*) \log \frac{p(\*h_{\*x})
 /
 \overline{Z}_{\*x}(f_*) % (1 - Z_{\*x}(f_*))
 }
 {p(\*h_{\*x})} {\rm d} \*h_{\*x}
 \! \biggr] \nonumber \\
 &=
 % \EE_{f_*} \bigl[ - \log (1 - Z_{\*x}(f_*)) \bigr].
 \EE_{f_*} \bigl[ - \log \overline{Z}_{\*x}(f_*) \bigr].
 \label{eq:IBO}
\end{align}
See Appendix~\ref{sec:IBO} for the detailed derivation.
In \eqref{eq:IBO}, $C+1$ dimensional integral with respect to $\*h_{\*x}$ is canceled out, which is one of important benefits of our selection of $q(\*h_{\*x} | f_*)$.
%
% By applying the Monte Carlo approximation to
By applying the MC approximation to
$\EE_{f_*}$
in \eqref{eq:IBO}, we obtain the CMES-IBO acquisition function
\begin{align}
 \alpha^{\rm IBO}(\*x) = - \frac{1}{K} \sum_{\tilde{f}_* \in \cF_*} \log  \overline{Z}_{\*x}(\tilde{f}_*), \label{eq:CMES-IBO}
\end{align}
where $\cF_*$ is a set of sampled $f_*$ from the current GPs, $\tilde{f}_* \in \cF_*$ represents each one of sampled values, and $K = |\cF_*|$.

Let $Z_{\*x}(\tilde{f}_*) \coloneqq \Pr(\*h_{\*x} \in \cA_{\tilde{f}_*}) = 1 - \overline{Z}_{\*x}(\tilde{f}_*)$.
For a given $\tilde{f}_* \in \cF_*$, we can compute $\overline{Z}_{\*x}(\tilde{f}_*)$ easily by using
% For any given $\tilde{f}_* \in \cF_*$, $\overline{Z}_{\*x}(\tilde{f}_*)$ can be computed easily by using
% $Z_{\*x}(\tilde{f}_*) \coloneqq \Pr(\*h_{\*x} \in \cA_{\tilde{f}_*}) = 1 - \overline{Z}_{\*x}(\tilde{f}_*)$,
% through which the following decomposition can be obtained from the independence assumption of $f$ and $g_c$ for $\forall c$:
% Since $\cA_{\tilde{f}_*}$ is a hyperrectangle, based on the independence assumption of $f$ and $g_c$ for $\forall c$, we can decompose $Z_{\*x}(\tilde{f}_*)$ as follows:
\begin{align*}
 & \overline{Z}_{\*x}(\tilde{f}_*) = 1 - Z_{\*x}(\tilde{f}_*)
 \\
 & \qquad = 1 - \Pr(f(\*x) \geq \tilde{f}_*) \mathord{\textstyle\prod}_{c = 1}^C \Pr(g(\*x) \geq z_c) \nonumber \\
  & \qquad = 1 - \bigl(1 - \Phi(\gamma^{(f)}_{\*x}(\tilde{f}_*))\bigl) \mathord{\textstyle\prod}_{c = 1}^C \bigl(1 - \Phi(\gamma^{(g_c)}_{\*x}(z_c))\bigl),
%  & Z_{\*x}(\tilde{f}_*) %& \coloneqq \Pr(\*h_{\*x} \in \cA_{\tilde{f}_*})
%  = \Pr(f(\*x) \geq \tilde{f}_*) \mathord{\textstyle\prod}_{c = 1}^C \Pr(g(\*x) \geq z_c) \nonumber \\
%  & \quad = \bigl(1 - \Phi(\gamma^{(f)}_{\*x}(\tilde{f}_*))\bigl) \mathord{\textstyle\prod}_{c = 1}^C \bigl(1 - \Phi(\gamma^{(g_c)}_{\*x}(z_c))\bigl),
 \label{eq:Z}
\end{align*}
where
$\gamma^{(f)}_{\*x}(\tilde{f}_*) \coloneqq {(\tilde{f}_* - \mu^{(f)}(\*x))}/{\sigma^{(f)}(\*x)}$,
$\gamma^{(g_c)}_{\*x}(z_c) \coloneqq {(z_c - \mu^{(g_c)}(\*x))}/{\sigma^{(g_c)}(\*x)}$,
% \begin{align*}
%  \gamma^{(f)}_{\*x}(\tilde{f}_*) &\coloneqq \frac{\tilde{f}_* - \mu^{(f)}(\*x)}{\sigma^{(f)}(\*x)},
%  \gamma^{(g_c)}_{\*x}(z_c) \coloneqq \frac{z_c - \mu^{(g_c)}(\*x)}{\sigma^{(g_c)}(\*x)},
% \end{align*}
and $\Phi$ is the cumulative distribution function (CDF) of the standard normal distribution.
This decomposition is possible because $f$ and $g_c$ for $\forall c$ are assumed to be independent, and $\cA_{\tilde{f}_*}$ is a hyperrectangle.

% Consequently, our acquisition function is reduced to a simple functional form.
Consequently, the functional form of our acquisition function \eq{eq:CMES-IBO} becomes quite simple.
The algorithm and computational complexity for our CMES-IBO is shown in Appendix~\ref{sec:computation}.
%
% Although we mainly focus on the setting that GPs are independent of each other, CMES-IBO can also apply to the correlated setting easily by computing $Z_{\*x}(f_*) = \Pr(\*h_{\*x} \in \cA_{f_*})$ as the CDF of the multivariate (correlated) normal distribution.
Although we assume that GPs are independent of each other, CMES-IBO can also apply to the correlated setting easily by computing $Z_{\*x}(\tilde{f}_*) = \Pr(\*h_{\*x} \in \cA_{\tilde{f}_*})$ as the CDF of the multivariate (correlated) normal distribution.

% --------------------------------------------------
\subsection{Sampling $f_*$}
\label{ssec:sampling-fstar}

% \blue{When $\cX$ is moderate size of a finite set, $f_*$ can be obtained by sampling all the candidates points from the joint predictive distributions of $f(\*x)$ for $\*x \in \cX$ and $g_c(\*x)$ for $\*x \in \cX$, respectively.}
%
When $\cX$ is a finite set of a moderate size, $f_*$ can be sampled from the joint predictive distributions of $\*h_{\*x}$ for all candidate points $\*x \in \cX$.
Otherwise (such as a continuous $\cX$), we employ an approach using the random Fourier feature (RFF) \citep{Rahimi2008-Random}.
We first generate approximate sample paths of objective and constraint functions from Bayesian linear regression models using RFF, and then, we solve a (white-box) constrained optimization problem defined by the generated sample paths by using a general solver (see Appendix~\ref{sec:sampling_fstar} for details).
A similar approach has been commonly employed in the entropy-based BO \citep[e.g.,][]{Wang2017-Max}.
%
% Particularly when the dataset $\cD_{t-1}$ does not contain any feasible solution (which typically occurs at the beginning of the optimization), sampled $\cX_{\rm fesasible}$ is often empty, as shown in the green sample path in Figure~\ref{fig:truncation}~(b).
Particularly when the dataset $\cD_{t-1}$ does not contain any feasible solution (which typically occurs at the beginning of the optimization), sampled $\cX_{\rm fesasible}$ is often empty, as shown in the green sample path in Fig.~\ref{fig:truncation}~(b).
%
% Therefore, if the solver cannot find any feasible solution, $f_*$ is set as $-\infty$ as defined in \eqref{eq:f_star}.
Therefore, if the solver cannot find any feasible solution, $\tilde{f}_*$ is set as $-\infty$ as defined in \eqref{eq:f_star}.
% In our definition \eqref{eq:f_star}, $f_*$ can be determined even in this infeasible case.
%

%%%%%%%%%%%%%%%%%%%%%%%%%%%%%%%%%%%%%%%%%%%%%%%%%%%%%%%%%%%%%%%%%%%%%%%
\subsection{Parallelization}
\label{sec:PCMES-IBO}

We consider a parallel extension, in which $Q$ queries can be evaluated at every iteration.
Due to a space limitation, we here only describe the main idea briefly (see Appendix~\ref{sec:parallel-detail} for more detail).
We employ a greedy selection of the $Q$ queries using the \emph{conditional mutual information} (CMI).
% For this purpose, we employ a greedy method using the \emph{conditional mutual information} (CMI).
%
Let
$\cX_q \coloneqq \{ \*x^{(1)}, \dots, \*x^{(q)} \}$
and
$\cH_q \coloneqq \{ \*h_{\*x^{(1)}}, \dots, \*h_{\*x^{(q)}} \}$
% be a set of input and output vectors already selected for a positive integer $q \leq Q$, respectively.
be sets of $q$ inputs and output vectors already selected in the greedy procedure ($q < Q$), respectively.
%
% This approach can be seen as an greedy approximation of
% $\max_{\cX_Q} {\rm MI}(\cH_Q ; f_*)$
% since
% We select a $(q+1)$-th query $\*x$ by maximizing
To maximize the MI after adding the $(q+1)$-th query $\*x$, i.e.,
${\rm MI}( \cH_{q} \cup \*h_{\*x} ; f_*)$,
we consider the decomposition
% \begin{align*}
\[
 {\rm MI}( \cH_{q} \cup \*h_{\*x} ; f_*) = {\rm MI}(\cH_{q} ; f_* ) + {\rm CMI}(\*h_{\*x} ; f_* | \cH_{q}),
\]
% \end{align*}
% and the first term does not depend on $\*x$.
where
${\rm CMI}(\*h_{\*x} ; f_* | \cH_{q}) \coloneqq {\EE_{\cH_{q}} \bigl[ {\rm MI}(\*h_{\*x} ; f_* | \cH_{q}) \bigl]}$.
Since the first term does not depend on $\*x$, we only need to maximize ${\rm CMI}(\*h_{\*x} ; f_* | \cH_{q})$.

% The $q$-th step of our greedy selection is defined as
% $\argmax_{\*x} {\rm MI}(\cH_q ; f_*)$,
% % $\argmax_{\*x} I(\*H_q ; f_*)$,
% where $\*x^{(1)}, \dots, \*x^{(q-1)}$ are already fixed by the previous steps.
% %
% This procedure maximizes the additional information gain produced by $\*h_{\*x}$, which can be seen through the following expansion:
% %
% \begin{align}
%  %I(\*H_q ; f_*) = I(\*H_{q-1} ; f_* ) + \EE_{\*H_{q-1}} \bigl[ I(\*h_{\*x} ; f_* | \*H_{q-1}) \bigl], \label{eq:joint_MI_sum}
%  % \rm{MI}(\*H_q ; f_*) = {\rm MI}(\*H_{q-1} ; f_* ) + \underbrace{\EE_{\*H_{q-1}} \bigl[ {\rm MI}(\*h_{\*x} ; f_* | \*H_{q-1}) \bigl]}_{\eqqcolon {\rm CMI}(\*h_{\*x} ; f_* | \*H_{q-1})},
%  % {\rm MI}(\cH_q ; f_*) = {\rm MI}(\cH_{q-1} ; f_* ) + \underbrace{\EE_{\cH_{q-1}} \bigl[ {\rm MI}(\*h_{\*x} ; f_* | \cH_{q-1}) \bigl]}_{\eqqcolon {\rm CMI}(\*h_{\*x} ; f_* | \cH_{q-1})},
%  {\rm MI}(\cH_q ; f_*) = {\rm MI}(\cH_{q-1} ; f_* ) + {\rm CMI}(\*h_{\*x} ; f_* | \cH_{q-1}), \nonumber
% %  \label{eq:joint_MI_sum}
% \end{align}
% where the second term is
% ${\rm CMI}(\*h_{\*x} ; f_* | \cH_{q-1}) \coloneqq {\EE_{\cH_{q-1}} \bigl[ {\rm MI}(\*h_{\*x} ; f_* | \cH_{q-1}) \bigl]}$.
% %
% Importantly, the first term does not depend on $\*x$ anymore.
% %
% Therefore, we obtain $\argmax_{\*x} {\rm MI}(\cH_q ; f_*) = \argmax_{\*x} {\rm CMI}(\*h_{\*x} ; f_* | \cH_{q-1})$, which means that the greedy selection can be performed by the maximization of CMI.

By the same approach as \eqref{eq:IBO}, we can obtain the lower bound of ${\rm CMI}(\*h_{\*x} ; f_* | \cH_{q})$ as
\begin{align}
 % \EE_{\*H_{q}} \bigl[ I(\*h_{\*x} ; f_* | \*H_{q}) \bigl] &\geq - \EE_{\*H_{q}, f_*} \bigl[ \log (1 - Z_{\*x}(f_*  |  \cH_{q}))  \bigl], \label{eq:PCMES-IBO}
 % {\rm CMI}(\*h_{\*x} ; f_* | \*H_{q}) &\geq - \EE_{\*H_{q}, f_*} \bigl[ \log (1 - Z_{\*x}(f_*  |  \cH_{q}))  \bigl], \label{eq:PCMES-IBO}
%  {\rm CMI}(\*h_{\*x} ; f_* | \cH_{q})
%  &\geq - \EE_{\cH_{q}, f_*} \bigl[ \log (1 - Z_{\*x}(f_*  |  \cH_{q}))  \bigl] \nonumber \\
%  &\eqqcolon L_{\rm par}(\*x) ,
%  L(\*x | \cX_{q} ) \coloneqq  - \EE_{\cH_{q}, f_*} \bigl[ \log \bigl(1 - Z_{\*x}(f_*  |  \cH_{q}) \bigl)  \bigl],
 L(\*x | \cX_{q} ) \coloneqq  - \EE_{\cH_{q}, f_*} \bigl[ \log \overline{Z}_{\*x}(f_*  |  \cH_{q})   \bigl],
 \label{eq:PCMES-IBO}
\end{align}
where
$\overline{Z}_{\*x}(f_*  |  \cH_{q}) \coloneqq \Pr (\*h_{\*x} \in \overline{\cA}_{f_*} | \cX_{q}, \cH_{q})$.
% $Z_{\*x}(f_*  |  \cH_{q}) \coloneqq \Pr (\*h_{\*x} \in \cA_{f_*} | \cX_{q}, \cH_{q})$.
%
See Appendix~\ref{sec:IBO} for the derivation.
% $Z_{\*x}(f_*  |  \cH_{q-1}) = \Pr (\*h_{\*x} \in \cA_{f_*} | \*X_{q-1}, \*H_{q-1})$.
%
% Let $m^{(f)}_{q}$ and $m^{(g_c)}_{q}$ be the predictive means, and ${s^{(f)}_{q}}^2$ and ${s^{(g_c)}_{q}}^2$ be the predictive variances after conditioning by $\cH_{q-1}$, respectively.
% Given $\cX_{q}, \cH_{q}$ and $f_*$, $Z_{\*x}(f_*  | \cH_{q})$ can be easily computed through the GPs updated by $\cX_{q}$ and $\cH_{q}$.
% \blue{by updating the GPs by $\cX_{q}$ and $\cH_{q}$}. \blue{(takeno:byの連続が気になりました.)}
%
% Suppose that $m^{(f)}_{\*x^{(q)}}$ and ${s^{(f)}_{\*x^{(q)}}}^2$ are the predictive mean and variance of $f(\*x)$ after conditioning by $\cH_{q}$, respectively, and that $\red{m^{(g_c)}_{\*x^{q}}$ and ${s^{(g_c)}_{q}}^2 (\*x)$ are those for $g_c(\*x)$.
% %
% Then, from the independence assumption,
% $Z_{\*x}(f_*  |  \cH_{q}) = \bigl(1 - \Phi(\eta^{(f)}_{q}(\*x))\bigl) \prod_{c=1}^C \bigl(1 - \Phi(\eta^{(g_c)}_{q}(\*x))\bigl)$, where
% $\eta^{(f)}_{q}(\*x) = \bigl(f_* - m^{(f)}_{\*x^{(q)}} \bigl) / s^{(f)}_{q}(\*x)$
% and
% $\eta^{(g_c)}_{q}(\*x) = \bigl(z_c - m^{(g_c)}_{\*x^{(q)}} \bigl) / s^{(g_c)}_{q}(\*x),\ c=1, \dots, C$.
%
% By applying the Monte Carlo estimation, we obtain the acquisition function for the $(q+1)$-th query as
By applying the MC estimation, we obtain the acquisition function for the $(q+1)$-th query as
\begin{align}
 % &\alpha^{\rm IBO}_{t}(\*x | \*X_{q}) = - \frac{1}{K} \sum_{(f_*, \*H_{q}) \in \cJ} \log(1-Z_{\*x}(f_*  |  \cH_{q})), \label{eq:PCMES-IBO_final}
 % &\alpha^{\rm IBO}(\*x | \cX_{q}) = - \frac{1}{K} \! \sum_{(f_*, \cH_{q}) \in \cJ} \! \! \! \log \bigl(1-Z_{\*x}(f_*  |  \cH_{q}) \bigl), \label{eq:PCMES-IBO_final}
 &\alpha^{\rm IBO}(\*x | \cX_{q}) = - \frac{1}{K} \! \sum_{(\tilde{f_*}, \cH_{q}) \in \cJ} \! \! \! \log \overline{Z}_{\*x}(\tilde{f}_*  |  \cH_{q}), \label{eq:PCMES-IBO_final}
\end{align}
where $\cJ$ is a set of $K$ sampled
$(\tilde{f}_*, \cH_{q})$
% $(f_*, \*H_{q})$
from the current GPs.
Note that jointly sampling $\tilde{f}_*$ and
$\cH_{q}$
can be easily performed by almost the same procedure as the sequential case,
and given $\tilde{f}_*$ and $\cH_{q}$, $\overline{Z}_{\*x}(\tilde{f}_*  | \cH_{q})$ can be easily computed through the GPs updated by $\cH_{q}$ (details are in Appendix~\ref{sec:sampling_fstar} and \ref{sec:parallel-detail}, respectively).
%
% Consequently, our parallel extension is also reduced to a computationally tractable form.
Consequently, our parallel extension is also reduced to a simple acquisition function.

% --------------------------------------------------
\section{Analysis}
\label{sec:analytical}

% In this section, we provide the error analysis for our CMES-IBO, which also holds in parallel extension.
% \red{
We here consider differences of CMES-IBO from a more direct application of the conventional MES, and we also provide analysis on the estimation variability of our estimator.
Although we focus on the sequential case, both of them also hold in the parallel case.
% }

% --------------------------------------------------
% \subsection{\red{Comparing Worst-case Error with Conventional MES}}
\subsection{Comparison with Conventional MES}
\label{sec:relation_MES}

Although we employ the lower bound based approach to derive CMES-IBO, \citet{Perrone2019-Constrained} have proposed a more direct extension of MES to CBO (Note that \citet{Perrone2019-Constrained} mainly focus on the binary setting, in which a binary value indicating feasible or not is only observed).
However, their derivation is restricted to only one constraint ($C=1$).
% \blue{, and \citet{Fernandezsanchez2020-maxvalue} described that the extension to multiple constraints is not obvious}.
%
Although \citet{Fernandezsanchez2020-maxvalue} described that the extension to multiple constraints is not obvious,
to consider the relation with CMES-IBO, we extend this original MES-based approach to the multiple constraints case, which we refer to as CMES.

In CMES, the MI is approximated as
\begin{align}
 & {\rm MI}(\*h_{\*x} ; f_*)
 =
 H(\*h_{\*x} )  - \mathbb{E}_{f_*}\bigl[ H(\*h_{\*x} \mid f_*) \bigl]
 \nonumber
 \\
 & \approx  H(\*h_{\*x} )  - \mathbb{E}_{f_*}\bigl[ H(\*h_{\*x} \mid \*h_{\*x} \in \overline{\cA}_{f_*}) \bigl]
 \label{eq:CMES_approx_truncated}
 \\
 & \approx
 \frac{1}{K}
 \! \!
 % \sum_{f_* \in \cF_*}
 \sum_{\tilde{f}_* \in \cF_*}
 \! \!
 \biggl\{
 % \frac{Z_{\*x}(f_*)}{2(1-Z_{\*x}(f_*))} R_{f_*} - \log(1-Z_{\*x}(f_*))
 \frac{Z_{\*x}( \tilde{f}_* )}{2 \overline{Z}_{\*x}( \tilde{f}_* ) } R_{\tilde{f}_*} - \log \overline{Z}_{\*x}( \tilde{f}_* )
 \biggl\}
 \nonumber
 \\
 &
 \eqqcolon \alpha^{\rm CMES}(\*x),
 \label{eq:CMES_final}
\end{align}
where
\begin{align*}
 % R_{f_*} &=  \frac{\gamma^{(f)}_{\*x}(f_*) \phi(\gamma^{(f)}_{\*x}(f_*))}{1 - \Phi(\gamma^{(f)}_{\*x}(f_*))} + \sum_{c=1}^C \frac{\gamma^{(g_c)}_{\*x}(z_c) \phi(\gamma^{(g_c)}_{\*x}(z_c))}{1 - \Phi(\gamma^{(g_c)}_{\*x}(z_c))},
R_{\tilde{f}_*} &=  \frac{\gamma^{(f)}_{\*x}(\tilde{f}_*) \phi(\gamma^{(f)}_{\*x}(\tilde{f}_*))}{1 - \Phi(\gamma^{(f)}_{\*x}(\tilde{f}_*))} + \sum_{c=1}^C \frac{\gamma^{(g_c)}_{\*x}(z_c) \phi(\gamma^{(g_c)}_{\*x}(z_c))}{1 - \Phi(\gamma^{(g_c)}_{\*x}(z_c))},
\end{align*}
and $H(\cdot)$ is the differential entropy and $\phi$ is the PDF of the standard normal distribution.
See Appendix~\ref{sec:CMES} for the derivation.
The first approximation \eqref{eq:CMES_approx_truncated} is
the replacement from
$p(\*h_{\*x} \mid f_*)$
to
$p(\*h_{\*x} \mid \*h_{\*x} \in \overline{\cA}_{f_*})$
in the entropy of the second term, and the second approximation that derives the next line
% \blue{\eqref{eq:CMES_final}}
is the MC estimation of $\EE_{f_*}$
% \blue{(takeno:番号を付ける位置が変わったため少しずれているので無くても良いかなと.)}
% \blue{(k1015:なんとかしてポインティングしたいです．読者視点で「二番目の近似」はそれほど明瞭ではないと思います)}
%
(Note that although the approximation strategy follows \citep{Perrone2019-Constrained}, $f_*$ here is based on our definition \eqref{eq:f_star} because $f_*$ should be a proper random variable to define the MI as we mentioned in Section~\ref{sec:max-value}).
% but the meaning of MI is essentially different from \citep{Perrone2019-Constrained} due to the definition of $f_*$ (The MI in \citep{Perrone2019-Constrained} is not mathematically well-defined).

Although CMES can be seen as a constrained counterpart of the original MES, the following lemma reveals that the MI approximation of CMES \eqref{eq:CMES_final} can be a negative value when $C > 5$.
\begin{lem}
 When $C > 5$, for every $\*x \in \cX$, there exist thresholds
 $\{z_c \}_{c=1}^C$
%  ${z_c (c = 1, \ldots, C)$
 that result in
 $\Pr(\alpha^{\rm CMES}(\*x) < 0) > 0$.
 \label{lem:CMES-negativity}
\end{lem}
\vspace{-.5em}
See Appendix~\ref{sec:proof-negative} for an interpretation and the proof.
%
% In contrast, the non-negativity of CMES-IBO is obvious from the definition \eqref{eq:CMES-IBO}.
In contrast, for CMES-IBO, we have the following remark:
% 一方，The non-negativity of CMES-IBO is obvious from the definition \eqref{eq:CMES-IBO}.
\begin{rema}
 % CMES-IBO guarantees non-negativity of the acquisition function $\alpha^{\rm IBO}(\*x) \geq 0$, and further, positivity $\alpha^{\rm IBO}(\*x) > 0$ is also guaranteed as far as the GPs have non-zero predictive variance ${\sigma^{(f)}_{n}}^2(\*x) > 0$ and ${\sigma^{(g)}_{n}}^2(\*x) > 0$.
 % Our acquisition function is positive $\alpha^{\rm IBO}(\*x) > 0$ for $\forall \*x \in \cX$.
 Our acquisition function is bounded from below by an average of probability of improvement (PI) from $\tilde{f}_*$, i.e.,
 % $\alpha^{\rm IBO}(\*x) \geq \sum_{f_* \in \cF_*} \Pr(\*h_{\*x} \in \cA_{f_*}) / K \geq 0$,
 $\alpha^{\rm IBO}(\*x) \geq \sum_{\tilde{f}_* \in \cF_*} \Pr(\*h_{\*x} \in \cA_{\tilde{f}_*}) / K \geq 0$,
 from which we also see non-negativity.
 \label{rem:IBO-non-negativity}
\end{rema}
\vspace{-.5em}
% \blue{(takeno: lower boundedよりこちらのほうが一般的かなと.)}
This remark is immediately derived by applying the well-known inequality $\log x \leq x - 1$ to \eqref{eq:CMES-IBO}.
Note that
% $\Pr(\*h_{\*x} \in \cA_{f_*})$
$\Pr(\*h_{\*x} \in \cA_{\tilde{f}_*})$
is PI from $\tilde{f}_*$ because it is the probability that
% $f(\*x) \geq f_*$
$f(\*x) \geq \tilde{f}_*$
and $g_c(\*x) \geq z_c$ for $\forall c$.
%
% The (unconstrained) MES also has shown a relation with PI \citep{Wang2017-Max}, while for CMES, it is difficult to see a clear relation with PI.
% Note that the unconstrained MES also has shown a relation with PI \citep{Wang2017-Max}.
The unconstrained MES also has shown a relation with PI \citep{Wang2017-Max}, while for CMES, it is difficult to see an interpretable relation with PI.
%
% \begin{rema}
Let the worst-case error be the largest error of the MI approximation among all the possible optimal-value samples $\cF_*$.
For problems in which $\Pr(\alpha^{\rm CMES}(\*x) < 0) > 0$, the worst-case error of CMES-IBO is smaller than that of CMES.
Another important implication of the remark is that although CMES-IBO considers a lower bound, $\alpha^{\rm IBO}(\*x)$ should be substantially larger than $0$ if $\*x$ is promising in the sense of the PI, which cannot be guaranteed for CMES as shown in the following toy example.

\begin{figure}
 \cigr{0.6}{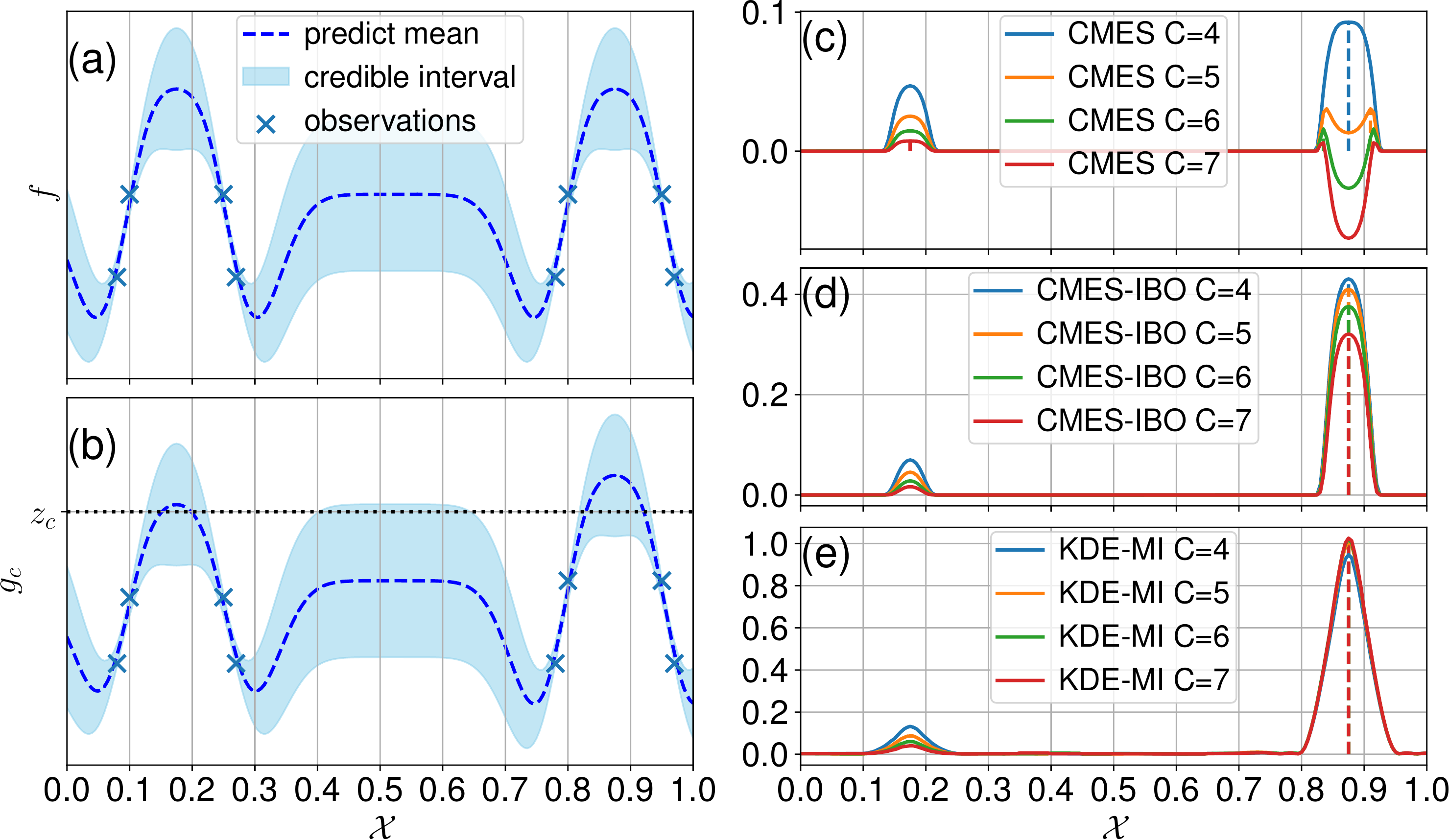}
 \caption{
 In (a) and (b), the GPs for $f$ and $g_c$ are shown, respectively.
 %
 % In (c) and (d), the acquisition functions of CMES-IBO and CMES \red{with $K=10$} for $C=4,5,6,$ and $7$ are shown, respectively.
 % \blue{, and the vertical dotted lines represent the maximum of each acquisition function.}
 % %
 % In (e), the MI approximated by \red{the} Monte Carlo \red{sampling} and kernel density estimations for $C=4,5,6,$ and $7$ are shown.
 %
 In (c)-(e), CMES, CMES-IBO, and the KDE-based MI approximation are shown, and each vertical dotted line indicates the maximum.
 % In (e), the MI approximated by Monte Carlo and kernel density estimations for $C=4,5,6,$ and $7$ are shown.
 }
 \label{fig:CMES-negative_example-10}
\end{figure}

% The negative value of CMES is not only a theoretical possibility.
%
% Figure~\ref{fig:CMES-negative_example-10} shows an example in a toy problem.
Figure~\ref{fig:CMES-negative_example-10} shows a toy example in which all constraints $g_1, \dots, g_C$ are the same function just for simplicity (note that the independence assumption on GP models does not mean that observations of different functions cannot take the same value) and change $C$ from $4$ to $7$.
%
% We assume that all constraints $g_1, \dots, g_C$ are the same function just for simplicity and change $C$ from $4$ to $7$.
%
CMES and CMES-IBO are calculated with $K = 10$.
In the figure, we also provide a numerical approximation of the original MI na{\"i}vely estimated by using the kernel density estimation (KDE), denoted as KDE-MI.
%
% For KDE-MI, we use a large number of samples ($10000$), which cannot be performed in practice.
For KDE-MI, since we used a large number of samples $10000$ (which cannot be performed in practice), we regard KDE-MI as a pseudo ground truth.
%
% To demonstrate negativity of CMES, the threshold $z_c$ (shown in Figure~\ref{fig:CMES-negative_example-10}~(b)) is set so that
To demonstrate negativity of CMES, the threshold $z_c$ (shown in Fig.~\ref{fig:CMES-negative_example-10}~(b)) is set so that
% $R_{f_*}$
$R_{\tilde{f}_*}$
in CMES \eqref{eq:CMES_final} becomes large negative value around $x = 0.9$ particularly when $C$ is large.
% The reason of this negativity is that $R_{f_*}$ in CMES \eqref{eq:CMES_final} becomes a large negative value in this problem particularly when $C$ is large.
% We design this problem so that \eqref{eq:CMES_final} results in negative value.
%
% For this purpose, we set $z_c$ is slightly smaller than $\mu_{t-1}^{(g_c)}(x)$ around $x = 0.9$, by which $R_{f_*}$ becomes large negative value particularly when $C$ is large.
%
% As a result, especially around $x = 0.9$, the acquisition function of CMES has a smaller value than that of CMES-IBO.
Then, in fact,
% CMES takes negative values around the global optima of the true MI approximation in the case of $C = 6$ and $7$.
% \blue{Moreover,}
CMES takes negative values around the global optimum of KDE-MI in the case of $C = 6$ and $7$.
% \blue{(this local decrease of CMES often occurs at points that have high $p(f_{\*x} \geq f_*)$ and $p(g_c(\*x) \geq z_c)$, which can also be interpreted from the definition \eqref{eq:CMES_final} explained in Appendix~\ref{sec:example_negative})}.
%
% \blue{On the other hand, CMES-IBO keeps the same global optimum as KDE-MI among $200$ grid points for all $C$.
% (takeno:200pointsの情報そこまで重要ではないかなと)}
%
In contrast, CMES-IBO keeps the same global optimum as KDE-MI for all $C$.
Although CMES has a negative value only when $C > 5$ as Lemma~\ref{lem:CMES-negativity} indicates, the acquisition values gradually decrease with the growth of $C$.
Although values of CMES-IBO are nearly half of KDE-MI, we can clearly see they are highly correlated.
Further detail of this example is in Appendix~\ref{sec:example_negative}.

Note that, in practice, the problematic behavior of CMES shown in Fig.~\ref{fig:CMES-negative_example-10}~(c) does not occur frequently depending on a given problem.
% the given problem setting.
%
Although we will see that CMES often reasonably works in later experiments, the potential risk of this negativity issue degrades the reliability of CMES.

% --------------------------------------------------
\subsection{Bounds for Estimation Variability}

We here analyze variability of our MI estimator \eqref{eq:CMES-IBO}, that is,
% $\alpha^{\rm IBO}(\*x) = - \frac{1}{K} \sum_{f_* \in \cF_*} \log  (1 - Z_{\*x}(f_*))$
how the MI estimator \eqref{eq:CMES-IBO} deviates from its expectation \eqref{eq:IBO}.
%
% From the simple form of \eqref{eq:CMES-IBO}, we obtain following theorem:
% \red{Since our MI estimator \eqref{eq:CMES-IBO} is simpler than CMES \eqref{eq:CMES_final}, we can obtain the following theorem:}
%
% \blue{The simplicity of our acquisition function \eqref{eq:CMES-IBO} enables us to derive the following theorem:}
The simplicity of our MI estimator \eqref{eq:CMES-IBO} enables us to derive the following theorem, while it is difficult to derive a similar guarantee for CMES:
\begin{theo}
 % For every $t \geq 1$ and $\*x \in \cX$, the variance of our MI estimator \eqref{eq:CMES-IBO} $\mathbb{V}[- \log(1 - Z_{\*x}(f_*))] \leq 2$.
 For every $\*x \in \cX$, our MI estimator \eqref{eq:CMES-IBO} satisfies
% $\mathbb{V}_{f_*}[- \log(1 - Z_{\*x}(f_*))] \leq 2$,
$\mathbb{V}_{f_*}[- \log\overline{Z}_{\*x}(f_*)] \leq 2$,
where $\mathbb{V}_{f_*}$ represents the variance with respect to $f_*$.
    Moreover, for every $\xi \geq 0$ and $\*x \in \cX$, following concentration inequality holds
    \begin{align}
        \Pr \bigl\{ |\alpha^{\rm IBO}(\*x) - L(\*x)| \geq \xi \bigl\}
        &\leq \min \left\{ U_{1}, U_{2} \right\},
        \label{eq:variability_bound}
    \end{align}
    where
 %\begin{align*}
 $U_1 = \frac{2}{K \xi^2}$ and
 $U_{2} = 2 \exp \biggl[ - A K \min \biggl( \frac{\xi^2}{B^2}, \frac{\xi}{B} \biggr)  \biggr]$
 with constants $A$ and $B$.
  % \end{align*}
 % and $A$ and $B$ are constants.
 \label{theo:CMES-IBO}
\end{theo}
See Appendix~\ref{sec:proof-ibo-var} for the proof.
%
% \blue{Since the acquisition function of CMES \eqref{eq:CMES_final} is more complicated, we could not obtain the same results for CMES. (takeno:短くしようとしました.)}

% The variance of our MI estimator can be bounded by small constant $2$, and this fact provides the bound by $U_1$ as a direct consequence of Chebyshev's inequality.
The variance of our MI estimator with $K=1$ can be bounded by the constant $2$, and this fact provides the bound by $U_1$ as a direct consequence of Chebyshev's inequality.
% \blue{(takeno: a direct consequence of/from の検索結果は ofのほうが多かったのでfromよりofのほうが自然かと思ったのですがいかがですか?)}
%
Most importantly, $U_{2}$ yields exponentially fast decay on $K$ and $\xi$ in contrast to the linear and quadratic dependence by $U_1$.
Although the convergence rate of $U_1$ is worse than $U_{2}$, $U_1$ may provide a tighter bound for specific values of $K$ and $\xi$.
% \red{(k1014:元の文章はちょっと意味が掴めなかったです. boundに対してin practiceって言うんですかね？漸近的な収束は良いけどある特定の設定では，と言うことでとどめていいかと思いました．variance by 2はU2の分子？のことなのかもですが，証明を見てないと出自がわからないので)}
% in practice owing to the tight bound of variance by $2$.
%
Note that our bound \eqref{eq:variability_bound} holds for any number of samplings $K$, i.e., it is not an asymptotic result such as the central limit theorem.
Moreover, $U_1$ and $U_{2}$ do not depend on any parameter of GPs and the problem setting, such as ${\sigma^{(f)}}^2(\*x), {\sigma^{(g_c)}}^2(\*x)$, and $C$.
%
% Consequently, we can expect an accurate estimation with a small $K$ in any situation, and it can be seen as a strong guarantee of the robustness of CMES-IBO.
% Consequently,
Our analysis can be seen as a guarantee of the estimation robustness of CMES-IBO even with a small $K$.

%%%%%%%%%%%%%%%%%%%%%%%%%%%%%%%%%%%%%%%%%%%%%%%%%%%%%%%%%%%%%%%%%%%%%%%%%%%%%%%%%%%%%%%%%%%%%%
\section{Related Work}
\label{sec:related_work}
%%%%%%%%%%%%%%%%%%%%%%%%%%%%%%%%%%%%%%%%%%%%%%%%%%%%%%%%%%%%%%%%%%%%%%%%%%%%%%%%%%%%%%%%%%%%%%

Extensions of unconstrained BO methods to CBO have been widely studied.
The standard expected improvement (EI) based approaches are called {\it EI with Constraint} (EIC) \citep{Schonlau1998-Global,Snoek2013-Bayesian,Gelbart2014-Bayesian,Gardner2014-Bayesian,Zhang2021-Constrained}.
However, when no feasible solution is obtained, the so-called `current best solution' can not be defined.
To avoid this difficulty, heuristic strategies, such as only using the feasible probability \citep{Gelbart2014-Bayesian} and introducing a threshold hyperparameter \citep{Letham2019-Constrained}, have been considered, but an appropriate remedy for this issue is still an open problem.
\citet{Gramacy2010-Optimization} and \citet{Picheny2014-Stepwise} consider the expected reduction of EI and PI, respectively.
However, since these approaches require expensive numerical integrations defined on the entire domain $\cX$, their applicability is limited.
A {\it Thompson Sampling} (TS) based method called {\it Scalable Constrained BO} (SCBO) \citep{Eriksson2020-Scalable} has been proposed, recently.
%
% Unlike the standard TS, which can be justified by strong theoretical results \citep{Kandasamy2018-Parallelised}, the convergence rate for the regret of SCBO has not been shown.
% The standard TS is justified by strong theoretical results \citep{Kandasamy2018-Parallelised}, but the convergence rate for the regret of SCBO has not been shown.
%
Since a next query in TS is directly determined by one sampled $\*x_*$, the decision often has high variability.
Further, when sampled $\cX_{\rm feasible}$ is empty, SCBO selects the input that has the smallest sum of the constraint violations, but the rationale of this selection in a sense of TS is not clarified.

The augmented Lagrangian (AL), which is a classical approach to white-box constrained problems, has also been combined with BO \citep{Gramacy2016-Modeling,Picheny2016-Bayesian}.
%
% ALBO needs Monte Carlo approximation, but Slack-AL replaces it with a one-dimensional quadrature by introducing \red{a} slack variable-based AL.
%
In AL-based methods, initial parameters of a Lagrange multiplier and a penalty coefficient should be specified.
%
% In the case of the `known' objective and `known' constraint setting in the classical numerical optimization, AL-based methods would be stable with respect to these settings because these values are refined during a large number of iterations.
% In the case of the `known' objective and constraints setting of the classical numerical optimization, AL-based methods are robust with respect to these settings because these values are refined during a large number of iterations.
In the classical white-box optimization, these parameters can be refined during a large number of iterations.
%
% However, for expensive black-box problems, the effect of the initial parameters is not negligible because the number of possible iterations is much smaller than the classical optimization.
However, for expensive black-box problems, the number of possible iterations is much smaller than the classical optimization, and the effect of the initial parameters can be significant.
%
% \citet{Ariafar2019-ADMMBO} proposed {\it Alternating Direction Method of Multipliers Bayesian Optimization} (ADMMBO) based on the famous ADMM algorithm.
\citet{Ariafar2019-ADMMBO} proposed {\it Alternating Direction Method of Multipliers BO} (ADMMBO) based on the famous ADMM algorithm.
ADMMBO is only for decoupled setting, in which each one of objective and constraint functions is observed separately.
In this paper, we only focus on the setting in which objective and constraint functions are simultaneously observed.

Entropy-based approaches are also studied in CBO literature.
\citet{Hernandez-Lobato2015-Predictive, Hernandez-Lobato2016-General} proposed {\it Predictive Entropy Search with Constraint} (PESC), which is an extension of a prior work on unconstrained problems \citep{Hernandez2014-Predictive}.
PESC considers the information gain of the optimum $\*x_*$, but complicated approximations, for which any relation with the original MI has not been clarified, are required.
% PESC considers the information gain of $\*x_*$ under the constraints, which is a measure of global utility and does not require any additional hyperparameters.
%
% However, PESC performs complicated approximations for which any relation with the original MI has not been clarified.
% However, PESC needs complicated and unstable approximations, for which any relation with the original MI has not been clarified.
% for which reliability has not been clarified \red{and a relation with the original MI has not been known.}
%
Another well-known entropy-based BO is MES, for which we already discussed in Section~\ref{sec:relation_MES}.
MES for multi-objective problems with constraints has been recently proposed by two papers \citep{Belakaria2020-maxvalue, Fernandezsanchez2020-maxvalue}.
%
% \citet{Fernandezsanchez2020-maxvalue} pointed out that the entropy evaluation of \citep{Belakaria2020-maxvalue} is obviously incorrect and further showed that the performance is inferior to the random search.
% \citet{Fernandezsanchez2020-maxvalue} pointed out that the entropy evaluation of \citep{Belakaria2020-maxvalue} is incorrect, and used the assumed density filtering for extending MES, but this approach results in complicated approximations by which the simplicity of the original MES is not maintained anymore.
\citet{Fernandezsanchez2020-maxvalue} pointed out that the entropy evaluation of \citep{Belakaria2020-maxvalue} is incorrect and derived a multi-objective counterpart of CMES \eqref{eq:CMES_approx_truncated}, but they introduced multiple additional approximations by which the relation with the original MI is further unclear.
Moreover, it is worth noting that none of the information-theoretic CBO studies have considered the possibility of $\cX_{\rm feasible} = \emptyset$.
See Appendix~\ref{sec:infeasibility} for a detailed discussion.
% on this problem.

A simple approach to incorporating multiple constraints is to transform it into a single constraint such as $g(\*x) \coloneqq \min_{c} (g_c(\*x) - z_c) \geq 0$.
% \red{with a threshold $0$}.
%
Both of fitting a GP directly to $g(\*x)$ and fitting GPs to $g_c(\*x)$ individually are possible strategies in this case, but the former discards observations of each $g_c(\*x)$ (only use the $\min$ value) and the latter makes $g(\*x)$ non-Gaussian.
The binary setting approach \citep{Perrone2019-Constrained} can also be used for this purpose, but then, observations of each $g_c(\*x)$ are discarded.
% ($g(\*x) \coloneqq 1 \text{ if } \min_{c} (g_c(\*x) - z_c) \geq 0$, otherwise $0$).
%

% In addition, by using CMI, we extend CMES-IBO (and CMES) to the parallel setting, which is made possible by the wide applicability and computational simplicity of MES.
%
The parallel extension of CBO has not been widely studied.
%
% \blue{
% \citet{Letham2019-Constrained} and \citep{Eriksson2020-Scalable} have shown the parallel CBO based on EIC and TS (SCBO), respectively.
% }
% \citet{Letham2019-Constrained} have proposed the parallel extension of EIC for noisy observations using the quasi-Monte Carlo method.
\citet{Letham2019-Constrained} have shown the parallel CBO based on EIC.
Although \citet{Eriksson2020-Scalable} have not focused on the parallelization, SCBO is applicable to the parallel setting easily because of the nature of TS.
%
% SCBO \citep{Eriksson2020-Scalable} is also applicable to the parallel setting easily because of the nature of TS.
%
% These approaches inherit the difficulties of their sequential counterparts mentioned in this section.
%
% in which lower fidelity observations are available with lower sampling costs,
For MES, although \citet{Takeno2020-Multifidelity} have proposed a CMI based parallel BO for the multi-fidelity setting, their approach is not based on a lower bound of CMI, and constrained problems have not been studied.
% A parallel extension of MES has been proposed by \citep{Takeno2020-Multifidelity} in the context of multi-fidelity BO.
%
% Although this also employs CMI, the acquisition function is not a lower bound of CMI, and constrained problems have not been studied.
%
\citet{Wilson2018-Maximizing} discussed the MC-based parallel BO in general, but they considered only `myopic maximal (MM) acquisition functions.'
MES is not included in the class of MM acquisition functions, and CBO was not considered in \cite{Wilson2018-Maximizing}.

To our knowledge, no existing studies use a lower bound of the MI for BO.
\citet{Moss2021-gibbon} recently proposed a lower bound based extension of MES.
However, their bound is a lower bound of the `approximate' MI used in the original MES, which is not a lower bound of the MI, and constrained problems have not been studied.
Although we only focus on constrained problems, considering the effectiveness of the lower bound approach to the unconstrained problem is one of our important future works.
%
% Therefore, the basic ideas are obviously different, and constrained problems have not been studied in \citep{Moss2021-gibbon}.
%
% Although we only focus on constrained problems in this paper, a similar lower bound can be derived even for the unconstrained case.
%
% Considering the effectiveness of the lower bound approach for the unconstrained problem is one of our important future works.

Lastly, we mention regret analysis of the entropy-based BO.
\citet{Wang2017-Max} show the regret analysis of `one sample MES,' in which only one optimal value is sampled in the MC estimation.
To our knowledge, no other regret analysis is known for the entropy-based BO (\cite{Belakaria2019-maxvalue} show an extension to the multi-objective problem, but their regret can be negative value as pointed out by \cite{Suzuki2020-multi}).
%
% However, we find that the original theorem of MES contains several critical flaws.
However, the original theorem of MES contains several technical problems.
For example, the theorem assumes that $f$ follows a GP, but the maximum of $f$ is regarded as a deterministic variable in the proof, which contradicts each other (more details including other issues are in Appendix~\ref{sec:about_regret_MES}).
%, among which we here briefly describe two examples (more details including other issues are shown in Appendix~\ref{sec:about_regret_MES}).
% However, we find that the original theorem of MES contains several critical flaws, among which we here briefly describe two examples (more details including other issues are shown in Appendix~\ref{sec:about_regret_MES}).
%
% First, the theorem assumes that $f$ follows GP, but the maximum of $f$ is regarded as a deterministic variable.
%
% Second, in their regret bound, the iteration $T$ is replaced with a subset of the entire iterations, and the dependency of the bound on the actual number of iterations is not clarified.
%
Therefore, convergence guarantee is still an open problem for the entropy-based BO.

%%%%%%%%%%%%%%%%%%%%%%%%%%%%%%%%%%%%%%%%%%%%%%%%%%%%%%%%%%%%%%%%%%%%%%%%%%%%%%%%%%%%%%%%%%%%%%
\section{Experiments}
\label{sec:experiments}
%%%%%%%%%%%%%%%%%%%%%%%%%%%%%%%%%%%%%%%%%%%%%%%%%%%%%%%%%%%%%%%%%%%%%%%%%%%%%%%%%%%%%%%%%%%%%%

We demonstrate the performance of sequential optimization by comparing with CMES, EIC \citep{Gelbart2014-Bayesian}, a TS-based method referred to as TSC, and PESC \citep{Hernandez-Lobato2015-Predictive} in Spearmint
(\url{https://github.com/HIPS/Spearmint/tree/PESC}).
% \footnote{\tiny \url{https://github.com/HIPS/Spearmint/tree/PESC}}.
% \blue{under an Academic and Non-Commercial Research Use License}
%
Note that CMES is based on our definition of $f_*$ \eqref{eq:f_star}.
TSC is a variant of SCBO \citep{Eriksson2020-Scalable} simplified by omitting the output transformation such as the Gaussian copula-based transformation of the objective and the trust region strategy.
Since these two are general strategies applicable to any acquisition functions, to only focus on the difference of acquisition functions, we do not employ them.
Although the AL-based methods \citep{Gramacy2016-Modeling,Picheny2016-Bayesian} are not shown as baselines, they are outperformed by PESC and SCBO in prior work \citep{Hernandez-Lobato2015-Predictive, Eriksson2020-Scalable}, respectively.

% Performances on GP-derived synthetic functions, benchmark functions, and a real-world problem were evaluated by using the \emph{utility gap}, which is also employed in various prior work \citep{Hernandez-Lobato2015-Predictive, Picheny2016-Bayesian, Eriksson2020-Scalable}.
Performances on GP-derived synthetic functions, benchmark functions, and two real-world problems were evaluated by using the \emph{utility gap} (UG), which is also employed in various prior work \citep{Hernandez-Lobato2015-Predictive, Picheny2016-Bayesian, Eriksson2020-Scalable}.
We set the recommendation at iteration $t$ as $\hat{\*x}_t = \argmax_{ \*x \in \cX } \mu^{(f)}(\*x),\ \text{s.t. } \forall c,\ \Pr(g_c(\*x) \geq z_c) \geq \sqrt[C]{0.95}$.
%
% Then, the utility gap is defined as $f_* - f(\hat{\*x}_t)$ if the recommendation is feasible, otherwise $f_* - \min_{x \in \cX} f(\*x)$, which indicates that an infeasible recommendation results in the worst utility gap.
Then, UG is defined as $f_* - f(\hat{\*x}_t)$ if the recommendation is feasible, otherwise $f_* - \min_{x \in \cX} f(\*x)$, which indicates that an infeasible recommendation results in the worst UG.
The initial inputs are sampled by Latin hypercube sampling \citep{Loh1996-Latin}.
The sample size of all MC approximations is set as $10$.
%
% For the parallel querying, we set $Q=3$.
%
% We used the method of moving asymptotes \citep{Krister2002-Class} in NLopt \citep{NLopt2008} and GPy \citep{GPy2014}, both of which are open-source libraries.
%
For the kernel function in GPs, we used a linear combination of the linear kernel
$k_{\rm LIN}: \cX \times \cX \to \RR$ and RBF kernel $k_{\rm RBF}: \cX \times \cX \to \RR$
defined as
$\sigma^2_{\rm LIN} k_{\rm LIN}(\*x,\*x') + \sigma^2_{\rm RBF} k_{\rm RBF}(\*x,\*x')$,
where $\sigma^2_{\rm LIN}$ and $\sigma^2_{\rm RBF}$ are updated by the marginal likelihood maximization every $5$ iteration, except for the GP-derived functions in which all the methods employed the same fixed kernels used to generate the true functions.
We used $k_{\rm LIN}$ because benchmark functions often contain linear functions.
For the GP-derived functions, we generated $10$ constrained problems for each of which the experiment is run $10$ times, and the mean and standard error of the total $100$ trials are reported.
For the benchmark and the real-word functions, we report the mean and the standard error of 10 random initializations.
%
% \blue{For PESC, due to the difficulty of rewriting the Spearmint specification, the default settings of kernels and inner optimization were used.
% %
% Although PESC has the same number of initial points, their location is not identical to other methods because of the same reason.}
%
% \blue{The infeasibility penalty $\zeta$ in CMES-IBO and CMES is set as $\zeta \rightarrow \infty$.}
%
Other experimental details such as the number of initial points are shown in Appendix~\ref{sec:detail_exp_setting}.
%
% The results for \red{sequential- and parallel-} querying are shown in Figure~\ref{fig:sequential} and \red{\ref{fig:parallel}}, respectively.

\begin{figure*}[!t]
    \centering
    %  \igr{.25}{Results_Inf_Gramacy_log_average.pdf}
    %  \igr{.249}{Results_Inf_Gardner2_log_average.pdf}%
    \igr{.16}{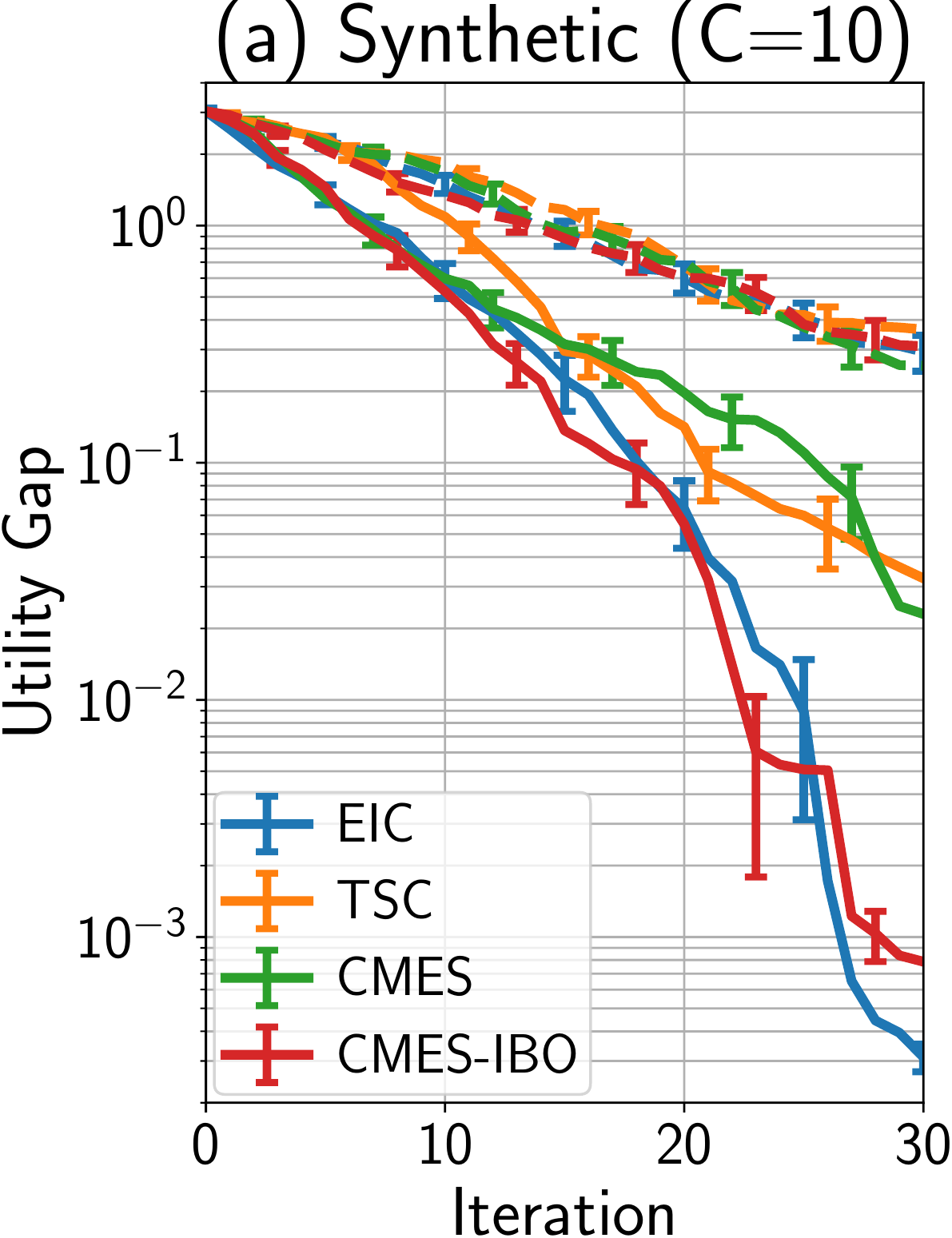}%
    \igr{.14}{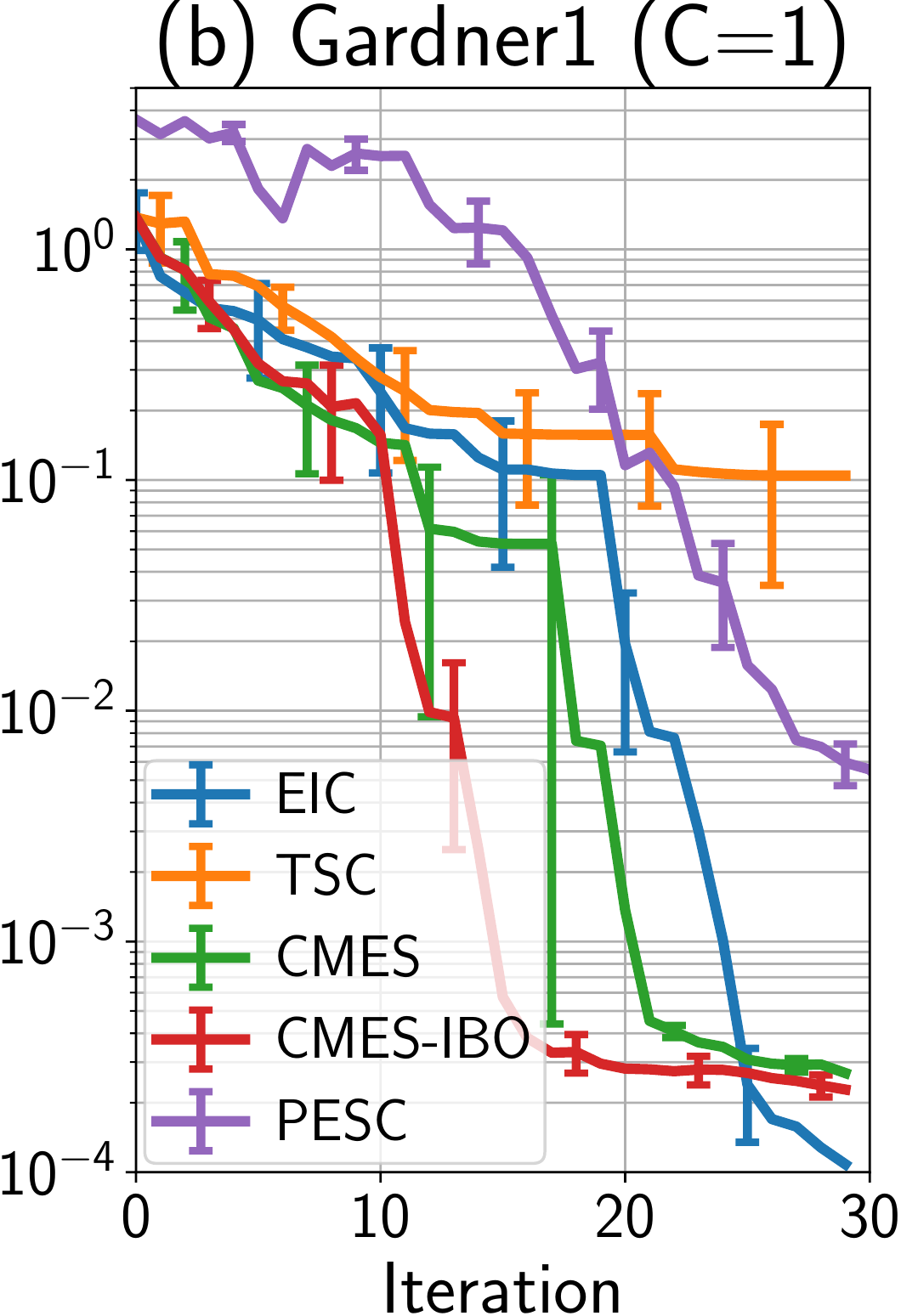}%
    % \igr{.16}{Results_Inf_Hartmann6_log_average.pdf}%
\igr{.14}{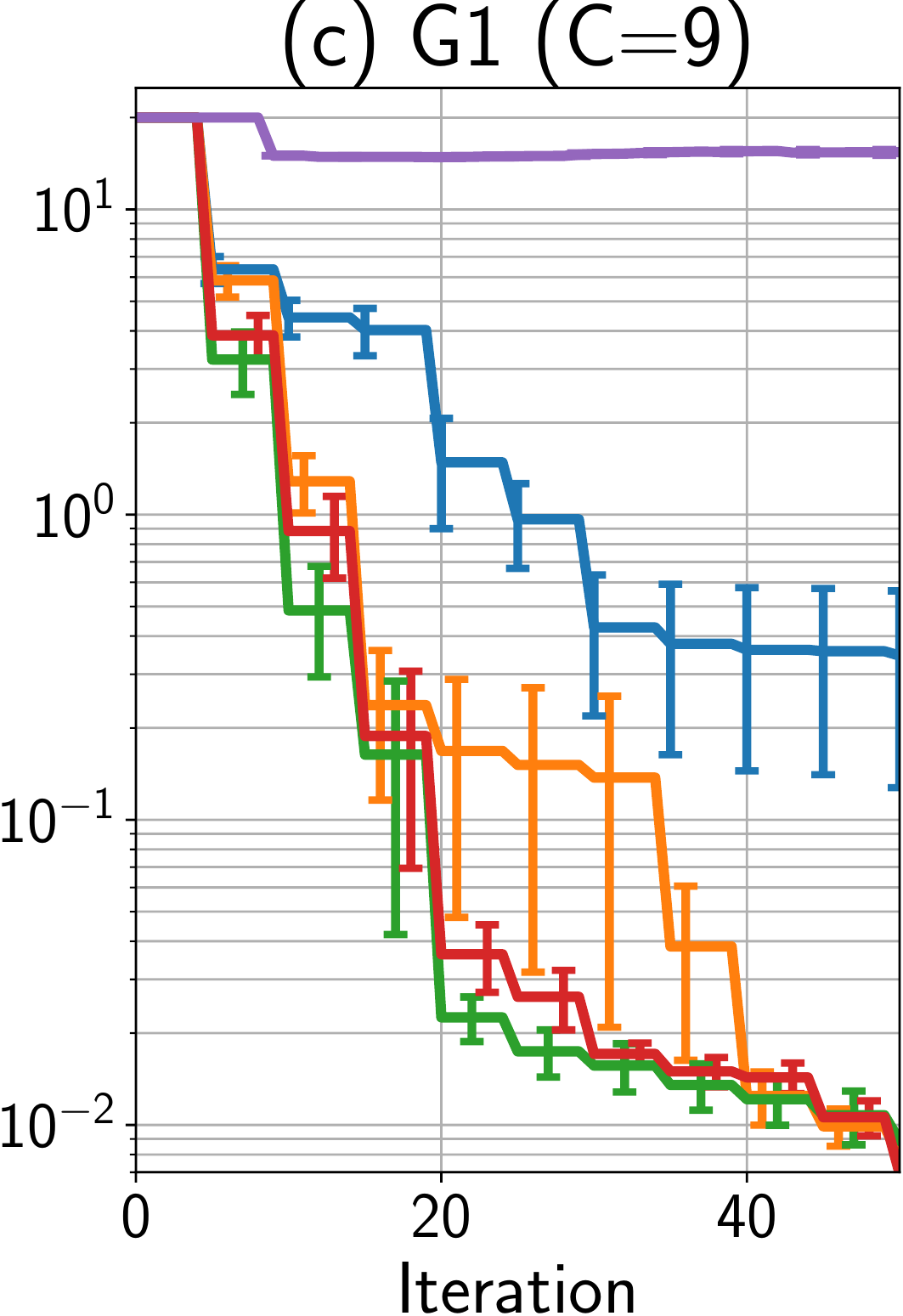}%
    \igr{.14}{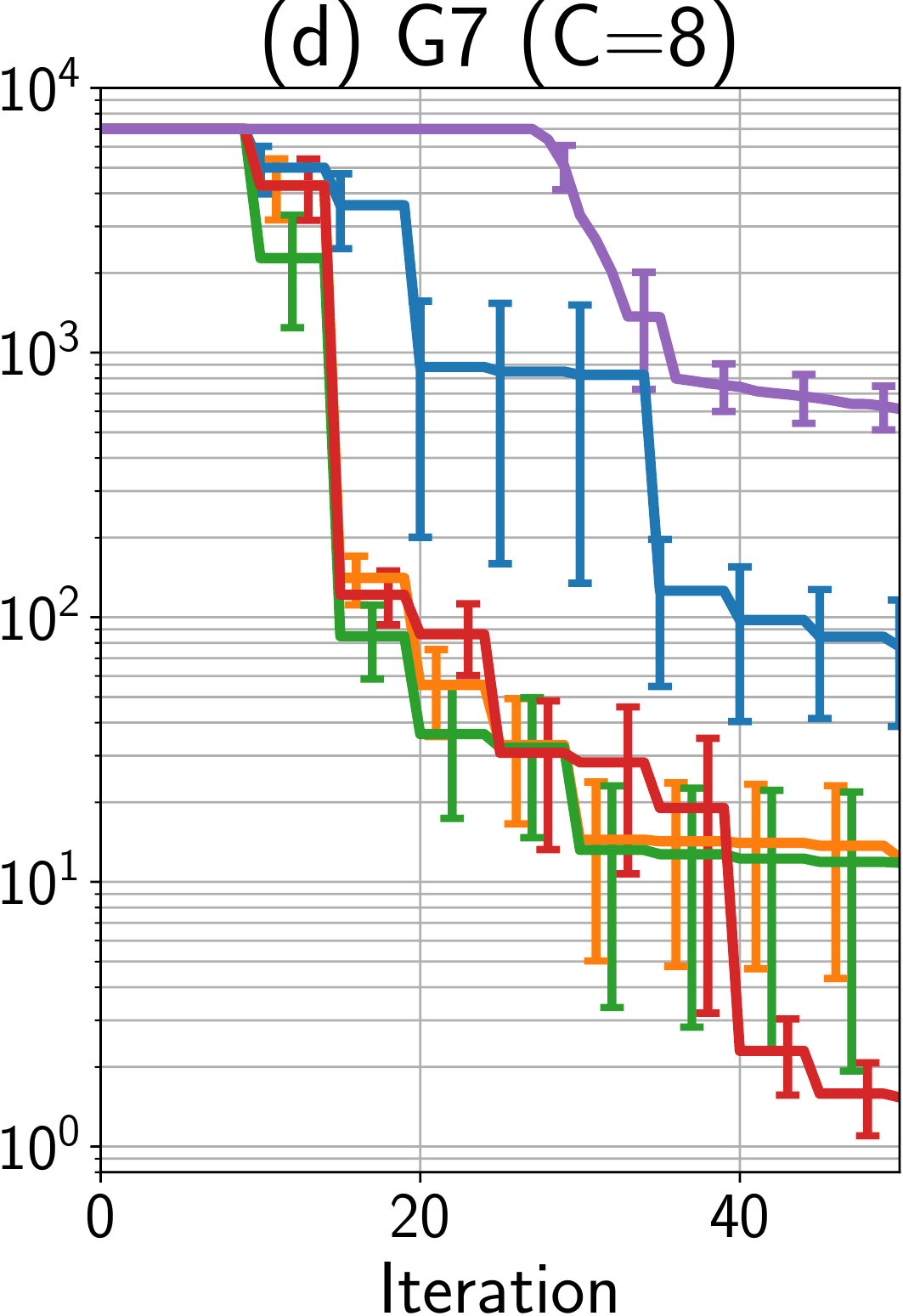}%
    \igr{.14}{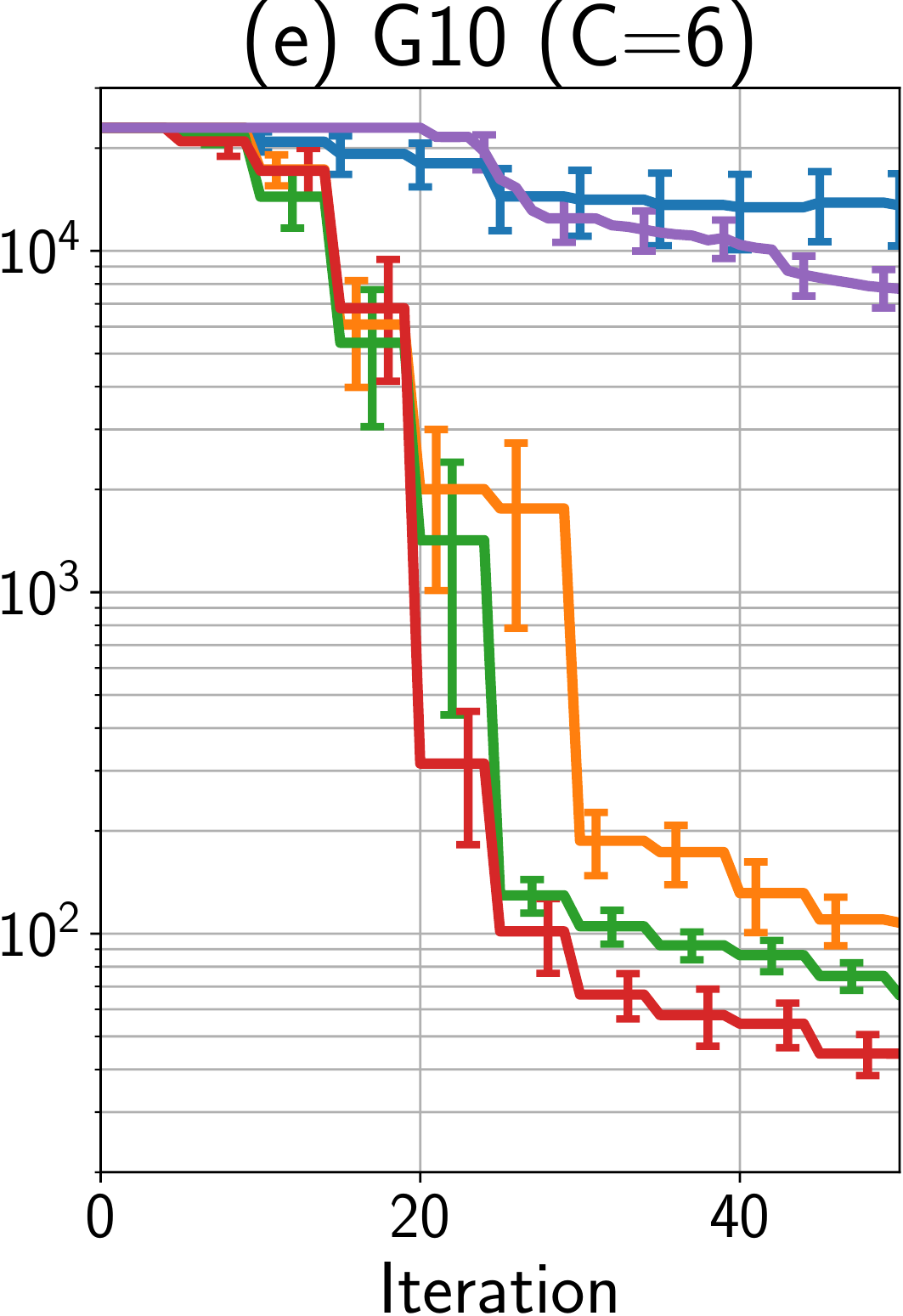}%
    \igr{.14}{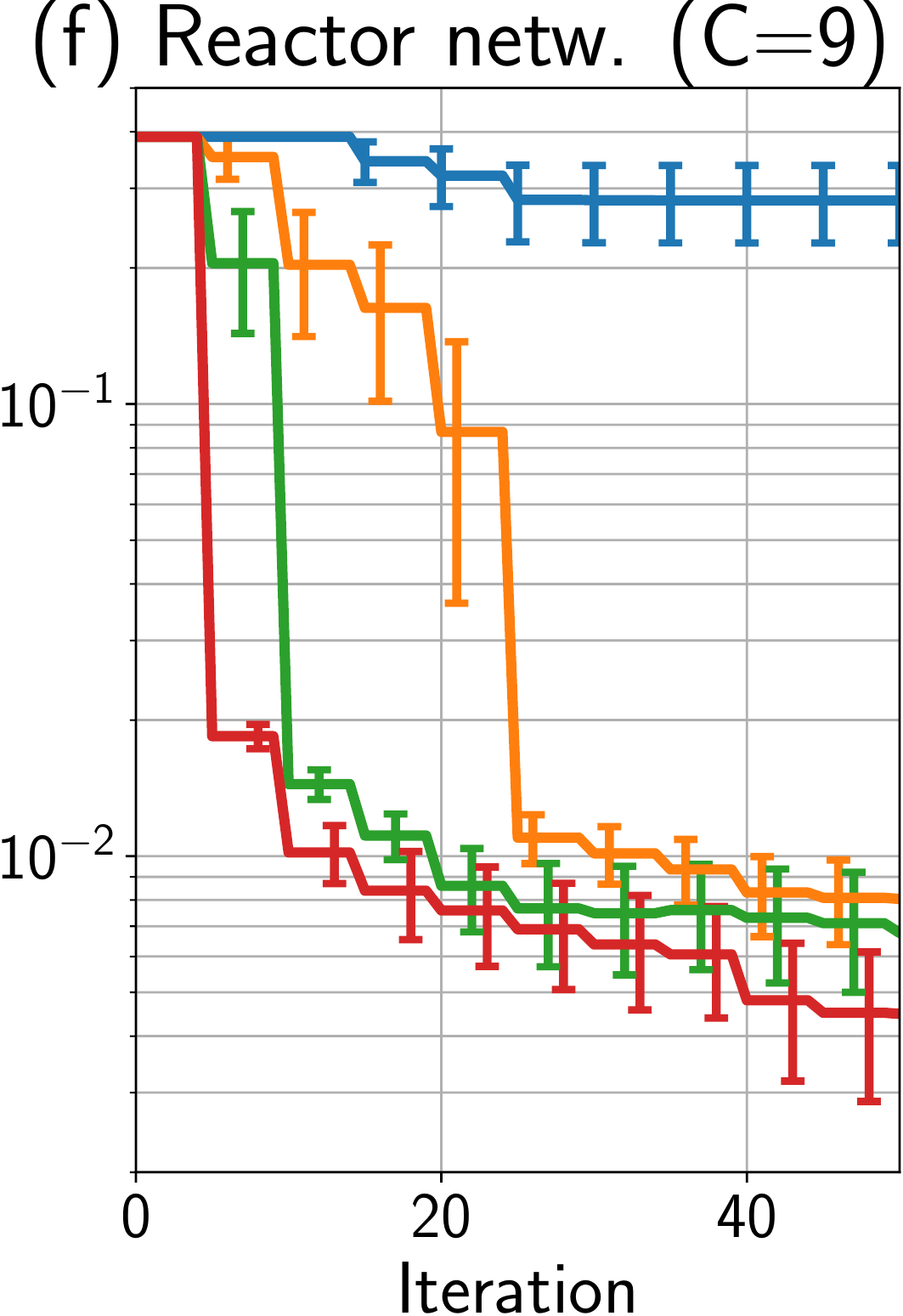}%
    \igr{.14}{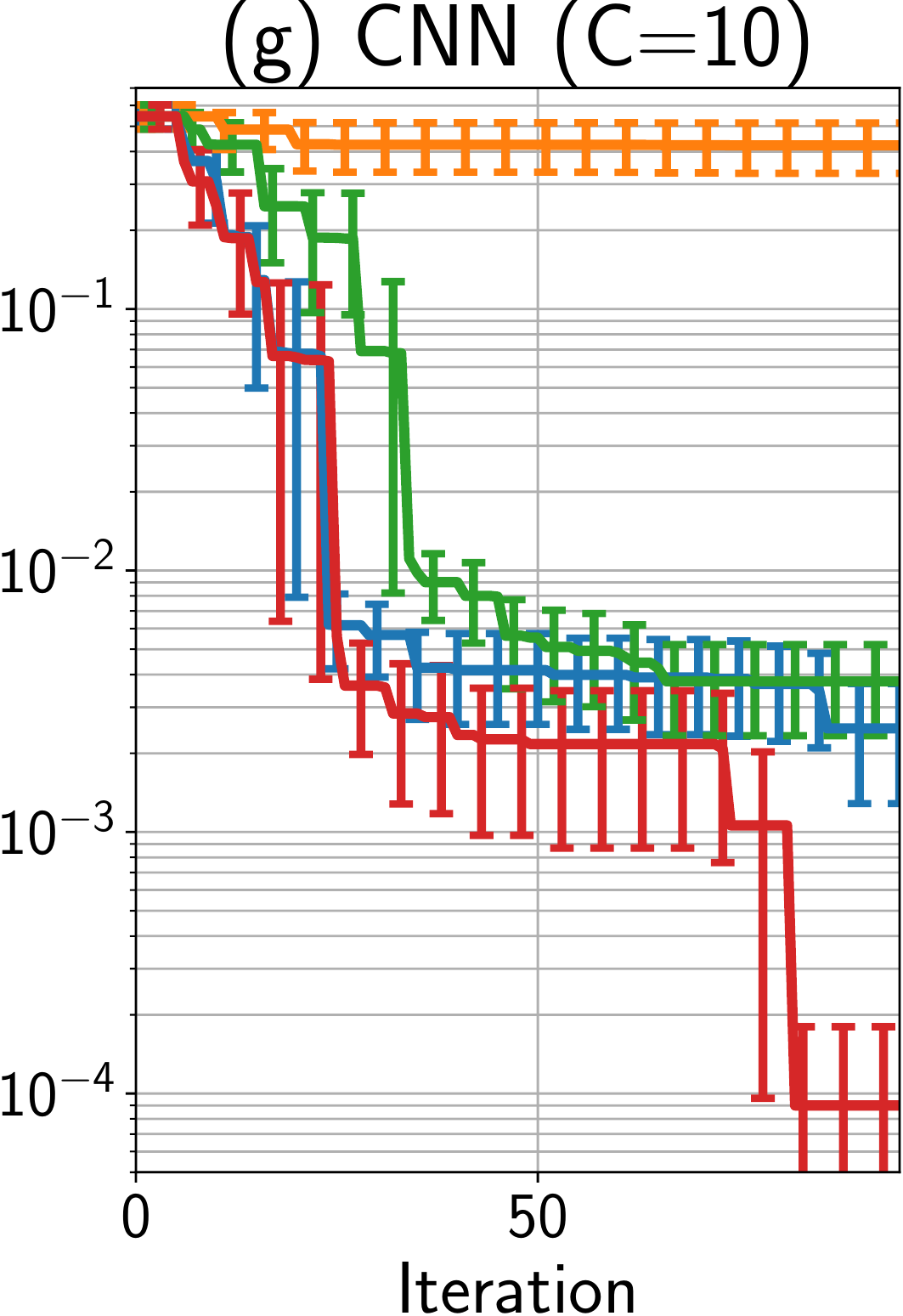}%

    \caption{
    Utility gap in sequential querying (average and standard error).
    %
    % The dashed lines in the synthetic function experiment represent one constraint counterpart of each method that has the same color.
    The dashed lines in the synthetic function experiment represent the `single constraint' counterpart of each method that has the same color.
    %
    % The solid line and error bar express the average and the standard error, respectively.
    }
    \label{fig:sequential}
\end{figure*}

% First, we focus on the results for the sequential querying shown in Figure~\ref{fig:sequential}.
First, we focus on the results for the sequential querying shown in Fig.~\ref{fig:sequential}.

%%%%%%%%%%%%%%%%%%%%%%%%%%%%%%%%%%%%%%%%%%%%%
% \paragraph{\red{Evaluation on synthetic functions}}
% \paragraph{\red{GP-derived} synthetic functions}
{\bf GP-derived synthetic functions (a)}:
%%%%%%%%%%%%%%%%%%%%%%%%%%%%%%%%%%%%%%%%%%%%%
% First, we evaluate the performance on GP-derived synthetic functions.
%
The objective and the $C = 10$ constraints are sampled from zero-mean GPs with RBF kernels whose length scales are $0.2$.
% and the length scales of the kernels are $0.2$.
%
% The threshold for constraints are set as $z_c = -0.75$ for $\forall c$, and the input domain is $[0, 1]^2$.
The thresholds are set as $z_c = -0.75$ for $\forall c$, and the input domain is $[0, 1]^2$.
%
% We sampled $10$ \red{constrained problems for each of which the experiment is run $10$ times, and the mean and standard error of the total $100$ trials are reported.}
%
% Thus, in the synthetic experiment, we report the mean and standard error of the $100$ trials.
%
% In this experiment, we set the same kernel used to sample each function for all the methods.
% \red{All the methods used the same kernels used to sample the problem.}
%
We see that EIC and CMES-IBO outperformed other methods, and in particular, CMES is clearly worse than CMES-IBO.
%
% \red{Note that a slow convergence of PESC may be caused by the misspecification of the kernels that we already described.}
%
Additionally, we run the experiments with the `single constraint' transformation of multiple constraints described in Section~\ref{sec:related_work}, in which a GP fits to $\min_{c=1,\dots, 10} (g_c(\*x) - z_c )$ as one constraint.
The results shown by dashed lines indicate that this approach deteriorated performance compared with their multiple constraints counterpart.
% Additionally, we run the experiments for one constraint counterpart, which fit a GP to $g(\*x) = \min_{c=1,\dots, 10} (g_c(\*x) - z_c )$ as one constraint, to validate the effectiveness of the formulation for multiple constraints.
%
% These results, shown by dashed lines, highly deteriorate from each multiple constraints counterpart.

{\bf Benchmark functions (b)-(e)}:
We here show results on four benchmark functions called Gardner1 ($C = 1$ and $d = 2$), G1 ($C=9$ and $d = 13$), G7 ($C=8$ and $d = 10$), and G10 ($C=6$ and $d = 8$).
Gardner1 \citep{Gardner2014-Bayesian} is a simple test problem constructed by sine and cosine functions (see Appendix~\ref{sec:detail_exp_setting} for the detailed definition).
G1, G7, and G10 are from \citep{Liang2006-problem}.
In CBO literature, empirical evaluations are typically performed with $C \leq 2$, and thus, the settings $C = 9, 8$ and $6$ are large.
Moreover, these three functions have large input dimensions $d$.
% Moreover, these three functions have large input dimensions $d$, and the estimated ratio between the feasible region and the input space is very small (less than \red{$0.1\%$}) \citep{Liang2006-problem}.
% \red{(takeno:G1は0.0111\%)}
%
We see that CMES-IBO has superior or comparable to the other methods in all the four plots.
% In Gardner1, the utility gap of CMES-IBO converged clearly faster than the other methods.
CMES also showed reasonable performance, but we observed substantial differences compared with CMES-IBO in Gardner1, G7, and G10.
% In G7, although the utility gaps of CMES and P-CMES were not lower than about $10$, CMES-IBO converged to lower values.
In G7, the average UG of CMES was larger than $10$, while CMES-IBO converged to a much smaller value than $10$.
%
% The difference of the average utility gap of CMES and CMES-IBO in G10 is larger than $30$ after iteration $20$.
%
% The confidence intervals of CMES and CMES-IBO do not overlap after iteration $30$, and this indicates that they have a significant difference.
%
% Note that the slow convergence of PESC can be because Spearmint does not employ the linear kernel though these test functions contain linear functions.
%
In G10, the differences of the average UG between CMES and CMES-IBO may seem small because of the large vertical scale, but after the iteration $30$, the differences are larger than $30$, and the error bars are not overlapped.
For the benchmark functions, the results of PESC are also shown.
Note that due to the difficulty of rewriting the Spearmint specification, some settings were not consistent with the other methods (see Appendix \ref{sec:detail_exp_setting} for detail).
However, we can see that the performance of PESC is particularly unstable for (c)-(e), which is presumably caused by the difficulty of approximating information gain of the high dimensional $\*x_*$.

{\bf Reactor network design (f):}
%
% \red{First, w}e used a reactor network design problem ($d=6$) \citep{Manousiouthakis1992-global,Kumar2020-test}, in which
%
In this problem, the product concentration from a sequence of two continuously stirred tank reactors is optimized with capital cost and physical phenomenon constraints \citep{Manousiouthakis1992-global,Kumar2020-test}.
The input dimension is $d=6$.
This problem has one inequality constraint and four equality constraints, and we replaced each equality constraint with two inequality constraints, by which we have $C = 9$ inequality constraints.
%
% Since the equality constraints cannot be strictly satisfied when evaluating the utility gap, we set the tolerance error as $10^{-3}$.
Since the equality constraints cannot be strictly satisfied when evaluating UG, we set the tolerance error as $10^{-3}$.
We can see that CMES-IBO also shows faster convergence than other methods in this problem.

{\bf Hyperparameter optimization of neural network (g):}
%
% Second, we tested the neural network (NN) hyperparameter optimization for class-imbalanced datasets.
We tested a hyperparameter optimization of convolutional neural network (CNN) for a class-imbalanced setting, in which, typically, performance for minority classes can deteriorate.
% Typically, weighting to the loss function is used to guarantee the recall for a minority class.
%
% To prevent recall of minority classes from
%
% Thus, we optimize the accuracy under the constraints that the recall of each class is larger than $0.5$.
We consider optimizing the average accuracy of all classes under the constraints that the recall of each one of classes is larger than $0.5$.
We fitted the two-layer CNN to the CIFAR10 dataset \citep{Krizhevsky2009-Learning}, in which we set the class sizes of labels $0,...,4$ as $2500$, and the others are $5000$.
To control the imbalanceness, we gave a weight $w_m$ to the loss function of class $m = 1, \ldots, 10$ defined by $w_m = \bigl(N / (M N_m)\bigl)^\rho$, where $M = 10$ is the number of classes, $N$ is the size of the training data, $N_m$ is the size of class $m$, and $\rho$ is a hyperparameter.
%
% Setting $\rho = 0$ and $\rho = 1$ correspond to no weighting, and the default weighting in the scikit-learn, respectively.
Setting $\rho > 0$ gives larger weights to minority classes (scikit-learn employs $\rho = 1$\footnote{\url{https://scikit-learn.org/stable/modules/generated/sklearn.utils.class_weight.compute_class_weight.html}}), while $\rho = 0$ is reduced to the uniform weighting.
% $\rho = 0$ and $\rho = 1$ correspond to no weighting, and the default weighting in the scikit-learn, respectively.
%
In addition to $\rho \in [0, 2]$, we also tune other hyperparameters such as learning rate (1 dim), dropout rate (1 dim), and the number of channels (2 dim).
% We tune $5$ hyperparameters: learning rate, dropout rate, the number of channels, and the coefficient of the weight $\rho \in [0, 2]$, in which the weights are set as $\bigl(N / (M N_m)\bigl)^\rho$, where $M = 10$ is the number of class, $N$ is the number of data, $N_m$ for $m=1,\dots, 10$ is the each class size.
%
%
As a result, we have $d = 5$ and $C = M = 10$.
% Therefore, $d = 5$ and $C = M = 10$.
%
% Other details of NN are shown in Appendix~\ref{sec:detail_exp_setting}.
Other details are shown in Appendix~\ref{sec:detail_exp_setting}.
%
% CMES-IBO and EIC showed fast convergence, and in contrast to EIC, the utility gap of CMES-IBO reduced to a highly small value.
CMES-IBO and EIC reduced UG rapidly at the beginning of the iterations, and CMES-IBO achieved a substantially smaller value than the other methods at the end of the iterations.

\begin{figure}[t]
% \centering
%  \subfloat[Sequential Querying]{%
%  \igr{.25}{Results_Inf_G7_log_average.pdf}%
%  \igr{.25}{Results_Inf_G10_log_average.pdf}%
%  }%
 % \centering
%  \subfloat[Parallel Querying]{%
 \centering
 \igr{.15}{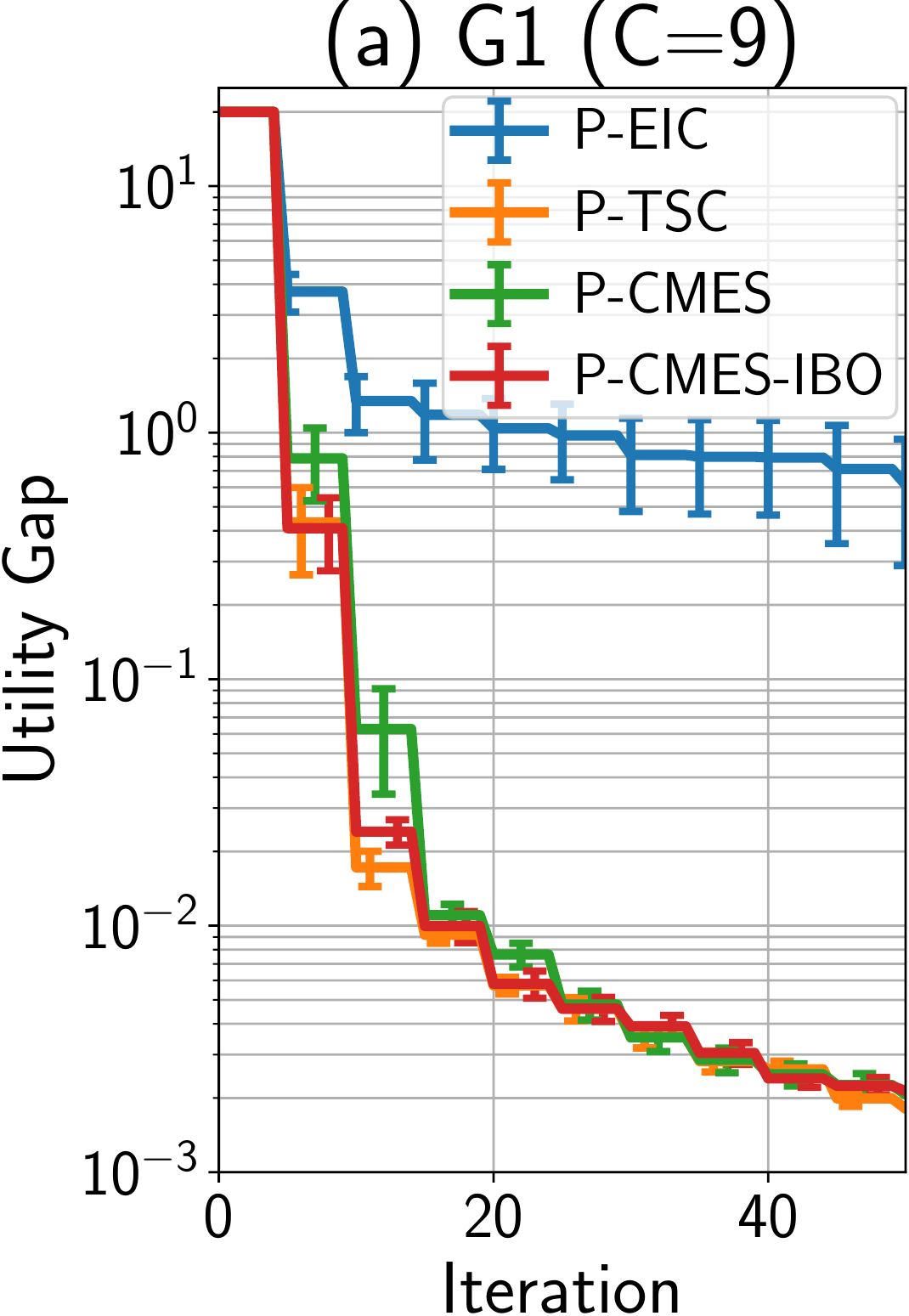}%
 \igr{.15}{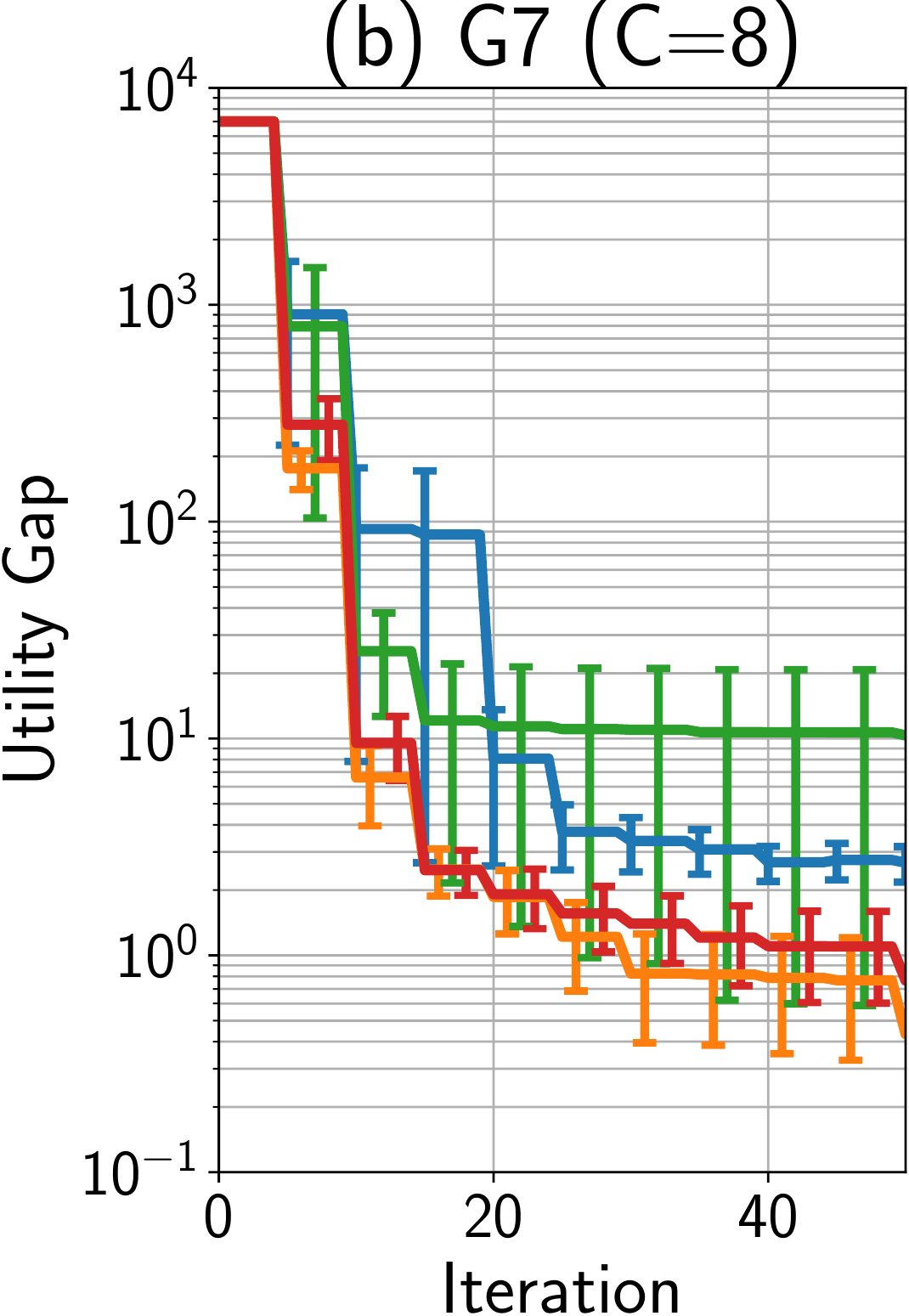}%
 \igr{.15}{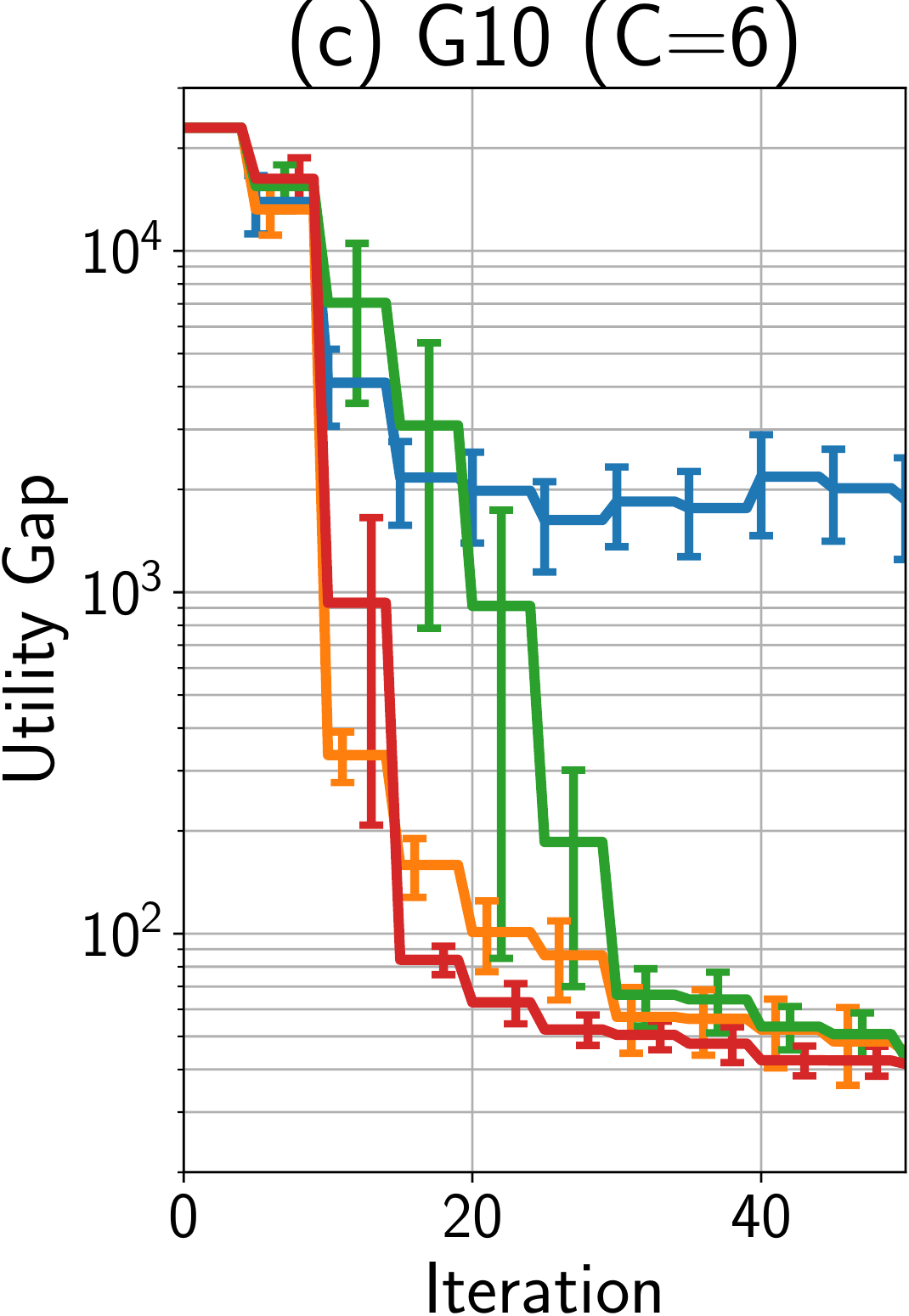}%
%  }
 \caption{
 Utility gap in parallel querying (average and standard error).
 % The solid line and error bar express the average and the standard error, respectively.
 }
 \label{fig:parallel}
\end{figure}

{\bf Parallel Querying}:
For the parallel setting, we parallelize EIC and TSC (see Appendix~\ref{sec:detail_exp_setting} for detail) and set the number of parallel queries $Q=3$.
The prefix `P-' indicates a parallelized variant of each method (e.g., P-EIC and P-CMES).
%
% The results are shown in Figure~\ref{fig:parallel}.
The results are shown in Fig.~\ref{fig:parallel}.
Here, G1, G7, and G10 benchmarks were used, and we see that, similarly to the sequential case, CMES-IBO shows efficient convergences compared with the other methods.

% Overall, EIC and TSC work well in some cases but often show worse performance.
Overall, EIC and TSC showed reasonable performance, but in some cases, their performances become much worse compared with the best method in each plot by being trapped at local optima.
%
% Although CMES showed the performance that does not depend on much on the function, in large $C$ experiments such as (a) and (f) in Fig.~\ref{fig:sequential}, CMES deteriorates compared to CMES-IBO.
The performance of CMES was mostly stable compared with EIC and TSC.
%
% However, particularly in large $C$ experiments such as (a) and (f) in Fig.~\ref{fig:sequential}, CMES deteriorates compared to CMES-IBO.
However, particularly in some large $C$ experiments such as (a) and (g) in Fig.~\ref{fig:sequential}, CMES deteriorates compared to CMES-IBO.
This result is consistent with our analysis that CMES can show problematic behavior when $C$ is large.

% Additional evaluations including for different benchmark functions and for different sample sizes $K$ are shown in Appendix~\ref{sec:additional_experiments}.
Additional evaluations for different sample sizes $K$, and the correlated extension are shown in Appendix~\ref{sec:additional_experiments}.

% We evaluate the performance of the CMES-IBO on correlated synthetic functions in Appendix~\ref{sec:experiment-correlated}.

%%%%%%%%%%%%%%%%%%%%%%%%%%%%%%%%%%%%%%%%%%%%%

%%%%%%%%%%%%%%%%%%%%%%%%%%%%%%%%%%%%%%%%%%%%%%%%%%%%%%%%%%%%%%%%%%%%%%%%%%%%%%%%%%%%%%%%%%%%%%
\section{Conclusion}
\label{sec:conclusion}
%%%%%%%%%%%%%%%%%%%%%%%%%%%%%%%%%%%%%%%%%%%%%%%%%%%%%%%%%%%%%%%%%%%%%%%%%%%%%%%%%%%%%%%%%%%%%%
%
We proposed an information-theoretic constrained Bayesian optimization method called CMES-IBO, derived from a lower bound of the MI.
We showed its several desired properties, such as the interpretation of the infeasible case, non-negativity, and the bounds of the estimation variability.
Moreover, we extended CMES-IBO to the parallel setting.
The effectiveness of our proposed method was shown through various benchmark functions and real-world problems.

\clearpage

%%%%%%%%%%%%%%%%%%%%%%%%%%%%%%%%%%%%%%%%%%%%%%%%%%%%%%%%%%%%%%%%%%%%%%%%%%%%%%%%%%%%%%%%%%%%%%
\section*{Acknowledgments}
This work was supported by MEXT KAKENHI 17H04694, 18K04700, 21H03498, JSPS KAKENHI Grant Number JP21J14673, and RIKEN Junior Research Associate Program.

%%%%%%%%%%%%%%%%%%%%%%%%%%%%%%%%%%%%%%%%%%%%%%%%%%%%%%%%%%%%%%%%%%%%%%%%%%%%%%%%%%%%%%%%%%%%%%
% \section*{Reference}
%%%%%%%%%%%%%%%%%%%%%%%%%%%%%%%%%%%%%%%%%%%%%%%%%%%%%%%%%%%%%%%%%%%%%%%%%%%%%%%%%%%%%%%%%%%%%%
\bibliographystyle{plainnat}
\bibliography{ref}
%%%%%%%%%%%%%%%%%%%%%%%%%%%%%%%%%%%%%%%%%%%%%%%%%%%%%%%%%%%%%%%%%%%%%%%%%%%%%%%%%%%%%%%%%%%%%%

\clearpage

\appendix
\onecolumn
% \thispagestyle{empty}
% \begin{center}
%  {\Large {\bf
%  Supplementary Materials for the Submission:\\
%  ``Sequential- and Parallel- Constrained Max-value Entropy Search via Information Lower Bound''
%  }
%  }
% \end{center}

%%%%%%%%%%%%%%%%%%%%%%%%%%%%%%%%%%%%%%%%%%%%%%%%%%%%%%%%%%%%%%%%%%%%%%%%%%%%%%%%%%%%%%%%%%%%%
%%%%%%%%%%%%%%%%%%%%%%%%%%%%%%%%%%%%%%%%%%%%%%%%%%%%%%%%%%%%%%%%%%%%%%%%%%%%%%%%%%%%%%%%%%%%%
\section{Definition of Gaussian Processes}
\label{sec:definition_GPs}
%%%%%%%%%%%%%%%%%%%%%%%%%%%%%%%%%%%%%%%%%%%%%%%%%%%%%%%%%%%%%%%%%%%%%%%%%%%%%%%%%%%%%%%%%%%%%%
% We use the standard GP models \citep{Rasmussen2005-Gaussian} for modeling $f$ and $g_1, \dots, g_C$.
We use standard GP models \citep{Rasmussen2005-Gaussian} for modeling $f$ and $g_1, \dots, g_C$.
Suppose that the observations are contaminated with additive Gaussian noises as $y^{(f)}_{\*x} = f(\*x) + \varepsilon$ and $y^{(g_c)}_{\*x} = g_c(\*x) + \varepsilon$ for $\forall c$, where $\varepsilon \sim \cN(0, \sigma^2_{\rm noise})$.
%
% Let $\cD^{(f)}_{n} = \{ (\*x_i, y^{(f)}_{\*x_i})\}_{i=1}^n$ and $\cD^{(g_c)}_{n} = \{ (\*x_i, y^{(g_c)}_{\*x_i})\}_{i=1}^n$ for $\forall c$.
%
% We employ the $\cG \cP(0, k(\*x, \*x^\prime))$ for the GP prior , where $k: \cX \times \cX \mapsto \RR$ is a kernel function .
For the GP prior, we employ $\cG \cP(0, k(\*x, \*x^\prime))$, which indicates that the prior mean is $0$ and covariance is specified by a kernel function $k: \cX \times \cX \mapsto \RR$.
%
% Note that, for simplicity, we here assume the same kernels and noise variances for all the functions, but both settings can be changed depending on the function.
%
Then, the predictive distribution for $f(\*x)$ given $\cD^{(f)}_{n} = \{ (\*x_i, y^{(f)}_{\*x_i})\}_{i=1}^n$ is derived as follows:
\begin{align*}
    f(\*x) | \cD^{(f)}_{n} &\sim \cN (\mu^{(f)}(\*x), {\sigma^{(f)}}^2(\*x)), \\
    \mu^{(f)}(\*x) &= \*k^\top \bigl( \*K + \sigma_{\rm noise}^2 \*I \bigl)^{-1} \*y^{(f)}, \\
    {\sigma^{(f)}}^2(\*x) &= k(\*x, \*x) - \*k^\top \bigl( \*K + \sigma_{\rm noise}^2 \*I \bigl)^{-1} \*k,
\end{align*}
where $\*y^{(f)} = \bigl( y^{(f)}_{\*x_1}, \dots, y^{(f)}_{\*x_n} \bigl)^\top$, $\*k = \bigl( k(\*x, \*x_1), \dots, k(\*x, \*x_n) \bigl)^\top$, $\*K \in \RR^{n\times n}$ is a kernel matrix whose $(i, j)$ element is defined as $k(\*x_i, \*x_j)$, and $\*I \in \RR^{n\times n}$ is the identity matrix.
The predictive distributions of $g_c$ can be obtained in the same way using the observations $y^{(g_c)}_{\*x_1}, \dots, y^{(g_c)}_{\*x_n}$.
% $\*y^{(g_c)}  = \bigl( y^{(g_c)}_{\*x_1}, \dots, y^{(g_c)}_{\*x_n} \bigl)^\top$ for $\forall c$.
%
Note that, for simplicity, we here assume the same kernels and noise variances for all the functions, but both the settings can be changed depending on the function.

%%%%%%%%%%%%%%%%%%%%%%%%%%%%%%%%%%%%%%%%%%%%%%%%%%%%%%%%%%%%%%%%%%%%%%%%%%%%%%%%%%%%%%%%%%%%%%
\section{Derivation of Information Lower Bound}
\label{sec:IBO}
%%%%%%%%%%%%%%%%%%%%%%%%%%%%%%%%%%%%%%%%%%%%%%%%%%%%%%%%%%%%%%%%%%%%%%%%%%%%%%%%%%%%%%%%%%%%%%
To derive our lower bound, we start from the following representation of the MI:
\begin{align}
 I(\*h_{\*x} ; f_*)
 &= \EE_{f_*} \biggl[ \int p(\*h_{\*x} | f_*) \log \frac{p(\*h_{\*x} | f_*)}{p(\*h_{\*x})} {\rm d} \*h_{\*x}  \biggl].
 \label{eq:MItmp}
\end{align}
%
% A simple way to define an MI between $\*h_{\*x}$ and $f_*$ is to interpret $p(f_*)$ as a mixture distribution in which one of the components (the green distribution in Fig.~\ref{fig:truncation}) consists of a single constant value ($-\infty$).
%
Note that since $p(f_*)$ can be seen as a mixture distribution in which one of the components (the green distribution in Fig.~\ref{fig:truncation}) consists of a single constant value ($-\infty$), the expectation regarding $f_*$ for a given function $b: \RR \cup \{-\infty\} \mapsto \RR$ can be defined as
\begin{align}
 \EE_{f_*} \bigl[ b(f_*) \bigl]
 &=
 \Pr(f_* = -\infty) b(-\infty) + \Pr( f_* \neq -\infty) \int_{\RR} p( f_* | f_* \neq -\infty) b(f_*) {\rm d} f_*.
 \label{eq:definition_expectation}
\end{align}
The MI \eq{eq:MItmp} defined through this expectation can be seen as a special case of an MI for a mixture distribution shown in \citep{Nair2007-On,Beknazaryan2019-On}.
% See \citep{Nair2007-On,Beknazaryan2019-On} for further details of an MI with a mixture distribution.
% which can be seen as an expectation over the mixture distribution of discrete and continuous random variables
% ($p(f_*)$ can be seen as a mixture distribution in which one of the components (the green distribution in Fig.~\ref{fig:truncation}) consists of a single constant value ($-\infty$))

% Since $f_*$ follows the mixture distribution of discrete and continuous random variables, the MI can be defined as follows \citep{Nair2007-On,Beknazaryan2019-On}:
% \begin{align*}
%     I(\*h_{\*x} ; f_*)
%     &= \EE_{f_*} \biggl[ \int p(\*h_{\*x} | f_*) \log \frac{p(\*h_{\*x} | f_*)}{p(\*h_{\*x})} {\rm d} \*h_{\*x}  \biggl].
% \end{align*}
% where the expectation regarding $f_*$ \red{for a function $b(f_*): \RR \cup \{-\infty\} \mapsto \RR$} can be expanded as
% \red{
%     \begin{align}
%         \EE_{f_*} \bigl[ b(f_*) \bigl]
%         &=
%         \Pr(f_* = -\infty) b(-\infty) + \Pr( f_* \neq -\infty) \int_{\RR} p( f_* | f_* \neq -\infty) b(f_*) {\rm d} f_*.
%         \label{eq:definition_expectation}
%     \end{align}
% }
% \blue{
% \begin{align*}
%     \EE_{f_*} \biggl[ \int p(\*h_{\*x} | f_*) \log \frac{p(\*h_{\*x} | f_*)}{p(\*h_{\*x})} {\rm d} \*h_{\*x}  \biggl]
%     &=
%     \Pr(f_* = -\infty) \int p(\*h_{\*x} | f_* = -\infty) \log \frac{p(\*h_{\*x} | f_* = -\infty)}{p(\*h_{\*x})} {\rm d} \*h_{\*x} \\
%     &\ + \Pr( f_* \neq -\infty) \int p( f_* | f_* \neq -\infty) \int p(\*h_{\*x} | f_*) \log \frac{p(\*h_{\*x} | f_*)}{p(\*h_{\*x})} {\rm d} \*h_{\*x} {\rm d} f_*.
% \end{align*}
% }
% %
Then, our information lower bound can be derived as follows:
\begin{align}
 I(\*h_{\*x} ; f_*)
%  &= \int \int p(\*h_{\*x}, f_*) \log \frac{p(\*h_{\*x}, f_*)}{p(\*h_{\*x}) p(f_*)} {\rm d} \*h_{\*x} {\rm d} f_*
%  \nonumber \\
 &= \EE_{f_*} \biggl[ \int p(\*h_{\*x} | f_*) \log \frac{p(\*h_{\*x} | f_*)}{p(\*h_{\*x})} {\rm d} \*h_{\*x}  \biggl]
 \nonumber \\
 % &= \EE_{f_*} \biggl[ \int p(\*h_{\*x} | f_*) \biggl( \log \frac{p(\*h_{\*x} | \*h_{\*x} \in \overline{\cA}_{f_*})}{p(\*h_{\*x})} + \log \frac{p(\*h_{\*x} | f_*)}{p(\*h_{\*x} | \*h_{\*x} \in \overline{\cA}_{f_*})} \biggl) {\rm d} \*h_{\*x}  \biggl]
 &= \EE_{f_*} \biggl[ \int p(\*h_{\*x} | f_*) \biggl( \log \frac{{q(\*h_{\*x} | f_*)}}{p(\*h_{\*x})} + \log \frac{p(\*h_{\*x} | f_*)}{{q(\*h_{\*x} | f_*)}} \biggl) {\rm d} \*h_{\*x}  \biggl]
 \nonumber \\
 % &= \EE_{f_*} \biggl[ \int p(\*h_{\*x} | f_*) \log \frac{p(\*h_{\*x} | \*h_{\*x} \in \overline{\cA}_{f_*})}{p(\*h_{\*x})} {\rm d} \*h_{\*x} +  D_{\rm KL} \bigl( p(\*h_{\*x} | f_*) || p(\*h_{\*x} | \*h_{\*x} \in \overline{\cA}_{f_*}) \bigl) \biggl]
 &= \EE_{f_*} \biggl[ \int p(\*h_{\*x} | f_*) \log \frac{{q(\*h_{\*x} | f_*)}}{p(\*h_{\*x})} {\rm d} \*h_{\*x} +  D_{\rm KL} \bigl( p(\*h_{\*x} | f_*) || {q(\*h_{\*x} | f_*)} \bigl) \biggl]
 \nonumber \\
 % &\geq \EE_{f_*} \biggl[ \int p(\*h_{\*x} | f_*) \log \frac{p(\*h_{\*x} | \*h_{\*x} \in \overline{\cA}_{f_*})}{p(\*h_{\*x})} {\rm d} \*h_{\*x}  \biggl]. \label{eq:VLB_App_B}
 &\geq \EE_{f_*} \biggl[ \int p(\*h_{\*x} | f_*) \log \frac{{q(\*h_{\*x} | f_*)}}{p(\*h_{\*x})} {\rm d} \*h_{\*x}  \biggl]
  \nonumber \\
 &= \EE_{f_*} \biggl[ \int p(\*h_{\*x} | f_*) \log \frac{p(\*h_{\*x} | \*h_{\*x} \in \overline{\cA}_{f_*})}{p(\*h_{\*x})} {\rm d} \*h_{\*x}  \biggl]. \label{eq:VLB_App_B}
\end{align}
In the last line, we replace $q(\*h_{\*x} | f_*)$ with $p(\*h_{\*x} | \*h_{\*x} \in \overline{\cA}_{f_*})$ as defined in \eqref{eq:TMN}.
Here, from the definition of $f_*$, we see $p(\*h_{\*x} | f_*) = p(\*h_{\*x} | \*h_{\*x} \in \overline{\cA}_{f_*}) = 0$ for $\*h_{\*x} \in \cA_{f_*}$.
Additionally, in information theory, $0 \log 0$ is treated as zero based on that $\lim_{x \rightarrow 0} x \log x = 0$ \citep{Cover2006-Elements}.
%
% \clearpage
Thus, we obtain
\begin{align*}
    % I(\*h_{\*x} ; f_*) &\geq
    {\eqref{eq:VLB_App_B}} &=
%    \EE_{f_*} \biggl[ \int_{\overline{\cA}_{f_*}} p(\*h_{\*x} | f_*) \log \frac{p(\*h_{\*x})}{(1 - Z_{\*x}(f_*))p(\*h_{\*x})} {\rm d} \*h_{\*x}  \biggl] \\
    \EE_{f_*} \biggl[ \int_{\overline{\cA}_{f_*}} p(\*h_{\*x} | f_*) \log \frac{p(\*h_{\*x})}{ \overline{Z}_{\*x}(f_*) p(\*h_{\*x})} {\rm d} \*h_{\*x}  \biggl] \\
%     &= - \EE_{f_*} \biggl[ \int_{\overline{\cA}_{f_*}} p(\*h_{\*x} | f_*) \log ( 1 - Z_{\*x}(f_*) ) {\rm d} \*h_{\*x}  \biggl]  \\
    &= - \EE_{f_*} \biggl[ \int_{\overline{\cA}_{f_*}} p(\*h_{\*x} | f_*) \log \overline{Z}_{\*x}(f_*) {\rm d} \*h_{\*x}  \biggl]  \\
%    &= - \EE_{f_*} \biggl[ \log ( 1 - Z_{\*x}(f_*) ) \int_{\overline{\cA}_{f_*}} p(\*h_{\*x} | f_*) {\rm d} \*h_{\*x}  \biggl]  \\
    &= - \EE_{f_*} \biggl[ \log \overline{Z}_{\*x}(f_*) ) \int_{\overline{\cA}_{f_*}} p(\*h_{\*x} | f_*) {\rm d} \*h_{\*x}  \biggl]  \\
%    &= - \EE_{f_*} [ \log ( 1 - Z_{\*x}(f_*) ) ].
    &= - \EE_{f_*} [ \log  \overline{Z}_{\*x}(f_*) ].
\end{align*}
Note that although the expectation $\EE_{f_*}$ is written as the mixture \eqref{eq:definition_expectation}, the MC approximation can be directly performed because a sample from the mixture distribution can be obtained through the procedure described in Section~\ref{ssec:sampling-fstar}, by which our acquisition function \eqref{eq:CMES-IBO} is derived.
% In more details, from the definition \eqref{eq:definition_expectation}, this can be expanded as
% \begin{align*}
%     - \EE_{f_*} [ \log  \overline{Z}_{\*x}(f_*) ]
%     &= - \Pr(f_* = -\infty) \log \overline{Z}_{\*x}(-\infty)
%     - \Pr( f_* \neq -\infty) \int_{\RR} p( f_* | f_* \neq -\infty) \log \overline{Z}_{\*x}(f_*) {\rm d} f_*.
% \end{align*}
% Our acquisition function \eqref{eq:CMES-IBO} is derived as the MC approximation of $- \EE_{f_*} [ \log  \overline{Z}_{\*x}(f_*) ]$.

%%%%%%%%%%%%%%%%%%%%%%%%%%%%%%%%%%%%%%%%%%%%%%%%%%%%%%%%%%%%%%%%%%%%%%%%%%%%%%%%%%%%%%%%%%%%%%
\section{Details of Parallelization}
\label{sec:parallel-detail}
%%%%%%%%%%%%%%%%%%%%%%%%%%%%%%%%%%%%%%%%%%%%%%%%%%%%%%%%%%%%%%%%%%%%%%%%%%%%%%%%%%%%%%%%%%%%%%
In this section, we provide the detailed derivation for the parallel setting omitted in the main paper.
% we consider a parallel extension of CMES-IBO.
%
We assume that $Q$ workers are available by which we can select $Q$ queries as a batch at every iteration.
Let
$\cX_q \coloneqq \{ \*x^{(1)}, \dots, \*x^{(q)} \}$
% $\*X_q$
be a set of input queries
and
$\cH_q \coloneqq \{ \*h_{\*x^{(1)}}, \dots, \*h_{\*x^{(q)}} \}$
% $\*H_q$
be a set of output vectors for an integer $1 \leq q \leq Q$.
A na{\"i}ve extension is to maximize MI between
% $\*H_Q$
$\cH_Q$
and $f_*$, i.e.,
% $I(\*H_Q ; f_*)$,
${\rm MI}(\cH_Q ; f_*)$,
with respect to the queries
$\cX_Q$.
% $\*X_Q$.
%
However, this approach results in a $Qd$ dimensional acquisition function maximization, which is often extremely hard.
Instead, we propose a greedy approximation of
$\max_{\cX_Q} {\rm MI}(\cH_Q ; f_*)$
% $\max_{\*X_Q} I(\*H_Q ; f_*)$
using the \emph{conditional mutual information} (CMI).

The $q$-th step of our greedy selection is defined as
$\argmax_{\*x^{(q)}} {\rm MI}(\cH_q ; f_*)$,
% $\argmax_{\*x^{(q)}} I(\*H_q ; f_*)$,
where $\*x^{(1)}, \dots, \*x^{(q-1)}$ are already fixed by the previous steps.
This procedure maximizes the additional information gain produced by $\*h_{\*x^{(q)}}$, which can be seen through the following expansion:
\begin{align}
 %I(\*H_q ; f_*) = I(\*H_{q-1} ; f_* ) + \EE_{\*H_{q-1}} \bigl[ I(\*h_{\*x^{(q)}} ; f_* | \*H_{q-1}) \bigl], \label{eq:joint_MI_sum}
 % \rm{MI}(\*H_q ; f_*) = {\rm MI}(\*H_{q-1} ; f_* ) + \underbrace{\EE_{\*H_{q-1}} \bigl[ {\rm MI}(\*h_{\*x^{(q)}} ; f_* | \*H_{q-1}) \bigl]}_{\eqqcolon {\rm CMI}(\*h_{\*x^{(q)}} ; f_* | \*H_{q-1})},
 % {\rm MI}(\cH_q ; f_*) = {\rm MI}(\cH_{q-1} ; f_* ) + \underbrace{\EE_{\cH_{q-1}} \bigl[ {\rm MI}(\*h_{\*x^{(q)}} ; f_* | \cH_{q-1}) \bigl]}_{\eqqcolon {\rm CMI}(\*h_{\*x^{(q)}} ; f_* | \cH_{q-1})},
 {\rm MI}(\cH_q ; f_*) = {\rm MI}(\cH_{q-1} ; f_* ) + {\rm CMI}(\*h_{\*x^{(q)}} ; f_* | \cH_{q-1}), \nonumber
%  \label{eq:joint_MI_sum}
\end{align}
where the second term is
${\rm CMI}(\*h_{\*x^{(q)}} ; f_* | \cH_{q-1}) \coloneqq {\EE_{\cH_{q-1}} \bigl[ {\rm MI}(\*h_{\*x^{(q)}} ; f_* | \cH_{q-1}) \bigl]}$.
Importantly, the first term does not depend on $\*x^{(q)}$ anymore.
Therefore, we obtain $\argmax_{\*x^{(q)}} {\rm MI}(\cH_q ; f_*) = \argmax_{\*x^{(q)}} {\rm CMI}(\*h_{\*x^{(q)}} ; f_* | \cH_{q-1})$, which means that the greedy selection can be performed by the maximization of CMI.

Our lower bound approach used in \eqref{eq:IBO} can be applied to CMI as follows:
\begin{align}
 % \EE_{\*H_{q-1}} \bigl[ I(\*h_{\*x} ; f_* | \*H_{q-1}) \bigl] &\geq - \EE_{\*H_{q-1}, f_*} \bigl[ \log (1 - Z_q)  \bigl], \label{eq:PCMES-IBO}
 % {\rm CMI}(\*h_{\*x} ; f_* | \*H_{q-1}) &\geq - \EE_{\*H_{q-1}, f_*} \bigl[ \log (1 - Z_q)  \bigl], \label{eq:PCMES-IBO}
 {\rm CMI}(\*h_{\*x^{(q)}} ; f_* | \cH_{q-1}) &\geq - \EE_{\cH_{q-1}, f_*} \bigl[ \log (1 - Z_{\*x^{(q)}}(f_* | \cX_{q-1}))  \bigl]
 \eqqcolon L_{\rm par}(\*x^{(q)}), \tag{\ref{eq:PCMES-IBO}}
%  \label{eq:PCMES-IBO}
\end{align}
where
$Z_{\*x^{(q)}}(f_* | \cX_{q-1})  \coloneqq \Pr (\*h_{\*x^{(q)}} \in \cA_{f_*} | \cX_{q-1}, \cH_{q-1})$.
The derivation of this lower bound is omitted because it is almost identical with the sequential case (see Appendix~\ref{sec:IBO}).
% This lower bound can be obtained as with the sequential case (see Appendix~\ref{sec:IBO}).
%
% (k1020:これはsequentialの場合と同じ導出でよいという意味？そうなら説明が必要)
% $Z_{\*x^{(q)}}(f_*) = \Pr (\*h_{\*x} \in \cA | \*X_{q-1}, \*H_{q-1})$.
%
% Let $m^{(f)}_{q}$ and $m^{(g_c)}_{q}$ be the predictive means, and ${s^{(f)}_{q}}^2$ and ${s^{(g_c)}_{q}}^2$ be the predictive variances after conditioning by $\cH_{q-1}$, respectively.
%
Suppose that $m^{(f)}_{\*x^{(q)}}$ and ${s^{(f)}_{\*x^{(q)}}}^2$ are the predictive mean and variance of $f(\*x)$ after conditioning by $\cH_{q-1}$, respectively, and that $m^{(g_c)}_{\*x^{(q)}}$ and ${s^{(g_c)}_{\*x^{(q)}}}^2$ are those for $g_c(\*x)$.
Then, from the independence assumption,
$Z_{\*x^{(q)}}(f_* | \cX_{q-1}) = \bigl(1 - \Phi(\eta^{(f)}_{\*x^{(q)}}(f_*))\bigl) \prod_{c=1}^C \bigl(1 - \Phi(\eta^{(g_c)}_{\*x^{(q)}}(z_c))\bigl)$, where
$\eta^{(f)}_{\*x^{(q)}}(f_*) = \bigl(f_* - m^{(f)}_{\*x^{(q)}} \bigl) / s^{(f)}_{\*x^{(q)}}$
and
$\eta^{(g_c)}_{\*x^{(q)}}(z_c) = \bigl(z_c - m^{(g_c)}_{\*x^{(q)}} \bigl) / s^{(g_c)}_{\*x^{(q)}},\ c=1, \dots, C$.
By applying the Monte Carlo estimation to the expectation in \eqref{eq:PCMES-IBO}, we obtain the acquisition function for the $q$-th query as
\begin{align}
 % &\alpha^{\rm IBO}_{t}(\*x | \*X_{q-1}) = - \frac{1}{K} \sum_{(f_*, \*H_{q-1}) \in \cJ} \log(1-Z_{\*x^{(q)}}(f_*)), \label{eq:PCMES-IBO_final}
 &\alpha^{\rm IBO} (\*x^{(q)} | \cX_{q-1}) = - \frac{1}{K} \sum_{(\tilde{f_*}, \cH_{q-1}) \in \cJ} \log(1-Z_{\*x^{(q)}}(\tilde{f_*} | \cX_{q-1})), \tag{\ref{eq:PCMES-IBO_final}}
%  \label{eq:PCMES-IBO_final}
\end{align}
where $\cJ$ is a set of $K$ sampled
$(\tilde{f_*}, \cH_{q-1})$
% $(\tilde{f_*}, \*H_{q-1})$
from the current GPs.

%%%%%%%%%%%%%%%%%%%%%%%%%%%%%%%%%%%%%%%%%%%%%%%%%%%%%%%%%%%%%%%%%%%%%%%%%%%%%%%%%%%%%%%%%%%%%%
\section{Proof of Theorem~\ref{theo:CMES-IBO}}
\label{sec:proof-ibo-var}
%%%%%%%%%%%%%%%%%%%%%%%%%%%%%%%%%%%%%%%%%%%%%%%%%%%%%%%%%%%%%%%%%%%%%%%%%%%%%%%%%%%%%%%%%%%%%%

Let $f_{*(1)}, \dots, f_{*(K)}$ be i.i.d. random variables sampled from the distribution of $f_*$, and let $D_k \coloneqq - \log  \bigl(1 - \Pr(\*h_{\*x} \in \cA_{f_*(k)}) \bigl)$, where $\cA_{f_*(k)} \coloneqq (f_{*(k)}, \infty) \times (z_1, \infty) \times \dots \times (z_C, \infty)$ for $\forall k {\in \{1, \ldots, K\}}$.
% \red{(takeno:$U_k$は使ってしまったので, $D_k$に変更)}
%
Then, our acquisition function can be written as
\begin{align*}
    \alpha^{\rm IBO}(\*x) = \frac{1}{K} \sum_{k=1}^K D_k.
\end{align*}
%

% \blue{We first show the variance of $D_k$ is equal to or less than $2$, i.e., $\VV_{{f_{*(k)}}} [ D_k ] \leq 2$.}
%
For any $b \in  \RR \cup \{-\infty\}$, the following inequality holds:
% The following inequality holds:
\begin{align*}
    \Pr(\*h_{\*x} \in \cB) \leq \Pr(f_* > b),
\end{align*}
% where $b \in \RR$ and $\cB \coloneqq (b, \infty) \times (z_1, \infty) \times \dots \times (z_C, \infty)$.
where $\cB \coloneqq (b, \infty) \times (z_1, \infty) \times \dots \times (z_C, \infty)$.
%
% This is because $f_* \geq f(\*x) \geq b$ from the definition if $\*h_{\*x} \in \cB$, i.e., $\*x$ is feasible and $f(\*x) \geq b$.
This is because $\*h_{\*x} \in \cB$ implies that (i) $\cX_{\rm feasible}$ is not empty since at least $\*x$ is feasible and (ii) $f_* = \max_{\*x^\prime \in \cX_{\rm feasible}} f(\*x^\prime) \geq f(\*x) > b$.
% } \blue{(takeno:この1文に対する質問だったと思うので, もう少し詳細にしてみました.)}
% \blue{
% This is because $f_* \geq f(\*x) \geq b$ if $\*h_{\*x} \in \cB$, which means $\*x$ is feasible and $f(\*x) \geq b$.
% }
%
% We can use sampled $f_{*(1)}, \dots, f_{*(K)}$ instead of $b$, and we see
By substituting the sampled $f_{*(1)}, \dots, f_{*(K)}$ into $b$, we see
\begin{align*}
    \Pr(\*h_{\*x} \in \cA_{f_*(k)}) \leq \Pr(f_* > f_{*(k)}),
\end{align*}
and
\begin{align}
    1 - \Pr(\*h_{\*x} \in \cA_{f_*(k)}) \geq  1 - \Pr(f_* > f_{*(k)})  =  F_*( f_{*(k)}),
    \label{eq:ineq_first}
\end{align}
where $F_*(b) \coloneqq \Pr(f_* \leq b)$ is CDF of $f_*$.

Let us consider the distribution of $F_* (f_{*(k)})$ induced by $f_{*(k)}$.
Define
$\tau \coloneqq \Pr(f_* = -\infty)$.
Then, $f_{*(k)} = -\infty$ with probability $\tau$ because $f_{*(k)}$ is a random sample of $f_*$.
This implies $F_* (f_{*(k)}) = \Pr(f_* = -\infty) = \tau$ with probability $\tau$.
%
%
% From the definition, $F_* (f_{*(k)}) = \tau \coloneqq \Pr(f_* = -\infty)$ with probability $\tau$.
% 定義より, $\tau \coloneqq \Pr(f_* = -\infty)$の確率で$\tau$を取る.
%
% Moreover, it has the continuous CDF for $\alpha \in (\tau, 1)$.
Moreover, the CDF of $F_* (f_{*(k)})$, defined as $\Pr( F_* (f_{*(k)}) \leq \alpha )$, is continuous for $\alpha \in (\tau, 1)$.
% また, $\alpha \in (\tau, 1)$については, 連続なCDFを持つ.
%
% Thus, as with the probability integral transform (e.g., see \citep{Casella2002-Statistical}), CDF of $F_* (f_{*(k)})$ for $\alpha$ can be obtained as
Thus, as with the probability integral transform (e.g., see \citep{Casella2002-Statistical}), $\Pr( F_* (f_{*(k)}) \leq \alpha )$ can be obtained as
\begin{align*}
    \Pr( F_* (f_{*(k)}) \leq \alpha ) &= \Pr\bigl( F_*^{-1} \bigl( F_* (f_{*(k)}) \bigr) \leq  F_*^{-1} (\alpha) \bigr) \\
    &= \Pr\bigl( f_{*(k)} \leq  F_*^{-1} (\alpha) \bigr) \\
    &= F_* \bigl( F_*^{-1} (\alpha) \bigr) \\
    &= \alpha,
\end{align*}
where $F_*^{-1}(\cdot)$ is the quantile function of $f_*$.
Consequently, CDF of $F_* (f_{*(k)})$ is obtained as,
\begin{align*}
    \Pr( F_* (f_{*(k)}) \leq \alpha ) &= \begin{cases}
        0 & \text{if } \alpha < \tau, \\
        % \tau & \text{if } \alpha = \tau, \\
        \alpha & \text{if } \alpha \geq \tau.
    \end{cases}
    % \label{eq:CDF_CDF_f_star}
\end{align*}
From above, $F_* (f_{*(k)}) = \tau$ with probability $\tau$, and $F_* (f_{*(k)}) \sim {\rm Unif}(\tau, 1)$ with probability $1 - \tau$.
% 以上より, $F_* (f_{*(k)})$は確率$\tau$で$\tau$を取り, 確率$1 - \tau$で区間$[\tau, 1]$の一様分布に従う.
%
Then, we define the random variable $U_k$ as follows:
% そこで, 以下の確率変数$U_k$を考える:
\begin{align*}
    U_k &=
    \begin{cases}
        \tilde{U}_k & \text{if } f_{*(k)} = -\infty, \\
        F_* (f_{*(k)}) & \text{if } f_{*(k)} \neq -\infty,
    \end{cases}
\end{align*}
where $\tilde{U}_k \sim {\rm Unif}(0, \tau)$.
In this definition, we have
% This
$U_k \sim {\rm Unif}(0, \tau)$ with probability $\tau$, and $U_k \sim {\rm Unif}(\tau, 1)$ with probability $1 - \tau$.
% この$U_k$は確率$\tau$で区間$[0, \tau]$の一様確率変数を取り, 確率$1 - \tau$で区間$[\tau, 1]$の一様分布に従う.
%
Therefore, $U_k \sim {\rm Unif}(0, 1)$.
% したがって, $U_k \sim {\rm Unif}(0, 1)$.
%
Since $F_* (f_{*(k)}) \geq U_k$ from the definition, by combining \eqref{eq:ineq_first}, we obtain
% また, 定義から明らかに$F_* (f_{*(k)}) \geq U_k$であり, \eqref{eq:ineq_first}とまとめれば,
\begin{align}
    1 - \Pr(\*h_{\*x} \in \cA_{f_*(k)}) \geq F_*( f_{*(k)}) \geq U_k.
    % \label{eq:ineq_second}
\end{align}

From this inequality and the monotonicity of logarithm, we can transform
\begin{align*}
    D_k = - \log \bigl(1 - \Pr(\*h_{\*x} \in \cA_{f_*(k)}) \bigl)
    \leq  - \log \bigl( F_*( f_{*(k)}) \bigl)
    \leq - \log ( U_k )
    \eqqcolon M_k.
\end{align*}
$D_k$ and $M_k$ are nonnegative random variables, and $\Pr(D_k \leq M_k) = 1$.
%  \blue{almost surely}.
% $D_k \leq M_k$ \blue{almost surely}.
%
Hence, we see that $0 \leq \EE_{{\cF_*}}[D_k] \leq \EE_{{\cF_*}}[M_k]$, $\EE_{{\cF_*}}[D_k^2] \leq \EE_{{\cF_*}}[M_k^2]$, and $\Pr(D_k \geq b) \leq \Pr(M_k \geq b)$ for $\forall b \in \RR$.

% \red{Then, we reveal the properties of $D_k$ from that of $M_k$.}
We will show that variance and concentration bounds for $D_k$ can be derived from that of $M_k$.
%
% \blue{Since $f_*$ is a continuous random variable, we can apply the probability integral transform and its inverse transformation.
%
% Thus, the CDF $\Pr(f_* \leq f_{*(k)})$ follows a standard uniform distribution and its transformation by $- \log(\cdot)$ follows an exponential distribution with the rate parameter $\lambda = 1$:
% \begin{align*}
%     \Pr(f_* \leq f_{*(k)}) &\sim {\rm Unif}(0, 1), \\
%     M_k = - \log \bigl( \Pr(f_* \leq f_{*(k)}) \bigl) &\sim {\rm exp}(\lambda = 1).
% \end{align*}
% }
%
Using the inverse probability integral transform \citep{Casella2002-Statistical}, $- \log(U_k)$ follows an exponential distribution with the rate parameter $\lambda = 1$:
\begin{align*}
    M_k = - \log ( U_k ) &\sim {\rm exp}(\lambda = 1).
\end{align*}
Therefore, $\EE_{{f_{*(k)}}}[M_k^2] = 2$, and we can derive the inequality below:
\begin{align*}
    \VV_{{f_{*(k)}}} [ D_k ] &= \EE_{{f_{*(k)}}}[D_k^2] - \EE_{{f_{*(k)}}}[D_k]^2 \\
    &\leq \EE_{{f_{*(k)}}}[D_k^2] \\
    &\leq \EE_{{f_{*(k)}}}[M_k^2] \\
    &= 2.
\end{align*}
Moreover, we see that $D_k$ satisfies the condition being a {\it sub-exponential random variable} \citep[Proposition~2.7.1 (a) in][]{Vershynin2018-high}, that is,
\begin{align*}
    \Pr(|D_k| \geq \xi) = \Pr(D_k \geq \xi) \leq \Pr(M_k \geq \xi) = \exp(-\xi) \text{ for all $\xi \geq 0$}.
\end{align*}
Consequently, $D_k$ is a sub-exponential random variable, whose variance is bounded from above $\VV_{{f_{*(k)}}} [ D_k ] \leq 2$.

We can directly apply Chebyshev's and Bernstein's inequalities from the bound of variance and the fact that $D_k$ is a sub-exponential random variable.
% We can directly apply Chebyshev's and Bernstein's inequalities from the bound of variance and that $D_k$ is the sub-exponential random variable, respectively.
%
First, applying Chebyshev's inequality, we can see immediately
\begin{align}
    \Pr( | \alpha^{\rm IBO}(\*x) - L(\*x) | \geq \xi ) \leq \frac{ 2 }{K\xi^2},
    \label{eq:Chebyshev}
\end{align}
for any $\xi \geq 0$.
Second, applying Bernstein's inequality \citep[Corollary 2.8.3 in ][]{Vershynin2018-high}, we can derive
\begin{align}
    \Pr( | \alpha^{\rm IBO}(\*x) - L(\*x) | \geq \xi ) \leq 2 \exp \biggl[ - A K \min \biggl( \frac{\xi^2}{B^2}, \frac{\xi}{B} \biggr)  \biggr],
    \label{eq:Bernstein}
\end{align}
where $A$ is a constant
% , which \red{does not change depending on the sub-exponential random variable}
%
and $B$ can be provided as a {\it sub-exponential norm}
\footnote{
The sub-exponential norm for a random variable $X$ is defined as follows:
% \begin{align*}
    $
    \| X \|_{\psi_1} \coloneqq \inf \bigl\{ \xi > 0 \mid \EE \bigl[ \exp \bigl( |X| / \xi \bigr) \bigr] \leq 2 \bigr\}.
    $
% \end{align*}
}
\citep[Definition 2.7.5 in ][]{Vershynin2018-high} of $D_k - \EE_{f_{*(k)}}[D_k]$.
For all $t \geq 0$ and $\*x \in \cX$, the sub-exponential norm $\| D_k  - \EE_{f_{*(k)}}[D_k] \|_{\psi_1}$ can be bounded as
\begin{align*}
    \| D_k  - \EE_{f_{*(k)}}[D_k] \|_{\psi_1} &\leq \| D_k \|_{\psi_1} + \| \EE_{f_{*(k)}}[D_k] \|_{\psi_1} \tag{triangle inequality} \\
    &\leq \| M_k \|_{\psi_1} + \| \EE_{f_{*(k)}}[M_k] \|_{\psi_1}  \tag{$\Pr(D_k \leq M_k) = 1$}\\
    &= \| M_k \|_{\psi_1} + \| 1 \|_{\psi_1}  \tag{$\EE_{f_{*(k)}}[M_k] = 1$} \\
    &= 2 + 1 / \log 2. \tag{$M_k \sim \exp (\lambda = 1)$}
\end{align*}
Therefore, $B$ can be set as $2 + 1 / \log 2$.
Consequently, we can derive Theorem~\ref{theo:CMES-IBO} combining these two inequalities \eqref{eq:Chebyshev} and \eqref{eq:Bernstein}.

% Above inequalities holds even after conditioning by $\cH_q$.
All the above derivations hold even after conditioning by $\cH_q$ for the parallel setting.
Let $D_k^{(q)} \coloneqq - \log \bigl( 1 - \Pr(\*h_{\*x} \in \cA_{f_*(k)} | \cH_{q-1})\bigl)$ and $\alpha^{\rm IBO}(\*x | \cX_{q-1}) = \nicefrac{1}{K} \sum_{k=1}^K D_k^{(q)}$.
Then, using $U_k^{(q)} \sim {\rm Unif}(0, 1)$, we can show that $D_k^{(q)} \leq M_k^{(q)} \coloneqq - \log(U_k^{(q)})$ from the same derivation.
Since $M_k^{(q)}$ obviously follows the exponential distribution, we see that $\VV_{{\cJ}}[D_k^{(q)}] \leq 2$ and $D_k^{(q)}$ is also still a sub-exponential random variable.
% }
%
The subsequent inequalities can be derived in the same way.

%%%%%%%%%%%%%%%%%%%%%%%%%%%%%%%%%%%%%%%%%%%%%%%%%%%%%%%%%%%%%%%%%%%%%%%%%%%%%%%%%%%%%%%%%%%%%%
\section{Discussion on Infeasibility}
\label{sec:infeasibility}
%%%%%%%%%%%%%%%%%%%%%%%%%%%%%%%%%%%%%%%%%%%%%%%%%%%%%%%%%%%%%%%%%%%%%%%%%%%%%%%%%%%%%%%%%%%%%%

Here, we discuss the case that sample paths generated for sampling $f_*$ do not have any feasible solution.
\citet{Perrone2019-Constrained} defined $f_*$ as $\max f(\*x),\ {\rm s.t.}\ g_c(\*x) \geq z_c$ for $\forall c = 1, \dots, C$, by which the infeasible case is not considered.
%
% We emphasize that \red{GPs usually can generate an infeasible sample path,} i.e.,$\Pr(\cX_{\rm feasible} = \emptyset) > 0$(in fact, this is always the case unless we observe a noiseless feasible solution).
However, particularly when the dataset $\cD_{t-1}$ does not contain any feasible solution (which typically occurs at the beginning of the optimization), it often occurs that a sample path generated for sampling $f_*$ does not have any feasible solution.
An illustrative example is shown in Figure~\ref{fig:truncation}~(a), in which the feasible region of the green sample path is empty.
In fact, GPs can generate an infeasible sample path, i.e., $\Pr(\cX_{\rm feasible} = \emptyset) > 0$ unless we observe a noiseless feasible solution.
Then, there is a possibility that $f_*$ is not defined, by which MI even cannot be theoretically defined.

Other than using our definition of $f_*$ \eqref{eq:f_star}, another possible approach to avoiding this problem is to assume the existence of feasible solutions in the GPs,
but this approach has at least the following three disadvantages.
% but it causes at least the following two computational difficulties.
%
First, this approach ignores uncertainty about the existence of a feasible solution.
Identifying the existence of a feasible solution for a given problem is a key issue in constrained problems (called the feasibility problem \citep{Chinneck2007-feasibility}).
However, information gain about feasibility cannot be incorporated in this approach.
% First,
Second, we need to generate sample paths that have feasible solutions for sampling $f_*$.
A na\"ive approach is the rejection sampling, in which generated constraint functions are rejected if no feasible solution exists, but this may require a huge number of samplings.
%
% Second,
Third, the predictive distribution of $\*h_{\*x}$ under this condition
$p(\*h_{\*x} \mid \text{at least one feasible $\*x$ exits})$
is not a Gaussian distribution anymore, and is analytically intractable. % , \red{though the existing studies ignore this issue (simply use $p(\*h_{\*x})$ without any justification)}.
This makes the entire computational procedures of both CMES-IBO and CMES much more complicated.
\citet{Perrone2019-Constrained} did not mention the above issues at all and used the usual predictive distributions of GPs in the entire acquisition function evaluation without any justification.
Note that the same problem exists in the PES-based CBO \citep{Hernandez-Lobato2015-Predictive, Hernandez-Lobato2016-General}.
On the other hand, our definition of $f_*$ \eqref{eq:f_star} resolves the above issues.

We further discuss how CMES-IBO balances the effect of the objective and constraint functions when infeasible sample paths are generated.
For simplicity, we first consider the case of $K = 1$.
The acquisition function \eqref{eq:CMES-IBO} can be written as
$-\log( 1 -  P_{\rm imp} \times P_{\rm fea})$, where
$P_{\rm imp} \coloneqq \Pr(f(\*x) \geq \tilde{f}_*)$
and
$P_{\rm fea} \coloneqq {\prod_{c = 1}^C \Pr(g_{c}(\*x) \geq z_c)}$.
In this criterion, $P_{\rm imp}$ and $P_{\rm fea}$ represent benefits for obtaining a larger value of $f(\*x)$ and for obtaining a feasible solution, respectively.
If the sample path does not have a feasible region, $P_{\rm imp} = \Pr(f(\*x) \geq -\infty) = 1$.
Thus, only the probability of being feasible $P_{\rm fea}$ is maximized.
On the other hand, when $K > 1$, the balance of the effect of the objective and the constraint functions are balanced through the frequency that sampled constraint functions have a feasible region among $K$ samplings.
%
% Note that when $K > 1$, the setting $\zeta \rightarrow \infty$ still can incorporate the effect of the objective function through the frequency that sampled constraint functions have a feasible region.
%
If the frequency is low, the probability of being feasible has a dominant effect, while if the frequency is high, the effect of the objective function becomes strong.
Note that sampled constraint functions can have feasible solutions even when the dataset $\cD_{t-1}$ does not contain any feasible solution (as illustrated in Figure~\ref{fig:truncation}~(a)).

%%%%%%%%%%%%%%%%%%%%%%%%%%%%%%%%%%%%%%%%%%%%%%%%%%%%%%%%%%%%%%%%%%%%%%%%%%%%%%%%%%%%%%%%%%%%%%
\section{Constrained Extension of Conventional Max-value Entropy Search}
\label{sec:CMES}
%%%%%%%%%%%%%%%%%%%%%%%%%%%%%%%%%%%%%%%%%%%%%%%%%%%%%%%%%%%%%%%%%%%%%%%%%%%%%%%%%%%%%%%%%%%%%%
We here extend cMES \citep{Perrone2019-Constrained} to the multiple constraints setting and the parallel setting, and further, we show an example that its approximate MI can be negative.
%
% \blue{Thus, we extend MES to multiple constraints and show that it is still reduced to a simple closed-form except for one-dimensional Monte Carlo integration.(takeno:このあたり適当に書き換え)}

\subsection{Acquisition Function of CMES}
Following the conventional MES and cMES \citep{Wang2017-Max, Perrone2019-Constrained}, we approximate MI as below:
\begin{align}
 \eqref{eq:MI}
 &= H(\*h_{\*x} ) - \mathbb{E}_{f_*}\bigl[ H(\*h_{\*x} \mid f_*) \bigl] \nonumber \\ % \label{eq:CMES_diff_entropy} \\
 &\approx  H(\*h_{\*x} )  - \mathbb{E}_{f_*}\bigl[ H(\*h_{\*x} \mid \*h_{\*x} \in \overline{\cA}_{f_*}) \bigl].
  \tag{\ref{eq:CMES_approx_truncated}}
\end{align}
%
% \eqref{eq:CMES_diff_entropy} is derived by a well-known relation in information theory.
%
% In this equation, the first term is easy to calculate because it is the entropy of the multivariate normal distribution.
% %
% However, the second term is difficult to evaluate directly.
%
For the approximation, the conditioning on $f_*$ in the second term is replaced with $\*h_{\*x} \in \overline{\cA}_{f_*}$ following the existing approach \citep{Perrone2019-Constrained}.
Further, the expectation over $p(f_*)$ in the second term of \eqref{eq:CMES_approx_truncated} is also analytically intractable.
This expectation can be approximated by the MC estimation, for which we employ the same RFF-based approach described in Appendix~\ref{sec:sampling_fstar}.
Thus, we obtain the following acquisition function as an approximation of \eqref{eq:CMES_approx_truncated}:
\begin{align}
 \begin{split}
  \alpha^{\rm CMES}(\*x) \coloneqq
  & H(\*h_{\*x}) - \frac{1}{K} \! \sum_{\tilde{f}_* \in \cF_*} \! H(\*h_{\*x} \mid \*h_{\*x} \in \overline{\cA}_{\tilde{f}_*}).
 \end{split}
 \label{eq:CMES_approx}
\end{align}
The first term is easy to calculate because it is the entropy of the multivariate normal distribution.
Hence, we need to calculate the entropy in the second term $H(\*h_{\*x} \mid \*h_{\*x} \in \overline{\cA}_{\tilde{f}_*})$.

Although \citet{Perrone2019-Constrained} derived a closed-form of the entropy of this TMN distribution specific for $C = 1$, for $C > 1$, directly evaluating it is not trivial as pointed out by \citep{Fernandezsanchez2020-maxvalue}.
We show that another analytical representation of this entropy actually can be derived for general $C$ by transforming the domain of the integration as follows:
\begin{align}
    &H(\*h_{\*x} \mid \*h_{\*x} \in \overline{\cA}_{\tilde{f}_*})
    =
    % \int_{\overline{\cA}_{\tilde{f}_*}} - \frac{p(\*h_{\*x} \mid \cD_{t-1})}{\overline{Z}_{\*x}(\tilde{f}_*)} \log \frac{p(\*h_{\*x} \mid \cD_{t-1})}{\overline{Z}_{\*x}(\tilde{f}_*)} \mathrm{d}\*h_{\*x} \\
    \int_{\overline{\cA}_{\tilde{f}_*}} - \frac{p(\*h_{\*x}) }{\overline{Z}_{\*x}(\tilde{f}_*)} \log \frac{p(\*h_{\*x}) }{\overline{Z}_{\*x}(\tilde{f}_*)} \mathrm{d}\*h_{\*x} \nonumber \\
    &= \int - \frac{p(\*h_{\*x}) }{\overline{Z}_{\*x}(\tilde{f}_*)} \log \frac{p(\*h_{\*x}) }{\overline{Z}_{\*x}(\tilde{f}_*)} \mathrm{d}\*h_{\*x}
     + \int_{\cA_{\tilde{f}_*}} \frac{p(\*h_{\*x}) }{\overline{Z}_{\*x}(\tilde{f}_*)} \log \frac{p(\*h_{\*x}) }{\overline{Z}_{\*x}(\tilde{f}_*)} \mathrm{d}\*h_{\*x} \nonumber \\
     &= \frac{1}{\overline{Z}_{\*x}(\tilde{f}_*)}  \int - p(\*h_{\*x}) \log p(\*h_{\*x}) \mathrm{d}\*h_{\*x}
     + \frac{1}{\overline{Z}_{\*x}(\tilde{f}_*)} \int_{\cA_{\tilde{f}_*}} p(\*h_{\*x}) \log p(\*h_{\*x}) \mathrm{d}\*h_{\*x} + \log(\overline{Z}_{\*x}(\tilde{f}_*))  \nonumber \\
     &= \frac{1}{\overline{Z}_{\*x}(\tilde{f}_*)} \int - p(\*h_{\*x}) \log p(\*h_{\*x}) \mathrm{d}\*h_{\*x}
     + \frac{Z_{\*x}(\tilde{f}_*)}{\overline{Z}_{\*x}(\tilde{f}_*)} \int_{\cA_{\tilde{f}_*}} \frac{p(\*h_{\*x}) }{Z_{\*x}(\tilde{f}_*)} \log \frac{p(\*h_{\*x}) }{Z_{\*x}(\tilde{f}_*)} \mathrm{d}\*h_{\*x} + \log(\overline{Z}_{\*x}(\tilde{f}_*)) + \frac{Z_{\*x}(\tilde{f}_*) \log (Z_{\*x}(\tilde{f}_*))}{\overline{Z}_{\*x}(\tilde{f}_*)} \nonumber \\
    %  &= \frac{1}{\overline{Z}_{\*x}(\tilde{f}_*)} \int - p(\*h_{\*x}) \log p(\*h_{\*x}) \mathrm{d}\*h_{\*x}
    %  + \frac{Z_{\*x}(\tilde{f}_*)}{\overline{Z}_{\*x}(\tilde{f}_*)} \int_{\cA_{\tilde{f}_*}} \frac{p(\*h_{\*x}) }{Z_{\*x}(\tilde{f}_*)} \log \frac{p(\*h_{\*x}) }{Z_{\*x}(\tilde{f}_*)} \mathrm{d}\*h_{\*x}
    %  + \frac{\log(\overline{Z}_{\*x}(\tilde{f}_*)) }{\overline{Z}_{\*x}(\tilde{f}_*)} - \frac{Z_{\*x}(\tilde{f}_*)\log(\overline{Z}_{\*x}(\tilde{f}_*))}{\overline{Z}_{\*x}(\tilde{f}_*)} + \frac{Z_{\*x}(\tilde{f}_*) \log (Z_{\*x}(\tilde{f}_*))}{\overline{Z}_{\*x}(\tilde{f}_*)}  \\
    %  &= \frac{1}{\overline{Z}_{\*x}(\tilde{f}_*)} \int - p(\*h_{\*x}) \log p(\*h_{\*x}) \mathrm{d}\*h_{\*x}
    %  - \frac{Z_{\*x}(\tilde{f}_*)}{\overline{Z}_{\*x}(\tilde{f}_*)} \int_{\cA_{\tilde{f}_*}} - \frac{p(\*h_{\*x}) }{Z_{\*x}(\tilde{f}_*)} \log \frac{p(\*h_{\*x}) }{Z_{\*x}(\tilde{f}_*)} \mathrm{d}\*h_{\*x}
    %  + \log(\overline{Z}_{\*x}(\tilde{f}_*)) + \frac{Z_{\*x}(\tilde{f}_*) \log (Z_{\*x}(\tilde{f}_*))}{\overline{Z}_{\*x}(\tilde{f}_*)} \\
     &= \frac{H(\*h_{\*x}) }{\overline{Z}_{\*x}(\tilde{f}_*)} - \frac{Z_{\*x}(\tilde{f}_*)}{\overline{Z}_{\*x}(\tilde{f}_*)} \int_{\cA_{\tilde{f}_*}} - p(\*h_{\*x} | \*h_{\*x} \in \cA_{\tilde{f}_*}) \log p(\*h_{\*x} | \*h_{\*x} \in \cA_{\tilde{f}_*}) \mathrm{d}\*h_{\*x}
     + \log(\overline{Z}_{\*x}(\tilde{f}_*)) + \frac{Z_{\*x}(\tilde{f}_*) \log (Z_{\*x}(\tilde{f}_*))}{\overline{Z}_{\*x}(\tilde{f}_*)} \nonumber \\
     &= \frac{H(\*h_{\*x}) }{\overline{Z}_{\*x}(\tilde{f}_*)} - \frac{Z_{\*x}(\tilde{f}_*) H(\*h_{\*x} \mid \*h_{\*x} \in \cA_{\tilde{f}_*})}{\overline{Z}_{\*x}(\tilde{f}_*)} + \log(\overline{Z}_{\*x}(\tilde{f}_*)) + \frac{Z_{\*x}(\tilde{f}_*) \log Z_{\*x}(\tilde{f}_*)}{\overline{Z}_{\*x}(\tilde{f}_*)}, \label{eq:second_term_entropy}
\end{align}
where TMN $p(\*h_{\*x} | \*h_{\*x} \in \cA_{\tilde{f}_*})$ is defined as
\begin{align*}
    p(\*h_{\*x} | \*h_{\*x} \in \cA_{\tilde{f}_*}) = \begin{cases}
        p(\*h_{\*x}) / Z_{\*x}(\tilde{f}_*) & {\rm if}\ \*h_{\*x} \in \cA_{\tilde{f}_*} \\
        0 & {\rm otherwise}
    \end{cases}.
\end{align*}
% $p(\*h_{\*x}) / Z$ if $\*h_{\*x} \in \cA_{\tilde{f}_*}$ and $0$ for otherwise.
%
% \begin{align}
%  H(\*h_{\*x} \mid \*h_{\*x} \in \overline{\cA}_{\tilde{f}_*})
%  &=
%  % \int_{\mathbb{R}^{C+1} \backslash \cA_{\tilde{f}_*}} - \frac{p(\*h_{\*x} \mid \cD_{t-1})}{1-Z} \log \frac{p(\*h_{\*x} \mid \cD_{t-1})}{1-Z} \mathrm{d}\*h_{\*x} \nonumber \\
%  \int_{\overline{\cA}_{\tilde{f}_*}} - \frac{p(\*h_{\*x}) }{1-Z} \log \frac{p(\*h_{\*x}) }{1-Z} \mathrm{d}\*h_{\*x} \nonumber \\
%  &= \int - \frac{p(\*h_{\*x}) }{1-Z} \log \frac{p(\*h_{\*x}) }{1-Z} \mathrm{d}\*h_{\*x}
%   + \int_{\cA_{\tilde{f}_*}} \frac{p(\*h_{\*x}) }{1-Z} \log \frac{p(\*h_{\*x}) }{1-Z} \mathrm{d}\*h_{\*x} \nonumber \\
%  %
%  % &= \frac{H(\*h_{\*x} }{1-Z} + \frac{\log(1-Z)}{1-Z} - \frac{Z\log(1-Z)}{1-Z} - \int_{\cA_{\tilde{f}_*}} - \frac{p(\*h_{\*x} }{1-Z} \log p(\*h_{\*x}  \mathrm{d}\*h_{\*x} \nonumber  \\
%  % &= \frac{H(\*h_{\*x} }{1-Z} - \frac{Z}{1-Z} \int_{\cA_{\tilde{f}_*}} - \frac{p(\*h_{\*x} }{Z} \log \frac{p(\*h_{\*x} }{Z} \mathrm{d}\*h_{\*x} + \log(1-Z) + \frac{Z \log Z}{1-Z}  \nonumber \\
%  &= \frac{H(\*h_{\*x}) }{1-Z} - \frac{Z H(\*h_{\*x} \mid \*h_{\*x} \in \cA_{\tilde{f}_*})}{1-Z} + \log(1-Z) + \frac{Z \log Z}{1-Z}.
%  \label{eq:second_term_entropy}
% \end{align}
%
In the last equation, $Z_{\*x}(\tilde{f}_*), \overline{Z}_{\*x}(\tilde{f}_*)$ and $H(\*h_{\*x})$ are easy to compute.
Although
$H(\*h_{\*x} \mid \*h_{\*x} \in \cA_{\tilde{f}_*})$
in the second term is still the entropy of TMN, we show the calculation of this entropy in Appendix~\ref{sec:derivation_entropy_tmvn} for both independent and correlated cases.

Consequently, in the independent case, by substituting \eqref{eq:second_term_entropy} into \eqref{eq:CMES_approx} and replacing $H(\*h_{\*x})$ and $H(\*h_{\*x} \mid \*h_{\*x} \in \cA_{\tilde{f}_*})$ with their closed-forms (shown in Appendix~\ref{sec:derivation_entropy_tmvn}), we obtain
% replacing each entropy with its closed-form \eqref{eq:entropy_in_A} (shown in Appendix~\ref{sec:derivation_entropy_tmvn}), we can obtain
\begin{align*}
 % \eqref{eq:CMES_approx} &= \frac{1}{|\cF_*|} \sum_{\tilde{f}_* \in \cF_*} - \frac{Z}{1-Z} H(\*h_{\*x} \mid \cD_{t-1}) + \frac{Z}{1-Z} H(\*h_{\*x} \mid \*h_{\*x} \geq \*l, \cD_{t-1}) - \log(1-Z) - \frac{Z \log Z}{1-Z} \nonumber \\
    % &= \frac{1}{|\cF_*|} \sum_{\tilde{f}_* \in \cF_*} \frac{Z}{2(1-Z)} \biggl( \frac{\gamma^{(f)}_{\*x}(\tilde{f}_*) \phi(\gamma^{(f)}_{\*x}(\tilde{f}_*))}{1 - \Phi(\gamma^{(f)}_{\*x}(\tilde{f}_*))} + \sum_{c=1}^C \frac{\gamma^{(g_c)}_{\*x}(z_c) \phi(\gamma^{(g_c)}_{\*x}(z_c))}{1 - \Phi(\gamma^{(g_c)}_{\*x}(z_c))} \biggl) - \log(1-Z),
 \alpha^{\rm CMES}&(\*x) = \frac{1}{K} \sum_{\tilde{f}_* \in \cF_*} \frac{Z_{\*x}(\tilde{f}_*)}{2(1-Z_{\*x}(\tilde{f}_*))} R_{\tilde{f}_*} - \log(1-Z_{\*x}(\tilde{f}_*)), % \label{eq:CMES_final}
 %\\
\end{align*}
where
\begin{align*}
    R_{\tilde{f}_*} &\coloneqq \frac{\gamma^{(f)}_{\*x}(\tilde{f}_*) \phi(\gamma^{(f)}_{\*x}(\tilde{f}_*))}{1 - \Phi(\gamma^{(f)}_{\*x}(\tilde{f}_*))} + \sum_{c=1}^C \frac{\gamma^{(g_c)}_{\*x}(z_c) \phi(\gamma^{(g_c)}_{\*x}(z_c))}{1 - \Phi(\gamma^{(g_c)}_{\*x}(z_c))}. \nonumber
\end{align*}
In the correlated case, $H(\*h_{\*x})$ and $H(\*h_{\*x} \mid \*h_{\*x} \in \cA_{\tilde{f}_*})$ are also replaced with their closed-forms (shown in Appendix~\ref{sec:derivation_entropy_tmvn}), and we obtain
% by replacing each entropy with its closed-form \eqref{eq:entropy_in_A_correlated} similarly, we also derive
\begin{align}
    \alpha^{\rm CMES}&(\*x) = \frac{1}{K} \sum_{\tilde{f}_* \in \cF_*} \frac{Z_{\*x}(\tilde{f}_*)}{2(1-Z_{\*x}(\tilde{f}_*))} \biggl( {\rm Tr}\Bigl( \*\Sigma^{-1}(\*x) \bigl( \*\Sigma^{\rm TN}(\*x) + \*d \*d^\top \bigl) \Bigl) - C - 1 \biggl) - \log(1-Z_{\*x}(\tilde{f}_*)),
 \label{eq:CMES_correlated_final}
\end{align}
where ${\rm Tr} (\cdot)$ is the trace of a matrix, $\*\mu(\*x)$ and $\*\Sigma(\*x)$ are the expectation and covariance matrix of $\*h_{\*x}$, respectively, $\*\mu^{\rm TN}(\*x)$ and $\*\Sigma^{\rm TN}(\*x)$ are those for $\*h_{\*x} | \*h_{\*x} \in \cA_{\tilde{f}_*}$, and $\*d = \*\mu^{\rm TN}(\*x) - \*\mu(\*x)$.
% $ \*\mu(\*x), \*\Sigma(\*x)$ and $\*\mu^{\rm TN}(\*x), \*\Sigma^{\rm TN}(\*x)$ are the expectation and covariance matrix of $\*h_{\*x}$ and $\*h_{\*x} | \*h_{\*x} \in \cA_{\tilde{f}_*} $, respectively, and $\*d \coloneqq \*\mu^{\rm TN}(\*x) - \*\mu(\*x)$.
%
See Appendix~\ref{sec:derivation_entropy_tmvn} for details.

%%%%%%%%%%%%%%%%%%%%%%%%%%%%%%%%%%%%%%%%%%%%%%%%%%%%%%%%%%%%%%%%%%%%%%%%%%%%%%%%%%%%%%%%%%%%%%
\subsection{Parallelization of CMES}
\label{sec:parallel_CMES}
%%%%%%%%%%%%%%%%%%%%%%%%%%%%%%%%%%%%%%%%%%%%%%%%%%%%%%%%%%%%%%%%%%%%%%%%%%%%%%%%%%%%%%%%%%%%%%
We can consider the CMI maximization in the same way as CMES-IBO.
Suppose that we already select the $q-1$ queries, and we need to select the next $q$-th query.
The CMI can be approximated as
\begin{align*}
    \EE_{\cH_{q-1}}\bigl[ I(\*h_{\*x^{(q)}} ; f_* | \cH_{q-1})\bigl] &= \EE_{\cH_{q-1}}\bigl[ H(\*h_{\*x^{(q)}} | \cH_{q-1} )\bigl] - \EE_{\cH_{q-1}, f_*}\bigl[ H(\*h_{\*x^{(q)}} | f_*, \cH_{q-1})\bigl] \\
    &\approx \EE_{\cH_{q-1}}\bigl[ H(\*h_{\*x^{(q)}} | \cH_{q-1} )\bigl] - \EE_{\cH_{q-1}, f_*}\bigl[ H(\*h_{\*x^{(q)}} | \*h_{\*x^{(q)}} \in \overline{\cA}_{f_*}, \cH_{q-1})\bigl]
\end{align*}
As described in Appendix~\ref{sec:parallel-detail}, we define
$f(\*x) | \cD_{t-1}, \cX_{q-1}, \cH_{q-1} \sim \cN\bigl(m^{(f)}_{\*x^{(q)}}, {s^{(f)}_{\*x^{(q)}}}^2\bigl)$ and
$g_c(\*x) | \cD_{t-1}, \cX_{q-1}, \cH_{q-1} \sim \cN\bigl(m^{(g_c)}_{\*x^{(q)}}, {s^{(g_c)}_{\*x^{(q)}}}^2\bigl)\ c=1,\dots, C$.
Importantly, ${s^{(f)}_{\*x^{(q)}}}^2, {s^{(g_c)}_{\*x^{(q)}}}^2, \dots, {s^{(g_C)}_{\*x^{(q)}}}^2$ only depend on $\cX_{q-1}$, not $\cH_{q-1}$.
Thus, given $\cX_{q-1}$, the first term $\EE_{\cH_{q-1}}\bigl[ H(\*h_{\*x^{(q)}} | \cH_{q-1} )\bigl]$ can be calculated analytically.
The second term is also calculated in the same manner as CMES with given $\tilde{f}_*$ and $\cH_{q-1}$ :
\begin{align*}
    H(\*h_{\*x^{(q)}} \mid \*h_{\*x^{(q)}} \in \overline{\cA}_{\tilde{f}_*}, \cH_{q-1}) &= H(\*h_{\*x^{(q)}} | \cH_{q-1}) - \frac{Z_{\*x^{(q)}}(\tilde{f}_* | \cX_{q-1})}{2(1-Z_{\*x^{(q)}}(\tilde{f}_* | \cX_{q-1}))} R^{(q)}_{\tilde{f}_*} + \log(1 - Z_{\*x^{(q)}}(\tilde{f}_* | \cX_{q-1})), %\\
\end{align*}
where
\begin{align*}
    R^{(q)}_{\tilde{f}_*} &\coloneqq \frac{\eta^{(f)}_{\*x^{(q)}}(\tilde{f}_*)\phi(\eta^{(f)}_{\*x^{(q)}}(\tilde{f}_*))}{1 - \Phi(\eta^{(f)}_{\*x^{(q)}}(\tilde{f}_*))}
    + \sum_{c=1}^C  \frac{\eta^{(g_c)}_{\*x^{(q)}}(z_c)\phi(\eta^{(g_c)}_{\*x^{(q)}}(z_c))}{1 - \Phi(\eta^{(g_c)}_{\*x^{(q)}}(z_c))}. \nonumber
\end{align*}
Finally, by applying the MC approximation, we obtain the acquisition function below
\begin{align}
 \alpha^{\rm CMES}(\*x^{(q)} | \cX_{q-1}) \! &= \! \frac{1}{|\cJ|} \! \sum_{(\tilde{f}_*, \cH_{q-1}) \in \cJ} \! \frac{Z_{\*x^{(q)}}(\tilde{f}_* | \cX_{q-1})}{2(1-Z_{\*x^{(q)}}(\tilde{f}_* | \cX_{q-1}))} R^{(q)}_{\tilde{f}_*} - \log(1 - Z_{\*x^{(q)}}(\tilde{f}_* | \cX_{q-1})).
 \label{eq:PCMES_final}
\end{align}
We can also derive the acquisition function in the correlated setting, but we omit the detailed derivation.

%%%%%%%%%%%%%%%%%%%%%%%%%%%%%%%%%%%%%%%%%%%%%%%%%%%%%%%%%%%%%%%%%%%%%%%%%%%%%%%%%%%%%%%%%%%%%%
% \subsection{Proof of Lemma~\ref{lem:CMES-negativity}}
\subsection{Interpretation and Proof of Lemma~\ref{lem:CMES-negativity}}
\label{sec:proof-negative}
%%%%%%%%%%%%%%%%%%%%%%%%%%%%%%%%%%%%%%%%%%%%%%%%%%%%%%%%%%%%%%%%%%%%%%%%%%%%%%%%%%%%%%%%%%%%%%

\paragraph{Interpretation of negativity}

At first, we provide an intuition that the acquisition function of CMES becomes negative when $C > 5$.
In \eqref{eq:CMES_approx_truncated}, CMES replaces
$\EE_{f_*}[H(\*h_{\*x} | f_*)]$
with
$\EE_{f_*}[H(\*h_{\*x} | \*h_{\*x} \in \overline{\cA}_{f_*})]$, in which
the variable of the expectation $f_*$ and the condition of the entropy $\*h_{\*x} \in \overline{\cA}_{f_*}$ are not identical unlike $\EE_{f_*}[H(\*h_{\*x} | f_*)]$.
Thus, \eqref{eq:CMES_approx_truncated} cannot be seen as an MI anymore, and the non-negative guarantee is not maintained.
The negativity is derived by minimizing $\alpha^{\rm CMES} (\*x)$ with respect to
$\{z_c\}_{c=1}^C$
in Lemma~4.1.
The `$C  >  5$' is the boundary that an upper bound of this minimum becomes less than $0$.

\paragraph{Proof}

Before the proof of Lemma~\ref{lem:CMES-negativity}, we prove the following auxiliary lemma:
\begin{lem}
 \label{lem:ub_min_a}
Let
$a(\gamma) \coloneqq \frac{\gamma \phi(\gamma)}{1 - \Phi(\gamma)}$
for
$\gamma \in \RR$.
Then,
$\min_{\gamma \in \RR} a(\gamma) < - 0.29$.
\end{lem}
\begin{proof}
 % Then, we numerically confirm $\min_{\gamma \in \RR} a(\gamma) < a(-0.84) < - 0.29$
Using the approximation of CDF \citep[26.2.17 in][]{abramowitz-Handbook}, we obtain
    \begin{align}
        \Phi(\gamma) &= 1 - \phi(\gamma)(b_1 r + b_2 r^2 + b_3 r^3 + b_4 r^4 + b_5 r^5) + \epsilon(\gamma),
        \label{eq:approximation_CDF}
    \end{align}
    where
    \begin{align*}
        | \epsilon(\gamma) | &< 7.5 \times 10^{-8},
        r = 1 / (1 + p \gamma), p = 0.2316419, \nonumber \\
        b_1 = 0.319381530, b_2 &= - 1.821255978,
        b_3 = - 0.356563782, b_4 = 1.330274429,
        b_5 = 1.781477937. \nonumber
    \end{align*}
    By substituting \eqref{eq:approximation_CDF} into $a(\gamma)$, we obtain
    \begin{align*}
        a(\gamma) = \frac{\gamma}{b_1 r + b_2 r^2 + b_3 r^3 + b_4 r^4 + b_5 r^5 + \epsilon(\gamma) / \phi(\gamma)} \text{ for } \forall \gamma \in \RR.
    \end{align*}
    Since $\epsilon(\gamma) / \phi(\gamma) = \sqrt{2 \pi} e^{\gamma^2} \epsilon(\gamma) < 9 \epsilon (\gamma) < 9 \times 7.5 \times 10^{-8} < 10^{-6}$ for $\gamma \in [-1, 0]$, we obtain an upper bound
    \begin{align}
        a(\gamma) < \frac{\gamma}{b_1 r + b_2 r^2 + b_3 r^3 + b_4 r^4 + b_5 r^5 + 10^{-6}} \text{ for } \forall \gamma \in [-1, 0]. \label{eq:upper_bound_a}
    \end{align}
    This upper bound consists of elementary arithmetic that can be computed without approximations such as the numerical integration.
 We evaluated this upper bound \eqref{eq:upper_bound_a} with $28$ digits of precision by \emph{decimal} package (\url{https://docs.python.org/3/library/decimal.html}) in Python, and we confirmed
 $\min_{\gamma \in \RR} a(\gamma) < a(-0.84) < - 0.29$.
 % $a(-0.84) < -0.294374 < -0.29$.
\end{proof}

Next, we prove Lemma~\ref{lem:CMES-negativity}.
%
% \blue{The possibility of negative values in CMES can also be seen analytically.}
%
By using
$\log(1-Z) \geq (1 - \frac{1}{1-Z}) = \frac{-Z}{1-Z}$ for $Z \in [0, 1)$,
an upper bound of the CMES acquisition function \eqref{eq:CMES_final} can be derived as
\begin{align*}
 &
 \frac{1}{K}
 \! \!
 \sum_{\tilde{f}_* \in \cF_*}
 \! \!
 \left\{
 \frac{Z_{\*x}(\tilde{f}_*)}{2(1-Z_{\*x}(\tilde{f}_*))} \biggl( \frac{\gamma^{(f)}_{\*x}(\tilde{f}_*) \phi(\gamma^{(f)}_{\*x}(\tilde{f}_*))}{1 - \Phi(\gamma^{(f)}_{\*x}(\tilde{f}_*))}
 +
 \sum_{c=1}^C \frac{\gamma^{(g_c)}_{\*x}(z_c) \phi(\gamma^{(g_c)}_{\*x}(z_c))}{1 - \Phi(\gamma^{(g_c)}_{\*x}(z_c))} \biggr) - \log(1-Z_{\*x}(\tilde{f}_*))
 \right\}
 \\ % ---
 & \leq
 \frac{1}{K}
 \! \!
 \sum_{\tilde{f}_* \in \cF_*}
 \! \!
 \left\{
 \frac{Z_{\*x}(\tilde{f}_*)}{2(1-Z_{\*x}(\tilde{f}_*))} \biggl( \frac{\gamma^{(f)}_{\*x}(\tilde{f}_*) \phi(\gamma^{(f)}_{\*x}(\tilde{f}_*))}{1 - \Phi(\gamma^{(f)}_{\*x}(\tilde{f}_*))}
 +
 \sum_{c=1}^C \frac{\gamma^{(g_c)}_{\*x}(z_c) \phi(\gamma^{(g_c)}_{\*x}(z_c))}{1 - \Phi(\gamma^{(g_c)}_{\*x}(z_c))} \biggr) - \left(\frac{-Z_{\*x}(\tilde{f}_*)}{1-Z_{\*x}(\tilde{f}_*)} \right)
 \right\}
 \\ % ---
 & =
 \frac{1}{K}
 \! \!
 \sum_{\tilde{f}_* \in \cF_*}
 \! \!
 \frac{Z_{\*x}(\tilde{f}_*)}{(1-Z_{\*x}(\tilde{f}_*))}
 \left\{
 \frac{1}{2}
 \biggl( \frac{\gamma^{(f)}_{\*x}(\tilde{f}_*) \phi(\gamma^{(f)}_{\*x}(\tilde{f}_*))}{1 - \Phi(\gamma^{(f)}_{\*x}(\tilde{f}_*))}
 +
 \sum_{c=1}^C \frac{\gamma^{(g_c)}_{\*x}(z_c) \phi(\gamma^{(g_c)}_{\*x}(z_c))}{1 - \Phi(\gamma^{(g_c)}_{\*x}(z_c))} \biggr) + 1
 \right\}.
\end{align*}
%
% \blue{Let
% $a(\gamma) \coloneqq \frac{\gamma \phi(\gamma)}{1 - \Phi(\gamma)}$.}
%
Since $\frac{Z_{\*x}(\tilde{f}_*)}{1-Z_{\*x}(\tilde{f}_*)} \geq 0$, we see that the sign of the above upper bound is determined by
\begin{align}
 s(z_1,\ldots,z_C)
 \coloneqq
 \frac{1}{2}
 a(\gamma^{(f)}_{\*x}(\tilde{f}_*))
 +
 \frac{1}{2}
 \sum_{c=1}^C a(\gamma^{(g_c)}_{\*x}(z_c))
 + 1.
 \label{eq:sign_of_CMES}
\end{align}
Note that $s$ is defined as a function of $z_c$.
%

% Therefore, if we set $z_c$ so that $\gamma^{(g_c)}_{\*x}(z_c) = -0.84$ for all $c$,
From Lemma~\ref{lem:ub_min_a}, we see
\begin{align*}
 % \eqref{eq:sign_of_CMES}
 \min_{z_1,\ldots,z_C} s(z_1,\ldots,z_C)
 < \frac{1}{2}
 a(\gamma^{(f)}_{\*x}(\tilde{f}_*))
 -
 \frac{0.29 C}{2}
 + 1.
\end{align*}
%
% Moreover, since $f_*$ is the continuous random variable, whose support is $\RR$, the probability that $\gamma^{(f)}_{\*x}(f_*)$ is also smaller than $-0.29$ $\Pr(\gamma^{(f)}_{\*x}(f_*) < -0.29) > 0$.
%
% \red{
Moreover, since $\tilde{f}_*$ is sampled from $p(f_*)$ whose support is $\RR$, $a(\gamma^{(f)}_{\*x}(\tilde{f}_*))$ can be smaller than $-0.29$.
Hence, if $0.29 (C+1) / 2  > 1$, \eqref{eq:sign_of_CMES} can be negative.
% $\Pr\bigl( \eqref{eq:sign_of_CMES} < 0 \bigr) > 0$.
%
Thus, \eqref{eq:CMES_final} can be negative when $C > 5$, since $0.29 (C+1) / 2 = 0.29 (6+1) / 2 = 2.03 / 2 > 1$ when $C=6$.
% }

% \blue{Moreover, since $f_*$ is the continuous random variable whose support is $\RR$, the probability that $a(\gamma^{(f)}_{\*x}(f_*))$ is smaller than $-0.29$ is also positive: $\Pr(a(\gamma^{(f)}_{\*x}(f_*)) < -0.29) > 0$.
%
% Hence, if $0.29 (C+1) / 2  > 1$, the probability that \eqref{eq:sign_of_CMES} becomes minus is not zero.
%
% Thus, \eqref{eq:sign_of_CMES} can be minus when $C > 5$, since $0.29 (C+1) / 2 = 0.29 (6+1) / 2 = 2.03 / 2 > 1$ when $C=6$.}

%%%%%%%%%%%%%%%%%%%%%%%%%%%%%%%%%%%%%%%%%%%%%%%%%%%%%%%%%%%%%%%%%%%%%%%%%%%%%%%%%%%%%%%%%%%%%%
\subsection{Illustrative Example that CMES Takes a Negative Value}
\label{sec:example_negative}
%%%%%%%%%%%%%%%%%%%%%%%%%%%%%%%%%%%%%%%%%%%%%%%%%%%%%%%%%%%%%%%%%%%%%%%%%%%%%%%%%%%%%%%%%%%%%%
%
% \begin{wrapfigure}[32]{r}[0pt]{0.4\textwidth}
\begin{figure}[t]
%  \cigr{0.4}{CMES_negative_example.pdf}
    \cigr{0.75}{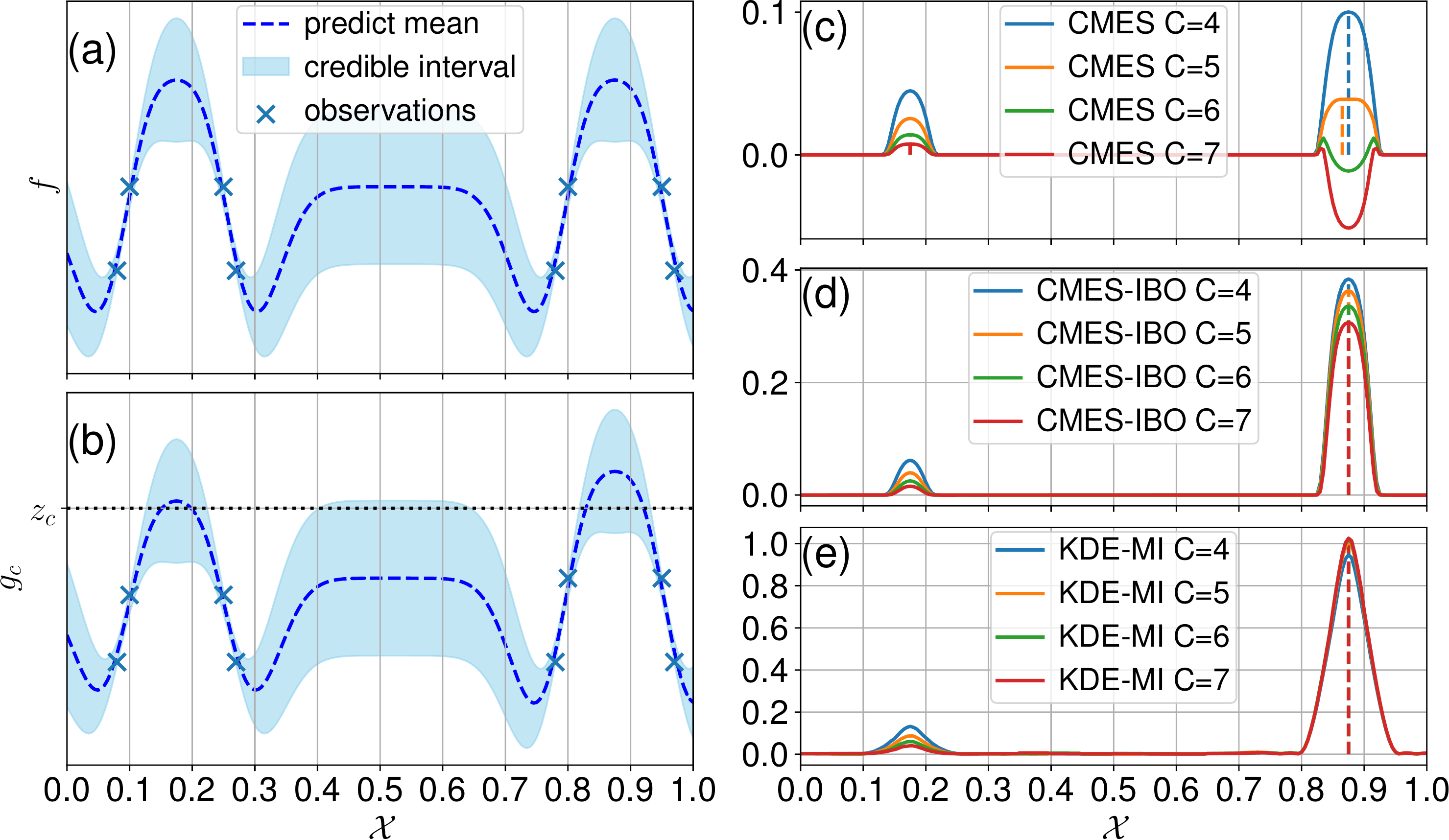}
    \caption{
    In (a) and (b), the predictions of $f$ and $g_c$ are shown, respectively.
    %The first and second from the top figures show the predictions of $f$ and $g_c$, respectively.
    %
    The solid line and shaded area represent the predictive mean and the credible interval, and the cross mark is the training data.
    The horizontal dotted line expresses $z_c$.
    In (c)-(e), CMES, CMES-IBO, and the KDE-based MI approximation for $C=4,5,6,$ and $7$ are shown, and each vertical dotted line indicates the maximum.
    }
    \label{fig:CMES-negative_example-1000}
\end{figure}
% \end{wrapfigure}

First, we describe the detailed setting of Figure~\ref{fig:CMES-negative_example-10} omitted in the main text.
In the plots, we have $200$ equally spaced grid points as $\cX$.
We sampled $f_*$ through the sampling from the predictive multivariate normal distribution on these grid points.
We assume the posterior of $g_c$ for $\forall c$ are the same for brevity, and this may seem to contradict the assumption that $f$ and $g_c$ for $\forall c$ are independent.
However, since the independence assumption does not mean that observations of different functions cannot take the same value, Fig.~\ref{fig:CMES-negative_example-10} does not contradict the assumption.
In fact, this setting (same observations for all constraints) is not an essential requirement for negativity.
% \blue{We used the same observations for the brevity of the plots (not an essential requirement).}
%
Typically, CMES \eqref{eq:CMES_final} is negatively biased when $(z_c - \mu^{(g_c)}(\*x)) / \sigma^{(g_c)}(\*x) \approx -0.84$ for $\forall c$ that can happen even if observations are different for each constraint.
%
% Moreover, since $(z_c - \mu_{t-1}^{(g_c)}(\*x)) / \sigma_{t-1}^{(g_c)}(\*x) \approx -0.84$ suggests a high probability satisfying the $c$-th constraint $\Pr(g_c(\*x) \geq z_c) \approx 0.8$, this negative bias is problematic.
Moreover, since $(z_c - \mu_{t-1}^{(g_c)}(\*x)) / \sigma_{t-1}^{(g_c)}(\*x) \approx -0.84$ suggests a high probability satisfying the $c$-th constraint $\Pr(g_c(\*x) \geq z_c) \approx 0.8$, this negative bias can be a significant problem.

The detailed procedure of KDE-MI is as follows.
We used the following representation of the MI \citep{Nair2007-On} because, in this form, KDE is required only for one dimensional densities:
% \blue{
% \begin{align*}
%     {\rm MI}(\*h_{\*x} ; f_*) &= \EE_{\*h_{\*x}} \biggl[ \int p(f_* | \*h_{\*x}) \log \frac{p(f_* | \*h_{\*x})}{p(f_*)} {\rm d}f_* \biggr],
% \end{align*}
% }
% \red{
\begin{align*}
    {\rm MI}(\*h_{\*x} ; f_*) = \EE_{\*h_{\*x}} \biggl[ &\Pr(f_* = -\infty | \*h_{\*x}) \log \frac{\Pr(f_* = -\infty | \*h_{\*x})}{\Pr(f_* = -\infty)}
    + \Pr(f_* \neq -\infty | \*h_{\*x}) \log \frac{\Pr(f_* \neq -\infty | \*h_{\*x})}{\Pr(f_* \neq -\infty)} \\
    &+ \Pr(f_* \neq -\infty | \*h_{\*x}) \int p(f_* | \*h_{\*x}, f_* \neq -\infty) \log \frac{p(f_* | \*h_{\*x}, f_* \neq -\infty)}{p(f_* | f_* \neq -\infty)} {\rm d} f_*
     \biggr],
\end{align*}
% }
where the expectation over $\*h_{\*x}$ is approximated by the MC estimation with $10000$ samples.
The probabilities $\Pr(f_* = -\infty | \*h_{\*x})$ and $\Pr(f_* = -\infty)$ are estimated by the percentage of infeasible samples in $10000$ samples ($\Pr(f_* \neq -\infty | \*h_{\*x})$ and $\Pr(f_* \neq -\infty)$ are estimated in the same way).
The one-dimensional density functions
% $p(f_*)$ and $p(f_* | \*h_{\*x})$
$p(f_* | f_* \neq -\infty)$ and $p(f_* | \*h_{\*x}, f_* \neq -\infty)$
are estimated by KDE fitted to feasible samples among $10000$ samples of $f_*$ and $f_* | \*h_{\*x}$, respectively.
%
% Note that $p(f_* | \*h_{\*x})$ is approximated for each MC sample of $\*h_{\*x}$.
Note that $p(f_* | \*h_{\*x})$ is approximated by generating $10000$ $f_*$ for each MC sample of $\*h_{\*x}$.
We approximate the one-dimensional integral with respect to $f_*$ in the outer expectation by Gauss--Legendre quadrature.
%
%
% We also empirically confirmed that the estimation variance was low enough.
We empirically confirmed that the estimation variance of this procedure was low enough.

Here, we show an additional example in Figure~\ref{fig:CMES-negative_example-1000}, in which the number of the MC samplings in CMES and CMES-IBO is set as $K=10000$.
The other settings are the same as Figure~\ref{fig:CMES-negative_example-10}.
Since we set $K$ as $10000$ in Figure~\ref{fig:CMES-negative_example-1000}, we expect that the MC estimation are highly accurate.
For CMES, we see that the behavior is slightly improved compared with the case of $K = 10$ in Figure~\ref{fig:CMES-negative_example-10} (In particular, the acquisition function of $C = 5$ in Figure~\ref{fig:CMES-negative_example-1000}~(c) is more reasonable than that of Figure~\ref{fig:CMES-negative_example-10}~(c)).
% In Figure~\ref{fig:CMES-negative_example-10}, CMES takes more unstable behavior by smaller $K = 10$ (e.g., when $C=5$, the acquisition function around $0.9$ is smaller than the case of $K=10000$).
However, in the case of $C = 6$ and $7$, CMES still takes negative values.
%
% However, In the case of $C = 6$ and $7$, we can clearly see that CMES takes negative values in the third plot.
%
% In this example, the inputs around $0.2$ and $0.9$ seem better because both of predictions of $f(\*x)$ and $g_c(\*x)$ is large.
In this example, querying to the inputs around $0.2$ and $0.9$ are obviously beneficial because both predictions of $f(\*x)$ and $g_c(\*x)$ are large.
%
% Moreover, the inputs around $0.9$ seem best because the probability of being feasible is the highest.
In particular, the inputs around $0.9$ can be a better selection than around $0.2$ because the probability of being feasible is the highest, while the uncertainty is almost the same as around $0.2$.
%
% \red{In practice, the $\argmax$ of KDE-MI and CMES-IBO are the same point around $0.9$ regardless of the number of constraints $C$.}
The $\argmax$ of KDE-MI and CMES-IBO are the same point around $0.9$ regardless of the number of constraints $C$.
%
% On the other hand, CMES select the same are for $C=4$, but the input selected by CMES shifts to the left for $C=5$.
On the other hand, the acquisition function values of CMES for a point around $0.9$ rapidly decrease with the number of constraints $C$.
Although CMES selects a point around $0.9$ for $C=4$ and $5$, but for $C = 6$ and $7$, CMES selects a point around $0.2$.
%

%%%%%%%%%%%%%%%%%%%%%%%%%%%%%%%%%%%%%%%%%%%%%%%%%%%%%%%%%%%%%%%%%%%%%%%%%%%%%%%%%%%%%%%%%%%%%%
\section{Derivation of Entropy of TMN}
\label{sec:derivation_entropy_tmvn}
%%%%%%%%%%%%%%%%%%%%%%%%%%%%%%%%%%%%%%%%%%%%%%%%%%%%%%%%%%%%%%%%%%%%%%%%%%%%%%%%%%%%%%%%%%%%%%
% In this section, we derive the entropy of TMN truncated by the one hyperrectangle $H(\*h_{\*x} | \*h_{\*x} \in \cA_{f_*})$ for both independent and correlated cases.
In this section, we derive the entropy of TMN truncated by a hyperrectangle $H(\*h_{\*x} | \*h_{\*x} \in \cA_{\tilde{f}_*})$ for both independent and correlated cases.

\subsection{Independent Case}
In this case, $H(\*h_{\*x} | \*h_{\*x} \in \cA_{\tilde{f}_*})$ can be decomposed to the sum of the entropy of each element \citep{Horrace2005-Some}:
\begin{align*}
    H(\*h_{\*x} | \*h_{\*x} \in \cA_{\tilde{f}_*}) = H\bigl( f(\*x) | f(\*x) \geq \tilde{f}_* \bigl) + \sum_{c=1}^C H\bigl( g_c(\*x) | g_c(\*x) \geq z_c \bigl).
\end{align*}
The analytical form of the entropy of the one-dimensional truncated normal distribution is known \citep{Michalowicz2014-Handbook}, by which we obtain
\begin{align*}
    H\bigl( f(\*x) | f(\*x) \geq \tilde{f}_* \bigl) &= \log\bigl(\sqrt{2 \pi e} \sigma^{(f)}(\*x) (1 - \Phi(\gamma^{(f)}_{\*x}(\tilde{f}_*)))\bigl) + \frac{\gamma^{(f)}_{\*x}(\tilde{f}_*) \phi(\gamma^{(f)}_{\*x}(\tilde{f}_*))}{2\bigl(1 - \Phi(\gamma^{(f)}_{\*x}(\tilde{f}_*))\bigl)}, \\
 H\bigl( g_c(\*x) | g_c(\*x) \geq z_c \bigl) &= \log\bigl(\sqrt{2 \pi e} \sigma^{(g_c)}(\*x) (1 - \Phi(\gamma^{(g_c)}_{\*x}(z_c)))\bigl) + \frac{\gamma^{(g_c)}_{\*x}(z_c)\phi(\gamma^{(g_c)}_{\*x}(z_c))}{2\bigl(1 - \Phi(\gamma^{(g_c)}_{\*x}(z_c))\bigl)} . %,\ c=1,\dots, C.
 %H\bigl( g_c(\*x) | f(\*x) \geq z_c \bigl) &= \log\bigl(\sqrt{2 \pi e} \sigma^{(g_c)}(\*x) (1 - \Phi(\gamma_t^{(g_c)}(\*x)))\bigl) + \frac{\gamma_t^{(g_c) }(\*x)\phi(\gamma_t^{(g_c)}(\*x))}{2\bigl(1 - \Phi(\gamma_t^{(g_c)}(\*x))\bigl)} . %,\ c=1,\dots, C.
\end{align*}
Moreover, from the independence of each elements of $\*h_{\*x}$, we see $H(\*h_{\*x}) = \log\bigl(\sqrt{2 \pi e} \sigma^{(f)}(\*x) \bigl) + \sum_{c=1}^C \log\bigl(\sqrt{2 \pi e} \sigma^{(g_c)}(\*x) \bigl) $.
Then, the entropy of TMN is obtained as
% \begin{align*}
%     H(\*h_{\*x} | \*h_{\*x} \in \cA_{\tilde{f}_*}) &= \log\bigl(\sqrt{2 \pi e} \sigma^{(f)}(\*x) (1 - \Phi(\gamma_t^{(f)}(\*x)))\bigl) + \frac{\gamma_t^{(f)}(\*x) \phi(\gamma_t^{(f)}(\*x))}{2\bigl(1 - \Phi(\gamma_t^{(f)}(\*x))\bigl)} \\
%     &\ + \sum_{c=1}^C\biggl( \log\bigl(\sqrt{2 \pi e} \sigma^{(g_c)}(\*x) (1 - \Phi(\gamma_t^{(g_c)}(\*x)))\bigl) + \frac{\gamma_t^{(g_c) }(\*x)\phi(\gamma_t^{(g_c)}(\*x))}{2\bigl(1 - \Phi(\gamma_t^{(g_c)}(\*x))\bigl)} \biggl)\\
%     &= \log\bigl(\sqrt{2 \pi e} \sigma^{(f)}(\*x) \bigl) + \frac{\gamma_t^{(f)}(\*x)\phi(\gamma_t^{(f)}(\*x))}{2\bigl(1 - \Phi(\gamma_t^{(f)}(\*x))\bigl)}
%     + \sum_{c=1}^C \biggl( \log\bigl(\sqrt{2 \pi e} \sigma^{(g_c)}(\*x) \bigl) + \frac{\gamma_t^{(g_c)}(\*x)\phi(\gamma_t^{(g_c)}(\*x))}{2\bigl(1 - \Phi(\gamma_t^{(g_c)}(\*x))\bigl)} \biggl) + \log(Z).
% \end{align*}
% %

% %
% % Thus, we obtain
% By using $H(\*h_{\*x})$, the last equation can be re-written as
\begin{align*}
 H(\*h_{\*x} | \*h_{\*x} \in \cA_{\tilde{f}_*})
    &= H(\*h_{\*x}) + \log(Z_{\*x}(\tilde{f}_*)) + \frac{R_{\tilde{f}_*}}{2},
    % \label{eq:entropy_in_A}
 % \\
\end{align*}
where
\begin{align}
    R_{\tilde{f}_*} &= \frac{\gamma^{(f)}_{\*x}(\tilde{f}_*)\phi(\gamma^{(f)}_{\*x}(\tilde{f}_*))}{1 - \Phi(\gamma^{(f)}_{\*x}(\tilde{f}_*))}
    + \sum_{c=1}^C  \frac{\gamma^{(g_c)}_{\*x}(z_c)\phi(\gamma^{(g_c)}_{\*x}(z_c))}{1 - \Phi(\gamma^{(g_c)}_{\*x}(z_c))}. \nonumber
\end{align}

%%%%%%%%%%%%%%%%%%%%%%%%%%%%%%%%%%%%%%%%%%%%%%%%%%%%%%%%%%%%%%%%%%
\subsection{Correlated Case}
\label{ssec:correlated_tmvn}
% We here consider the correlated setting.
%
Supposed that $\*h_{\*x}$ has a correlated distribution defined as
\begin{align*}
    \*h_{\*x} | \cD_{t-1} \sim \cN \bigl(\*\mu(\*x), \*\Sigma(\*x) \bigl),
\end{align*}
where $\*\mu(\*x)$ {and} $\*\Sigma(\*x)$ can be obtained by an arbitrary multi-output GP model \citep{Rasmussen2005-Gaussian}.
Let $\*\mu^{\rm TN}(\*x)$ {and} $\*\Sigma^{\rm TN}(\*x)$ be the expectation and covariance matrix of $\*h_{\*x} | \*h_{\*x} \in \cA_{\tilde{f}_*}$, respectively, whose analytical expressions are known \citep{Manjunath2012-moments}.
We denote $\EE_{\*h_{\*x} | \*h_{\*x} \in \cA_{\tilde{f}_*}}$, which is the expectation with respect to $\*h_{\*x} | \*h_{\*x} \in \cA_{\tilde{f}_*}$, as $\EE_{\rm TN}$ for brevity.
Then, we can transform the entropy $H(\*h_{\*x} | \*h_{\*x} \in \cA_{\tilde{f}_*})$ into
\begin{align*}
    H(\*h_{\*x} | \*h_{\*x} \! &\in \! \cA_{\tilde{f}_*}) \! = \! \EE_{\rm TN} \biggl[ - \log p(\*h_{\*x} | \*h_{\*x} \in \cA_{\tilde{f}_*}) \biggl] \\
    &= \EE_{\rm TN} \biggl[ - \log \frac{p(\*h_{\*x})}{Z} \biggl] \\
    &= \EE_{\rm TN} \bigl[ - \log p(\*h_{\*x}) \bigl] + \log Z \\
    &= \frac{1}{2} \EE_{\rm TN} \bigl[ \bigl(\*h_{\*x} - \*\mu(\*x)\bigl)^\top \*\Sigma^{-1}(\*x) \bigl(\*h_{\*x} - \*\mu(\*x)\bigl) \bigl] + \frac{1}{2} | 2 \pi \*\Sigma(\*x) | + \log Z \\
    &= \frac{1}{2} {\rm Tr} \Bigl( \! \*\Sigma^{-1}(\*x) \underbrace{\EE_{\rm TN} \bigl[ \bigl(\*h_{\*x} - \*\mu(\*x)\bigl) \! \bigl(\*h_{\*x} - \*\mu(\*x)\bigl)^\top \bigl] \! \Bigl)}_{\eqqcolon \*V} + \frac{1}{2} | 2 \pi \*\Sigma(\*x) | + \log Z,
\end{align*}
where $|\cdot|$ is the determinant of a matrix.
Further, by using $\*d = \*\mu^{\rm TN}(\*x) - \*\mu(\*x)$, we obtain
\begin{align*}
    \*V &= \EE_{\rm TN} \bigl[ \bigl(\*h_{\*x} - \*\mu^{\rm TN}(\*x) + \*d \bigl) \bigl(\*h_{\*x} - \*\mu^{\rm TN}(\*x) + \*d \bigl)^\top \bigl] \\
    &= \EE_{\rm TN} \bigl[ \bigl(\*h_{\*x} - \*\mu^{\rm TN}(\*x) \bigl) \bigl(\*h_{\*x} - \*\mu^{\rm TN}(\*x) \bigl)^\top + \*d \bigl(\*h_{\*x} - \*\mu^{\rm TN}(\*x) \bigl)^\top + \bigl(\*h_{\*x} - \*\mu^{\rm TN}(\*x) \bigl) \*d^\top + \*d \*d^\top  \bigl] \\
    &= \EE_{\rm TN} \bigl[ \bigl(\*h_{\*x} - \*\mu^{\rm TN}(\*x) \bigl) \bigl(\*h_{\*x} - \*\mu^{\rm TN}(\*x) \bigl)^\top \bigl] + \*d \*d^\top \\
    &= \*\Sigma^{\rm TN}(\*x) + \*d \*d^\top,
\end{align*}
where we use $\EE_{\rm TN} \bigl[ \bigl(\*h_{\*x} - \*\mu^{\rm TN}(\*x) \bigl) \bigl] = 0$.
% where we use the fact that $\EE_{\rm TN} \bigl[ \bigl(\*h_{\*x} - \*\mu^{\rm TN}(\*x) \bigl) \bigl] = 0$.
%
Consequently, we can derive
\begin{align*}
    H(\*h_{\*x} | \*h_{\*x} \in \cA_{\tilde{f}_*}) = \frac{1}{2} {\rm Tr} \Bigl( \*\Sigma^{-1}(\*x) \bigl( \*\Sigma^{\rm TN}(\*x) + \*d \*d^\top \bigl) \Bigl)
     + \frac{1}{2} | 2 \pi \*\Sigma(\*x) | + \log Z.
    %  \label{eq:entropy_in_A_correlated}
\end{align*}
%%%%%%%%%%%%%%%%%%%%%%%%%%%%%%%%%%%%%%%%%%%%%%%%%%%%%%%%%%%%%%%%%%%%%%%%%%%%%%%%%%%%%%%%%%%%%%
\section{Computation of CMES-IBO and CMES}
\label{sec:computation}

In this section, we describe several computational details.
First, we present the general algorithm for sequential- and parallel- CMES-IBO in Algo.~\ref{alg:CMES-IBO}, where sequential CMES-IBO corresponds to the case that $Q=1$.
%
% By replacing \eqref{eq:CMES-IBO} and \eqref{eq:PCMES-IBO_final} to that of CMES, this algorithm can also be seen as the algorithm for sequential- and parallel- CMES.
By replacing \eqref{eq:CMES-IBO} and \eqref{eq:PCMES-IBO_final} in Algo.~\ref{alg:CMES-IBO} with those of CMES \eqref{eq:CMES_final} and \eqref{eq:PCMES_final}, this algorithm can also be seen as the algorithm for sequential- and parallel- CMES.

%%%%%%%%%%%%%%%%%%%%%%%%%%%%%%%%%%%%%%%%%%%%%%%%%%%%%%%%%%%%%%%%%%%%%%%%%%%%%%%%%%%%%%%%%%%%%%
% Algorithm
%%%%%%%%%%%%%%%%%%%%%%%%%%%%%%%%%%%%%%%%%%%%%%%%%%%%%%%%%%%%%%%%%%%%%%%%%%%%%%%%%%%%%%%%%%%%%%
\begin{algorithm}[t]
    \caption{Sequential- and parallel- CMES-IBO.}
    \label{alg:CMES-IBO}
    \begin{algorithmic}[1]
     \Function{CMES-IBO}{$\cD_0, \cX, Q, K$}
     \For{$t = 0, \ldots, T$}
        \For{$k = 1, \ldots, K$}
        \State Sample $\tilde{f}_{(k)}, \tilde{g}_{1(k)}, \dots, \tilde{g}_{C(k)}$ from current GPs
        % \State $f_{*(k)} \leftarrow \max_{\*x \in \cX} \tilde{f}_{(k)}(\*x), {\rm s.t.} \forall c, \tilde{g}_{c(k)}(\*x) \geq z_c$%,\ c=1,\dots, C$
        \State {\parbox{30em}{
     % \begin{align*}
     $\tilde{f}_{*(k)} \leftarrow
        \begin{cases}
        \max_{\*x \in \tilde{\cX}_{\rm feasible}} \tilde{f}_{(k)}(\*x), & \text{if } \tilde{\cX}_{\rm feasible} \neq \emptyset, \\
         - \infty, & \text{otherwise,}
        \end{cases}$ \\
       % \end{align*}
        where
        $\tilde{\cX}_{\rm feasible} \coloneqq \{ \*x \mid \tilde{g}_c(\*x) \geq z_c, c = 1, \ldots, C \}$}
        }
        \EndFor
     \State $\cF_* = \{ \tilde{f}_{*(k)} \}_{k=1}^K$
     \State $\*x^{(1)}_{t} \leftarrow \argmax_{\*x \in \cX} \alpha^{\rm IBO}(\*x)$ \eqref{eq:CMES-IBO}
     \If{$Q > 1$}
        \For{$q = 2, \ldots, Q$}
        \State $\cJ = \Bigl\{ \Bigl( {\tilde{f}}_{*(k)},
        \bigl\{
            % \tilde{\*h}_{\*x^{(q^\prime)}_t } \coloneqq
            \bigl( \tilde{f}_{(k)}(\*x^{(q^\prime)}_t ), \tilde{g}_{1(k)}(\*x^{(q^\prime)}_t ), \dots, \tilde{g}_{C(k)}(\*x^{(q^\prime)}_t ) \bigl)^\top
            \bigl\}_{q^\prime=1}^{q-1}
            \Bigl) \Bigl\}_{k=1}^K$
            % , where $\tilde{\*h}_{\*x^{(q^\prime)}_t } \coloneqq \bigl( \tilde{f}_{(k)}(\*x^{(q^\prime)}_t ), \tilde{g}_{1(k)}(\*x^{(q^\prime)}_t ), \dots, \tilde{g}_{C(k)}(\*x^{(q^\prime)}_t ) \bigl)^\top $
        \State $\*x^{(q)}_{t} \leftarrow \argmax_{\*x \in \cX} \alpha^{\rm IBO}_{t}(\*x | \cX_{q-1})$ \eqref{eq:PCMES-IBO_final}
        \EndFor
     \EndIf
     \State Evaluate $f, g_1, \dots, g_C$ at $\cX_q$
     \State Update $\cD^{(f)}_{t}, \cD^{(g_1)}_{t}, \dots, \cD^{(g_C)}_{t}$ adding new observations
     \EndFor
     \EndFunction
    \end{algorithmic}
\end{algorithm}
%%%%%%%%%%%%%%%%%%%%%%%%%%%%%%%%%%%%%%%%%%%%%%%%%%%%%%%%%%%%%%%%%%%%%%%%%%%%%%%%%%%%%%%%%%%%%%
%%%%%%%%%%%%%%%%%%%%%%%%%%%%%%%%%%%%%%%%%%%%%%%%%%%%%%%%%%%%%%%%%%%%%%%
\subsection{Sampling from Posterior}
\label{sec:sampling_fstar}
%%%%%%%%%%%%%%%%%%%%%%%%%%%%%%%%%%%%%%%%%%%%%%%%%%%%%%%%%%%%%%%%%%%%%%%
Our proposed method needs to sample the maximum value $f_*$ defined by \eqref{eq:f_star}.
% $f_* \coloneqq \max_{x \in \cX} f(\*x),\ {\rm s.t.}\ g_c(\*x) \geq z_c,\ c=1, \dots, C$.
%
We employ an approach using random Fourier features (RFF) \citep{Rahimi2008-Random}, which has been used in various entropy-based BO and CBO methods \citep{Hernandez2014-Predictive, Hernandez-Lobato2015-Predictive, Wang2017-Max}.
In the RFF-based sampling, RFF $\*\phi(\*x) \in \RR^D$ are first generated, and a Bayesian linear regression model $f(\*x) \approx \*\omega^\top \*\phi(\*x)$ with the coefficients $\*\omega \in \RR^D$ is constructed.
By sampling the coefficients $\*\omega$ from the Gaussian posterior, a continuous sample path can be derived.
%% Then, the posterior of $\*\omega$ is normally distributed, and a continuous sample path can be derived
%
Let $\tilde{f}_{(k)}$ and $\tilde{g}_{c(k)} (c \in \{1,\ldots,C\})$ be the $k$-th set of sample paths ($k \in \{1,\ldots,K\}$) for the objective and constraint functions.
The $k$-th sample of $f_*$, can be obtained by solving a constrained optimization problem
$\max_{x \in \cX} \tilde{f}_{(k)}(\*x),\ {\rm s.t.}\ \tilde{g}_{c(k)}(\*x) \geq z_c,\ c=1, \dots, C$.
We can apply any constrained optimization modules such as the method of moving asymptotes \citep{Krister2002-Class}.
%
% Given sample paths of objective $f$ and constraint functions $g_c$, the maximum $f_*$ can be obtained by using some constrained optimization module such as the method of moving asymptotes \citep{Krister2002-Class}.
%
Another approach would be the discretization-based sampling through multivariate normal distribution \citep{Perrone2019-Constrained}, though the number of discretization points strongly depends on the input dimension $d$ for sufficiently accurate sampling.

For parallel querying, we need to sample the $f_*$ and $\cH_{q-1}$ simultaneously.
By using RFF-based sampling, these random variables are sampled from a joint distribution easily, i.e., $\cH_{q-1}$ can be sampled as a value of the sample paths
$\tilde{f}_{(k)}(\*x)$ and $\tilde{g}_{c(k)}(\*x)$ at $\*x \in \cX_{q-1}$.
% $\tilde{f}_{(k)}, \tilde{g}_{1(k)}, \dots, \tilde{g}_{C(k)}$ at $\cX_{q-1}$.
%
Moreover, in the MC approximation for $q > 1$, we can reuse the sample paths $\tilde{f}_{(k)}, \tilde{g}_{1(k)}, \dots, \tilde{g}_{C(k)}$ that are sampled at $q = 1$ as described in Algo.~\ref{alg:CMES-IBO}.
Because of this reuse, the computational cost of parallel CMES and CMES-IBO are much smaller than $Q$ times that of sequential CMES and CMES-IBO, respectively.

%%%%%%%%%%%%%%%%%%%%%%%%%%%%%%%%%%%%%%%%%%%%%%%%%%%%%%%%%%%%
\subsection{Computational Complexity of CMES-IBO and CMES}
\label{sec:computational_complexity}
%%%%%%%%%%%%%%%%%%%%%%%%%%%%%%%%%%%%%%%%%%%%%%%%%%%%%%%%%%%%

\paragraph{Independent Case}

% To generate sample paths, the Bayesian linear regression requires $O(C \blue{K} D^3 \red{+ K D^2})$ for the given RFF, but $C, K$, and $D$ are usually not large values.
To generate sample paths, the Bayesian linear regression requires $O(C D^3 + K D^2)$ for the given RFF, but $C, K$, and $D$ are usually not large values.
The number of constraints $C$ is often less than $10$.
In our empirical observations, the performance of CMES-IBO were stable with a small number of MC samplings $K$ such as $10$.
The dimension of the RFF features $D$ is typically set as less than $1000$.
The complexity for solving the generated constraint problems depends on the solver.
Many gradient-based methods have been known for the white-box constrained optimization such as the sequential quadratic programming and the interior point method.
The gradient of a sample path can be obtained by $O(d D)$, and thus, those gradient-based solvers can be applied efficiently.

Once we obtain the samples of $f_*$, the acquisition functions CMES-IBO \eqref{eq:CMES-IBO} and CMES \eqref{eq:CMES_final} can be easily calculated for $\forall \*x$ by using the GP posterior of $f$ and $g_c$.
Since we employ a standard GP model, the posterior can be computed with $O(n^3)$ for a given kernel matrix, where $n$ is the number of the observed points.
Note that the $O(n^3)$ computation is required only once through the acquisition function maximization, because by storing $(\*K + \sigma_{\rm noise}^2 \*I)^{-1}$, the predictive mean and variance can be evaluated by $O(n^2)$ for a given $\*x$.
In our setting, the function evaluation is assumed to be expensive, by which we usually only have a moderate size of a training set (typically, at most several hundreds of points).

In the case of the parallel setting, we need $m_{\*x^{(q)}}^{(f)}$ and $s_{\*x^{(q)}}^{(f)2}$, which are the predictive mean and variance of $f(\*x)$ after conditioning $\cH_{q-1}$.
This conditional density is written as
$p(f(\*x) \mid \tilde{f}_{(k)}(\*x^{(1)}), \ldots, \tilde{f}_{(k)}(\*x^{(q-1)}))$,
which can be easily determined through the standard conditional Gaussian formula.
Given the $q \times q$ predictive covariance matrix of the current GP for $\*x^{(q)}$ and $\cX_{q-1}$, we can obtain $m_{\*x^{(q)}}^{(f)}(\*x)$ and $s_{\*x^{(q)}}^{(f)2}(\*x)$ with $O(q^3)$.
For $m_{\*x^{(q)}}^{(g_c)}$ and $s_{\*x^{(q)}}^{(g_c)2}$, which are for constraint functions, the same calculation can be applied.

\paragraph{Correlated Case}
For multi-output GP model, the posterior can be computed with $O(C^3 n^3)$.
% For CMES-IBO and CMES, the computations of CDF of the multivariate normal distribution are needed in the correlated case.
For CMES-IBO and CMES, the computations of CDF of the multivariate normal distribution are needed in the correlated case, as shown in Appendix~\ref{ssec:correlated_tmvn}.
CMES-IBO performs $K$ times $C+1$ dimensional CDF {for the acquisition function \eqref{eq:CMES-IBO}, in which $Z_{\*x}(\tilde{f}_*) = \Pr(\*h \in \cA_{\tilde{f}_*})$ is now CDF of a correlated multivariate normal distribution.}
On the other hand, in addition to these CDF computations, CMES performs $K$ times computations of the expectation and covariance matrix of TMN in {the acquisition function \eqref{eq:CMES_correlated_final}}, each of which needs computing $C$ dimensional CDF $C+1$ times and $C-1$ dimensional CDF $C(C+1)/2$ times, respectively \citep{Manjunath2012-moments}.
Using a well-known MC-based algorithm \citep{Genz1992-numerical}, the computational complexity of CDF with respect to the dimension $C$ is known to be $O(C^2)$.
Therefore, the computational complexity of CMES-IBO with respect to $K$ and $C$ is $O(KC^2)$, but that of CMES is $O(KC^4)$, which severely limits the applicability of CMES to the large number of constraints.

%%%%%%%%%%%%%%%%%%%%%%%%%%%%%%%%%%%%%%%%%%%%%%%%%%%%%%%%%%%%%%%%%%%%%%%%%%%%%%%%%%%%%%%%%%%%%%
\section{Considerations on Regret Bound of MES}
\label{sec:about_regret_MES}
%%%%%%%%%%%%%%%%%%%%%%%%%%%%%%%%%%%%%%%%%%%%%%%%%%%%%%%%%%%%%%%%%%%%%%%%%%%%%%%%%%%%%%%%%%%%%%
In this section, we describe flaws in the theoretical analysis of MES \citep{Wang2017-Max}.
We use the same notation as \citep{Wang2017-Max} throughout this section.
\citet{Wang2017-Max} showed the bound of simple regret $r_T \coloneqq \max_{\*x \in \cX} f(\*x) - \max_{t \in [1, T]}f(\*x_t)$ as follows:
\begin{theo}[Theorem~3.2 in \citep{Wang2017-Max}]
    Let $F$ be the cumulative probability distribution for the maximum of any function $f$ sampled from $GP(\mu, k)$ over the compact search space $\cX \subset \RR^d$, where $k(\*x, \*x^\prime) \leq 1,\ \forall\*x, \*x^\prime \in \cX$.
    Let $f_* = \max_{\*x \in \cX} f(\*x)$ and $w = F(f_*) \in (0, 1)$, and assume the observation noise is iid $\cN(0, \sigma^2)$.
    If in each iteration $t$, the query point is chosen as $\*x_t = \argmax_{\*x \in \cX} \gamma_{y^{t}_*}(\*x) \frac{\psi(\gamma_{y^{t}_*}(\*x))}{2\Psi(\gamma_{y^{t}_*}(\*x))} - \log(\Psi(\gamma_{y^{t}_*}(\*x)))$, where $\gamma_{y^{t}_*}(\*x) = \frac{y^{t}_* - \mu_t(\*x)}{\sigma_t(\*x)}$ and $y^{t}_*$ is drawn from $F$, then with probability at least $1 - \delta$, in $T^\prime = \sum_{i=1}^T \log_{w}\frac{\delta}{2 \pi_i}$ number of iterations, the simple regret satisfies
    \begin{align}
        r_{T^\prime} \leq \sqrt{\frac{C \rho_T}{T}}(\nu_{t^*} + \zeta_T), \label{eq:MES_regret_bound}
    \end{align}
    where $C = 2/\log(1 + \sigma^{-2})$ and $\zeta_T = (2 \log (\frac{\pi_T}{\delta}))^{\frac{1}{2}}$, $\pi_i$ satisfies $\sum_{i=1}^T \pi_i^{-1} \leq 1$ and $\pi_t > 0$, and $t^* = \argmax_{t} \nu_t$ with $\nu_t \triangleq \min_{\*x \in \cX, y^{t}_* > f_*} \gamma_{y^{t}_*}(\*x)$, and $\rho_T$ is the maximum information gain of at most $T$ selected points.
    \label{theo:MES_regret_bound}
\end{theo}
In this theorem, the simple regret for `one sample MES' is analyzed, in which only one max-value is sampled at every iteration in the algorithm.
We summarize five main flaws in this theorem as follows (note that each of them is related to each other):
\begin{enumerate}
    \item From the assumption $f \sim \cG \cP(\mu, k)$, the max-value $f_*$ is a random variable, but it is not treated as a random variable in their analysis.
          For example, Lemma~C.1 in \citep{Wang2017-Max} obviously assumes that $f(\*x_t)$ follows the Gaussian distribution, and therefore, treating $f_*$ as a deterministic variable contradicts the assumption of this lemma.
            This problem closely related to other flaws, and thus, it is difficult to correct.
    %%%%%%%%%%%%%%%%%%%%%%%%%%%%%%%%%%%%%%%%%%%%%%%%%%%%%%%%%%%%%%%%%%%%%%%%%%
    \item In BO literature, usually, the number of iterations is represented by $T$, and \citet{Wang2017-Max} use this notation except for the theorem.
          On the other hand, in the theorem, the number of iteration is represented by $T^\prime$ instead of $T$, and the variable $T$ in the bound \eqref{eq:MES_regret_bound} is defined by the number of `partitioning' of the entire iterations in their proof.
          Each partition is defined so that it satisfies a probabilistic condition related to max-values, which we omit details here.
          An important issue is that, for this replaced $T$, dependency on the actual number of iterations is not clarified.
          Therefore, \eqref{eq:MES_regret_bound} does not reveal the convergence rate with respect to the number of iterations.
    %%%%%%%%%%%%%%%%%%%%%%%%%%%%%%%%%%%%%%%%%%%%%%%%%%%%%%%%%%%%%%%%%%%%%%%%%%%
    \item
          Since $\nu_{t^*}$ depends on $y^{t}_*$ in each iteration, $\nu_{t^*}$ is a random variable.
          This variable remains in the final upper bound, but the convergence rate of $\nu_{t^*}$ has not been shown.
    %%%%%%%%%%%%%%%%%%%%%%%%%%%%%%%%%%%%%%%%%%%%%%%%%%%%%%%%%%%%%%%%%%%%%%%%%%%
 \item
       In the theorem, \citet{Wang2017-Max} assumed that $y^{t}_*$ is drawn from the prior $F$.
       This assumption clearly does not match the basic idea of MES that selects the next query based on the mutual information estimated through samples of the max-value from the `posterior' distribution.
    %%%%%%%%%%%%%%%%%%%%%%%%%%%%%%%%%%%%%%%%%%%%%%%%%%%%%%%%%%%%%%%%%%%%%%%%%%%
    \item
          \citet{Wang2017-Max} claimed that the above simple regret bound can adapt to the setting that $y^{t}_*$ is drawn from the posterior.
          However, this claim is not proven and not obvious.
          In the proof, \citet{Wang2017-Max} implicitly assumed $\Pr(y^{t}_* < f_*, y^{t+1}_* < f_*) = \Pr(y^{t}_* < f_*) \Pr(y^{t+1}_* < f_*)$.
          Through the basic formula of the joint probability of independent variables, this decomposition is allowed if $f_*$ is deterministic, and $y^{t}_*$ and $y^{t+1}_*$ are independent of each other.
          However, both of these two conditions are not satisfied in the case of sampling from the posterior.
          For the first condition, as we already mention, $f_*$ should be a random variable (actually, this first condition is not satisfied even for the case of sampling from the prior).
          For the second condition, when $y^{t}_*$ and $y^{t+1}_*$ are sampled from posterior, $y^{t}_*$ depends on $\cD_{t-1}$ and $y^{t+1}_*$ depends on $\cD_t$.
          However, since $\cD_{t-1} \subset \cD_{t}$, dependency obviously exists between $y^{t}_*$ and $y^{t+1}_*$ (note that now $\cD_{t-1}$ and $\cD_{t}$ are also random variables not only because of the random noise of observations, but also the randomness in the acquisition function).
          In the proof, the decomposition is not justified when the two conditions are not satisfied.
          This decomposition plays a key role in the proof, and we consider that the above issue is not easy to avoid.
\end{enumerate}
We first tried to extend this theorem to our constrained problem, but we consider that correcting the above issues is at least not trivial.
Therefore, theoretical analysis of the MES-based methods (even for the simple unconstrained case) is still an open problem, to the best of our knowledge.
%%%%%%%%%%%%%%%%%%%%%%%%%%%%%%%%%%%%%%%%%%%%%%%%%%%%%%%%%%%%%%%%%%%%%%%%%%%%%%%%%%%%%%%%%%%%%%
\section{Details of Experimental Settings}
\label{sec:detail_exp_setting}
%%%%%%%%%%%%%%%%%%%%%%%%%%%%%%%%%%%%%%%%%%%%%%%%%%%%%%%%%%%%%%%%%%%%%%%%%%%%%%%%%%%%%%%%%%%%%%
In this section, we describe additional details of experimental settings.
%%%%%%%%%%%%%%%%%%%%%%%%%%%%%%%%%%%%%%%%%%%%%%%%%%%%%%%%%%%%%%%%%%%%%%%%%%%%%%%%
% \subsection{Other experimental settings for benchmark functions and materials data}
\subsection{Other Experimental Settings}
% {We used the method of moving asymptotes \citep{Krister2002-Class} in NLopt \citep{NLopt2008} and GPy \citep{GPy2014}, both of which are open-source libraries.}
%
Observations for objective and constraint functions are standardized so that they have zero mean and unit variance, respectively, before the regression at every iteration.
We used the open-source library called GPy \citep{GPy2014} for the GP regression.
The candidate intervals of hyperparameters $\sigma^2_{\rm LIN}$ and $\sigma^2_{\rm RBF}$ are set to $[0, 1]$, and the variance of the noise $\sigma^2$ is set to $10^{-6}$.
As we mentioned in the main paper, other hyperparameters are chosen by marginal likelihood maximization.
We employed an Automatic Relevance Determination (ARD) for the RBF kernel.
% except for the experiment of LLTO.
%
% \blue{Since the input dimension of LLTO data is extremely large ($d = 2185$), we cannot employ ARD.}
%
Thus, the RBF kernel has a hyperparameter $\*\ell \coloneqq (\ell_1, \dots, \ell_d)^\top \in \RR^d$, and it is written as
\begin{align*}
 % k_{\rm RBF}(\*x, \*x^\prime) \coloneqq \exp\biggl(- \frac{1}{2} \sum_{i=1}^d \frac{\| x_i - x_i^\prime\|^2}{\ell_i^2}\biggl),
 k_{\rm RBF}(\*x, \*x^\prime) \coloneqq \exp\biggl(- \frac{1}{2} \sum_{i=1}^d \frac{ (x_i - x_i^\prime)^2}{\ell_i^2}\biggl),
\end{align*}
where $x_i$ and $x^\prime_i$ are the $i$-th elements of $\*x$ and $\*x^\prime$, respectively.
%
% \blue{Hence, in the case of LLTO, $\ell_1 = \ell_2 = \dots = \ell_d$.}
%
For each one of $\ell_i$ for $i=1, \dots, d$, we set the same interval of the candidate value.
For the benchmark functions and the reactor network design problem, we set
$\ell_i \in [10^{-1} s_i, 10 s_i]$,
where $s_i$ is the length of the interval for the possible value of $x_i$ in the input domain, and the thresholds $z_c = 0$ for $c=1, \dots, C$.
For the CNN hyperparameter optimization problem, we set $\ell_i \in [10^{-1} , 2]$ and the thresholds $z_c = 0.5$ for $c=1, \dots, C$.
%
% \blue{
% For materials data, we set the interval as $\ell_i \in [10^{-2} b, 10^2 b]$, where $b$ is the median of the distance between all the pairs created from the pooled $\*x_i$.
% }
%
We used the method of moving asymptotes \citep{Krister2002-Class} in NLopt \citep{NLopt2008} for the constrained optimization of the sample path in CMES-IBO, CMES, TSC, and their parallel extensions.
The numbers of initial points are set as $3$ for the GP-derived synthetic function, $5$ for the two-dimensional benchmark functions, $5d$ for the Hartmann6 function, and $25$ for others.
Finally, the hyperparameter $M$ of P-EIC is selected as the minimum of the current observations.

\subsection{Settings for PESC}

Due to the difficulty of rewriting the Spearmint specification, the default settings such as for kernels and inner optimization were used.
Although PESC has the same number of initial points, their locations are not identical to other methods because of the same reason.

%%%%%%%%%%%%%%%%%%%%%%%%%%%%%%%%%%%%%%%%%%%%%%%%%%%%%%%%%%%%%%%%%%%%%%%%%%%%%%%%
\subsection{Details of Benchmark Functions}
Here, we provide detailed information on benchmark functions.
Note that we change the sign from the original functions if it is required to formalize as the maximization problem.
\begin{description}
    \item[Gardner1] Gardner1 is a simple test problem in which functions are constructed by sine and cosine functions \citep{Gardner2014-Bayesian}.
    The input dimension $d=2$, the input domain $\cX = [0, 6]^2$, and the number of constraints $C=1$.
    The detailed forms of functions are
    \begin{align*}
        f(\*x) &= - \cos(2 x_1) \cos(x_2) - \sin(x_1), \\
        g_1(\*x) &= - \cos(x_1) \cos(x_2) + \sin(x_1) \sin(x_2) + 0.5.
    \end{align*}
    \item[Gardner2] Gardner2 also has a simple form, but its feasible region is very small \citep{Gardner2014-Bayesian}.
    The input dimension $d=2$, the input domain $\cX = [0, 6]^2$, and the number of constraints $C=1$.
    The detailed forms of functions are
    \begin{align*}
        f(\*x) &= - \sin(x_1) - x_2, \\
        g_1(\*x) &= - \sin(x_1) \sin(x_2) - 0.95.
    \end{align*}
    \item[Gramacy] The Gramacy function is from \citep{Gramacy2016-Modeling}.
    The input dimension $d=2$, the input domain $\cX = [0, 1]^2$, and the number of constraints $C=2$.
    The detailed forms of functions are
    \begin{align*}
        f(\*x) &= - x_1 - x_2, \\
        g_1(\*x) &= \frac{1}{2} \sin(2 \pi \bigl(x_1^2 - 2 x_2)\bigl)  + x_1 + 2x_2  - 1.5, \\
        g_2(\*x) &= - x_1^2 - x_2^2 + 1.5.
    \end{align*}
    \item[Hartmann6] The Hartmann6 function is used in \citep{Letham2019-Constrained}.
    The input dimension $d=6$, the input domain $\cX = [0, 1]^6$, and the number of constraints $C=1$.
    The detailed forms of functions are
    \begin{align*}
        f(\*x) &= \sum_{i=1}^4 \alpha_i \exp \biggl( -\sum_{j=1}^6 A_{ij} (x_j - P_{ij})^2 \biggl), \\
        g_1(\*x) &= - \| \*x - \*x^\prime \| + 1,
     %\\
    \end{align*}
    where
    \begin{align*}
        \bm{\alpha} &= (1.0, 1.2, 3.0, 3.2)^\top, \\
        \bm{A} &= \left( \begin{array}{cccccc}
            10 & 3 & 17 & 3.5 & 1.7 & 8 \\
            0.05 & 10 & 17 & 0.1 & 8 & 14 \\
            3 & 3.5 & 1.7 & 10 & 17 & 8 \\
            17 & 8 & 0.05 & 10 & 0.1 & 14
        \end{array}\right), \\
        \bm{P} &= 10^{-4} \left( \begin{array}{cccccc}
            1312 & 1696 & 5569 & 124 & 8283 & 5886 \\
            2329 & 4135 & 8307 & 3736 & 1004 & 9991 \\
            2348 & 1451 & 3522 & 2883 & 3047 & 6650 \\
            4047 & 8828 & 8732 & 5743 & 1091 & 381
        \end{array}\right).
    \end{align*}
\end{description}

%%%%%%%%%%%%%%%%%%%%%%%%%%%%%%%%%%%%%%%%%%%%%%%%%%%%%%%%%%%%%%%%%%%%%%%%%%%%%%%%
\subsection{Details of CNN hyperparameter optimization problem}
We used PyTorch \citep{Paszke2019-PyTorch} to constract the CNN model.
The search domaion for this problem is set as follows:
the learning rate in $\{ 10^{-3}, 10^{-2}, 10^{-1}, 10^{0} \}$,
the dropout rate in $\{ 2^{-4}, 2^{-3}, 2^{-2}, 2^{-1} \}$,
the number of chennels in $\{ 2^{3}, 2^4, 2^5, 2^6 \}$ for two layers,
and the coefficient $\rho$ in $\{ 0, 0.1, \dots, 1.9\}$.
Thus, the size of input domain $|\cX| = 5120$.
To define $\*x$, we apply $\log_{10}$ and $\log_2$ transformation to the learning rate and the other inputs except for $\rho$, respectively.
%
% Moreover, we apply \red{the} logit transformation to accuracy and recall before the standardization for the regression.
Moreover, we apply the logit transformation to accuracy and recall for transforming the output domain from $[0,1]$ to $\RR$.
Under the setting $z_c = 0.5$ for $c = 1, \dots, 10$, the number of feasible points is $28$, which is only
about 0.5\% $(\approx 28/5120)$
of the entire candidates $\cX$.
% and thus the feasible region is tiny.
%
The unconstrained optimal point of the accuracy $\argmax_{\*x \in \cX} f(\*x)$ is different from the constrained optimal point $\*x_*$, meaning that the unconstrained optimization of the average accuracy sacrifices recalls of some classes.
% Note that the global optimal point of the accuracy $\argmax_{\*x \in \cX} f(\*x)$ is different from the constrained optimal point $\*x_*$.
%
% For a train of CNN, we used the default weight cross-entropy loss function in PyTorch, which is defined as follows:
For the training of CNN, we used the default weighted cross-entropy loss function in PyTorch, which is defined as follows:
\begin{align*}
    {\rm LOSS}(\*l, \hat{\*l}) &= \sum_{i=1}^N - \frac{w_{l_i}}{\sum_{m=1}^M w_m} \log \hat{l}_{i, l_i},
\end{align*}
where $\*l = (l_1, \dots, l_N) \in \RR^N$, $l_i$ is a correct label of $i$-th data, $\hat{l}_{i, m}$ is a predicted probability that $i$-th data has label $m$, and a $(i, m)$ element of $\hat{\*l} \in \RR^{N \times M}$ is $\hat{l}_{i, m}$.

%%%%%%%%%%%%%%%%%%%%%%%%%%%%%%%%%%%%%%%%%%%%%%%%%%%%%%%%%%%%%%%%%%%%%%%%%%%%%%%%%%%%%%%%%%%%%%
\section{Additional Experiments}
\label{sec:additional_experiments}
%%%%%%%%%%%%%%%%%%%%%%%%%%%%%%%%%%%%%%%%%%%%%%%%%%%%%%%%%%%%%%%%%%%%%%%%%%%%%%%%%%%%%%%%%%%%%%
% \subsection{Experiments on small \texorpdfstring{$C$}{C} problems for the parallel setting}
% \subsection{Experiments on \red{other benchmark functions}}
\subsection{Results with the Different Number of Samplings}

\begin{figure}
    \igr{.24}{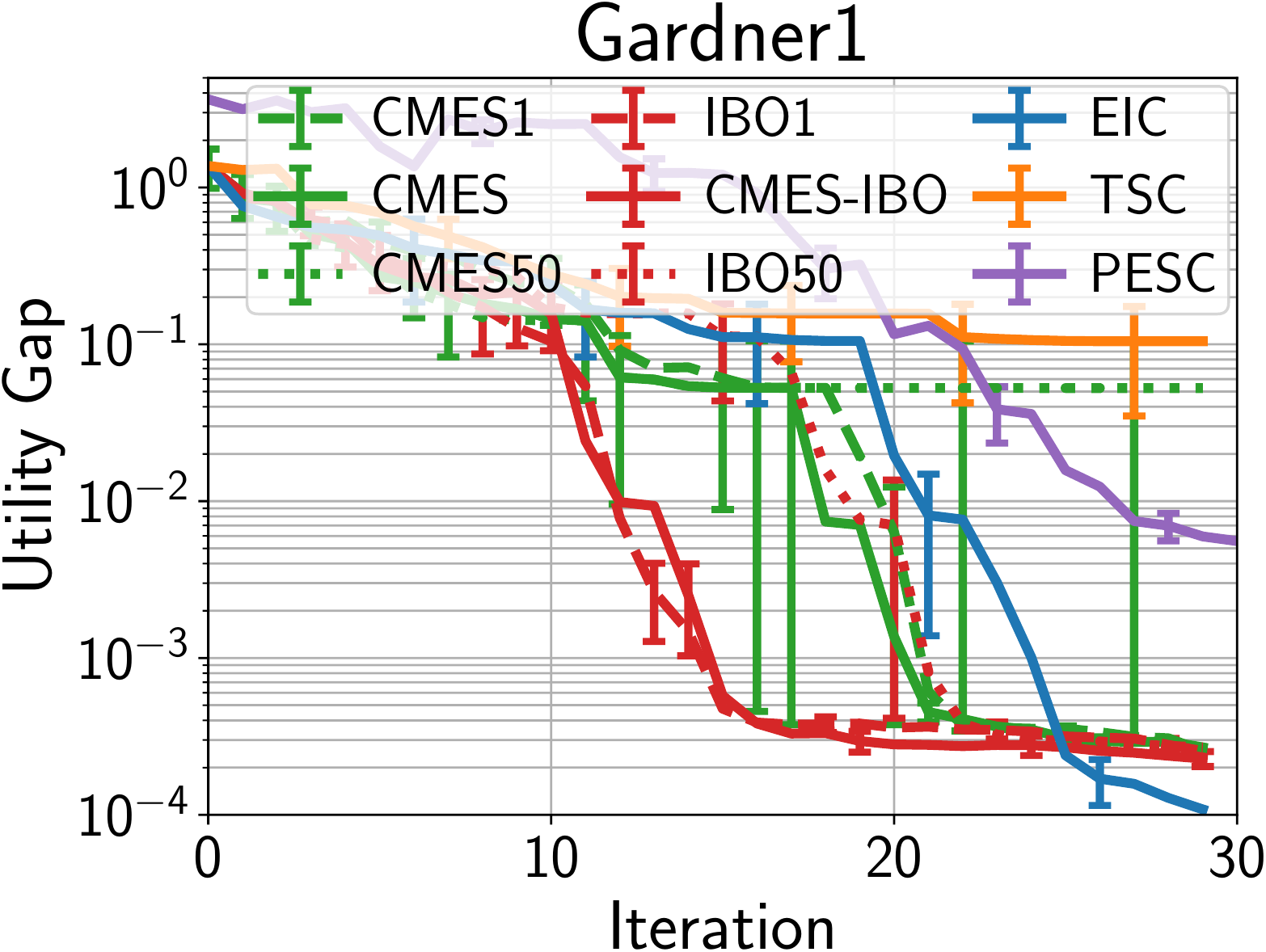}
    \igr{.24}{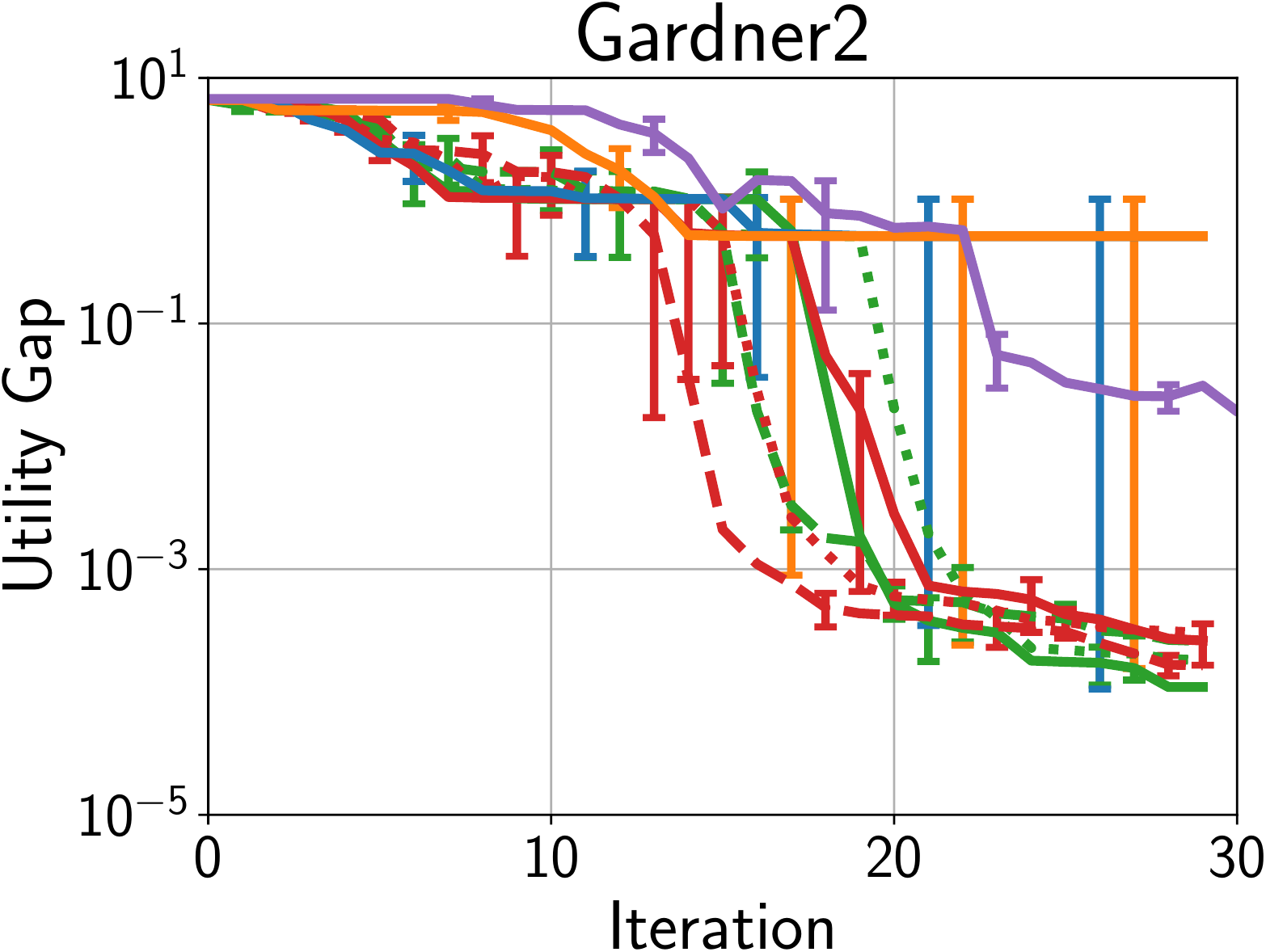}
    \igr{.24}{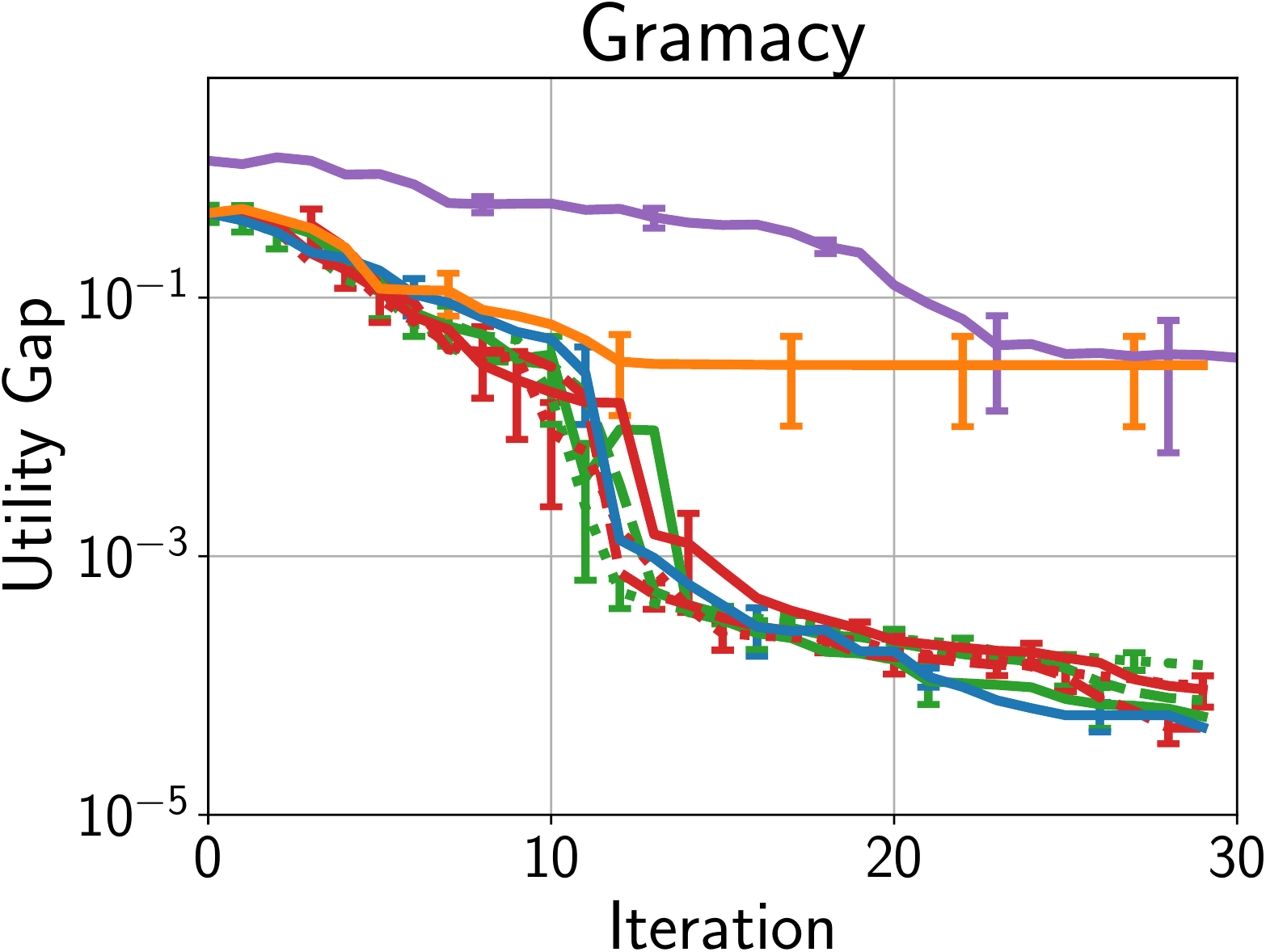}
    \igr{.24}{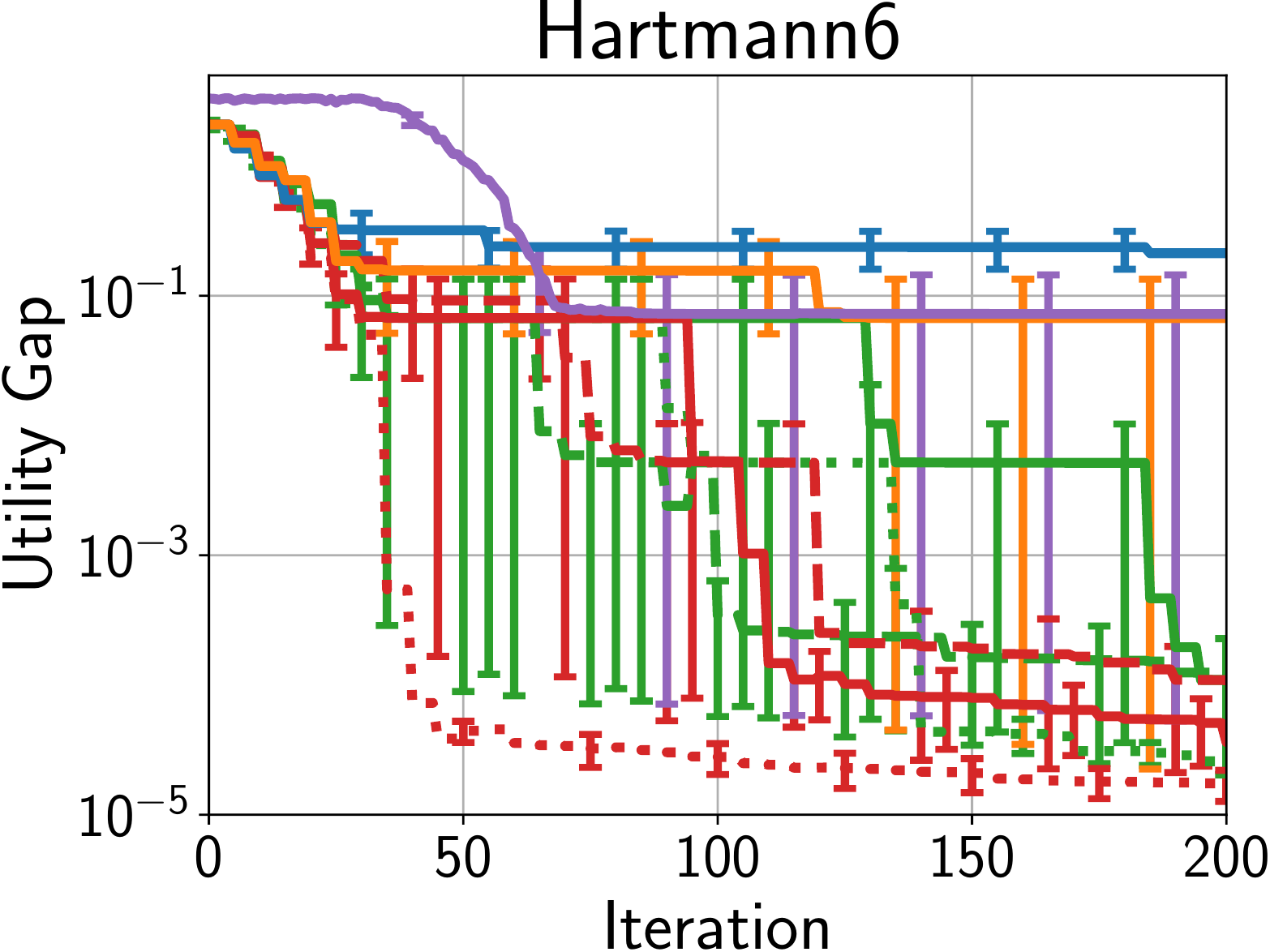}
    \caption{
        % Utility gap of small $C$ benchmark functions with different $K$ (average and standard error).
        Utility gap of benchmark functions with different $K$ (average and standard error).
    }
    \label{fig:benchmarks_with_diffK}
\end{figure}

By using benchmark functions from CBO literature, we here show the results with three different numbers of samplings $K = 1, 10$, and $50$ of CMES and CMES-IBO.
Figure~\ref{fig:benchmarks_with_diffK} shows that even with small $K = 1$, these two methods maintain reasonable performance compared with other methods.
In particular, for CMES-IBO, this result can be explained by the fact that $\alpha^{\rm IBO}(\*x)$ can be seen as the PI from $\tilde{f}_*$ even with $K = 1$ (Remark~\ref{rem:IBO-non-negativity}).

\subsection{Experimental Results of the Utility Gap Defined by Best Observed Value}
% \label{sec:experiment-}

\begin{figure*}[!t]
    \centering
    \igr{.16}{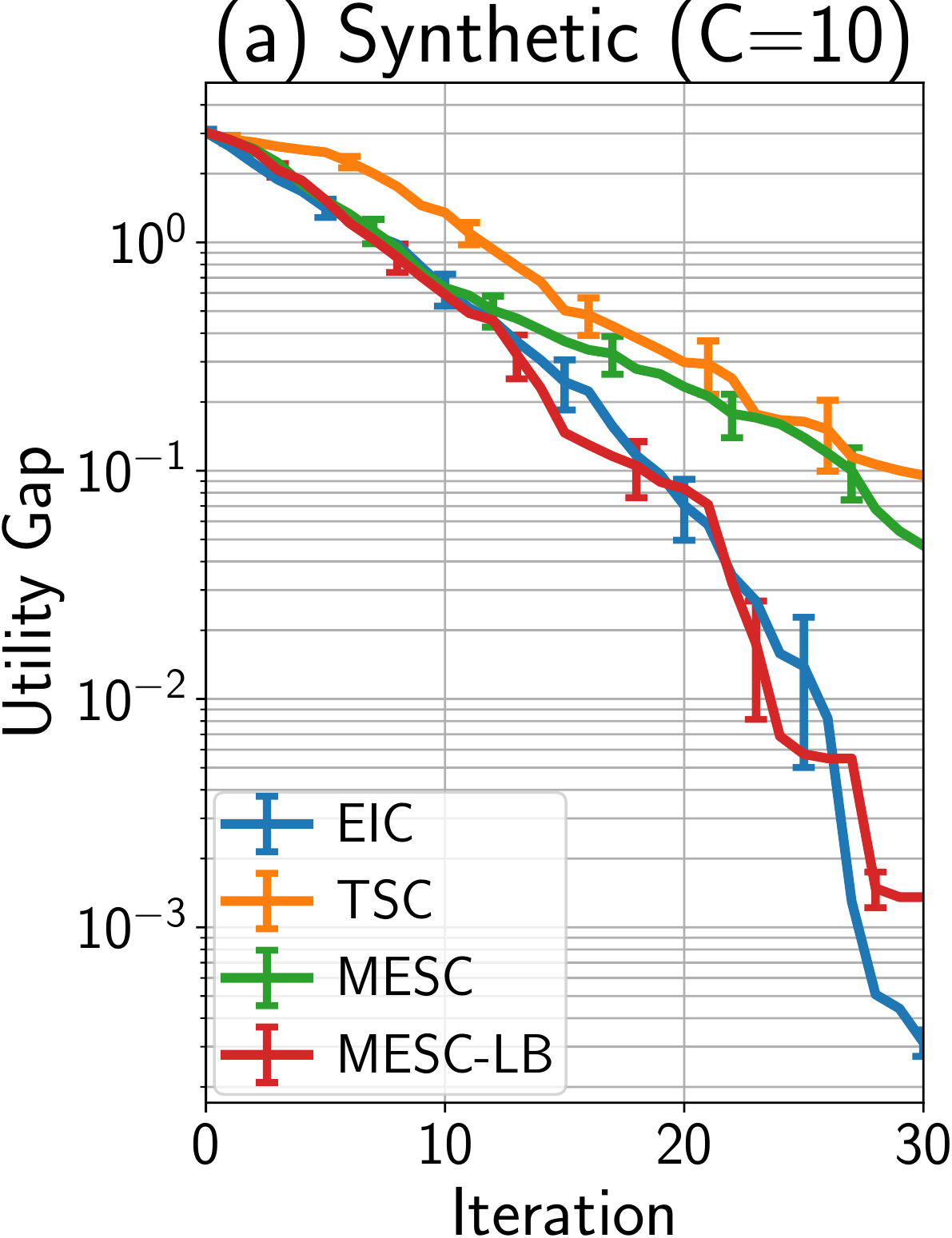}%
    \igr{.14}{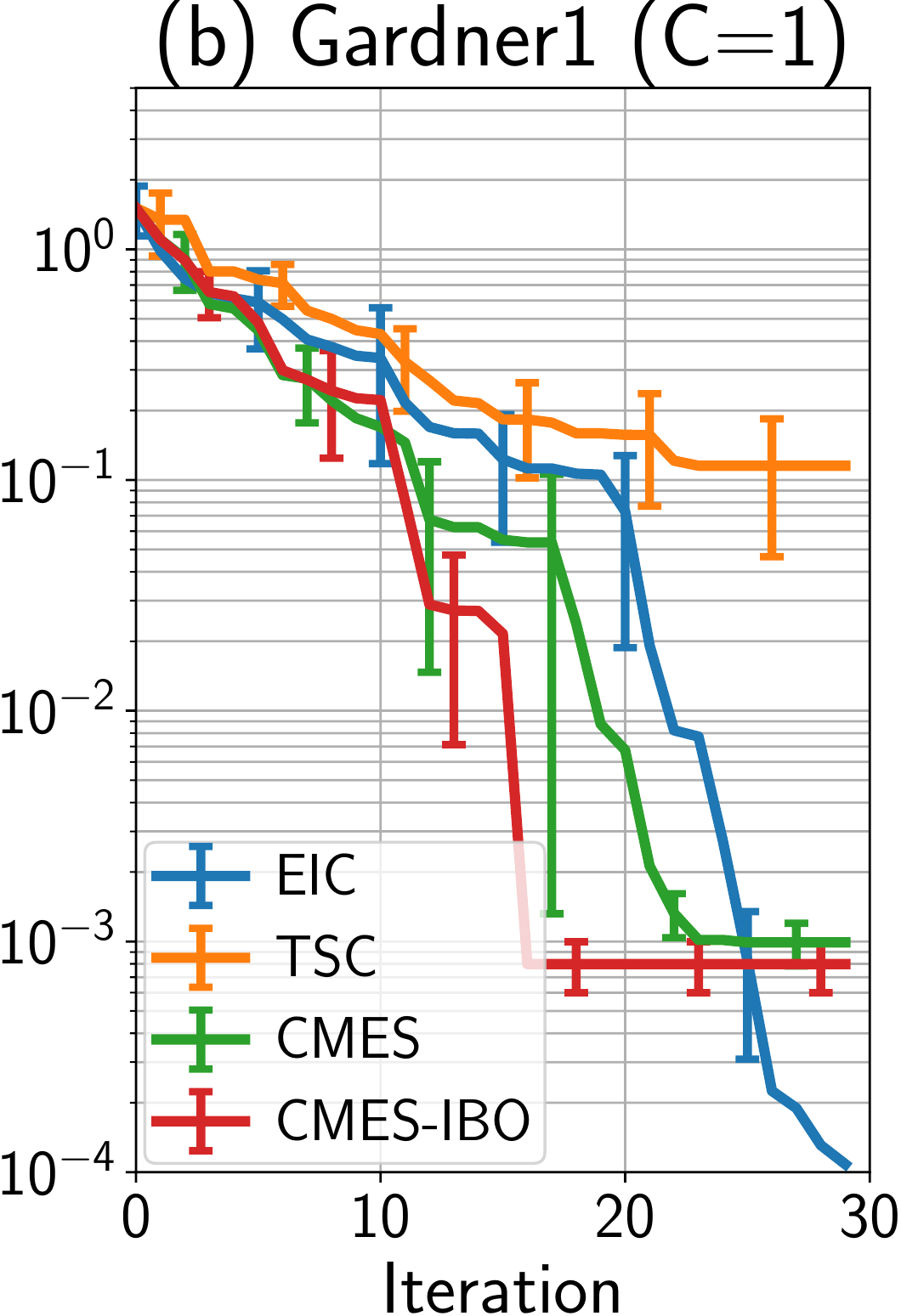}%
    \igr{.14}{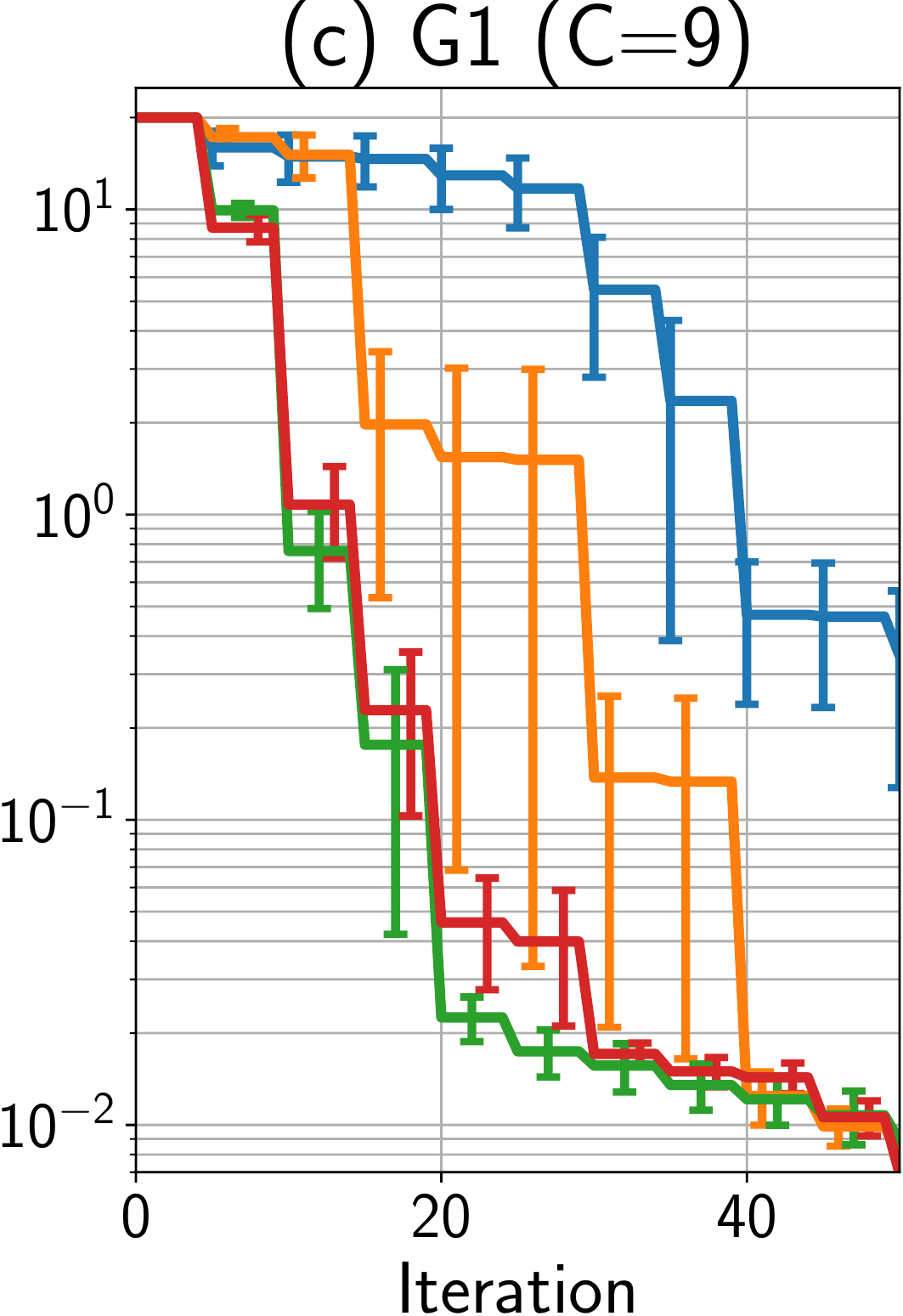}%
    \igr{.14}{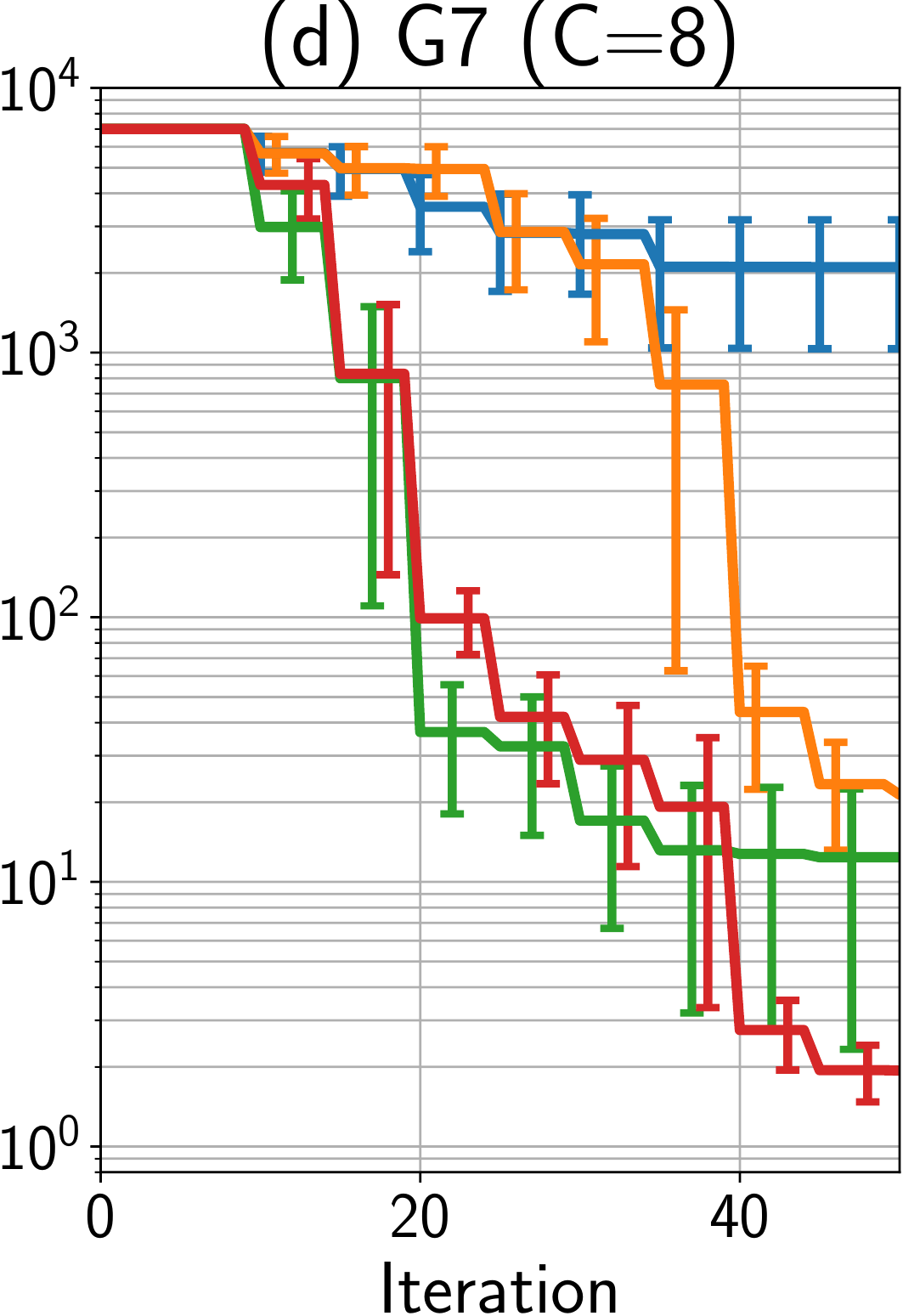}%
    \igr{.14}{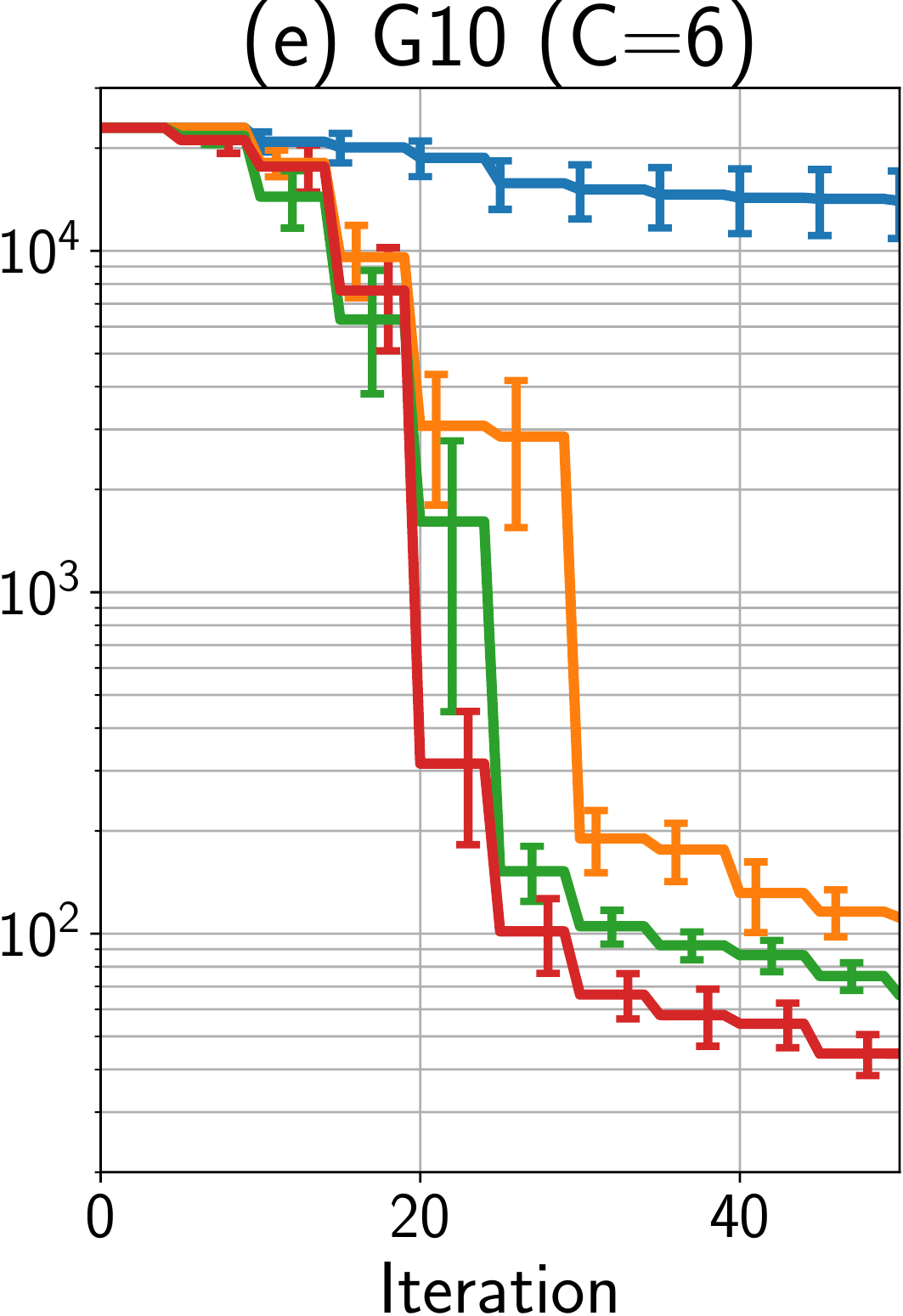}%
    \igr{.14}{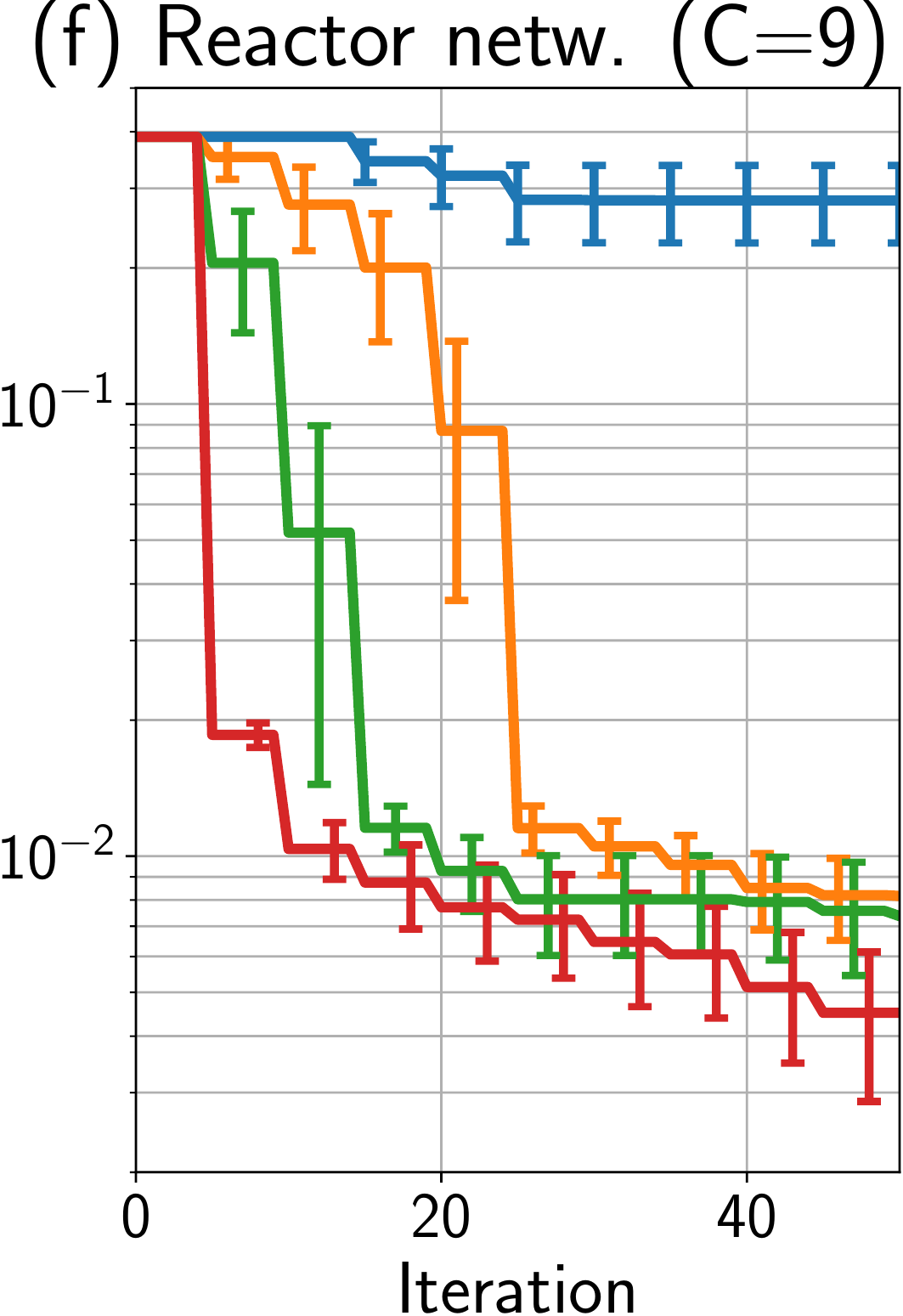}%
    \igr{.14}{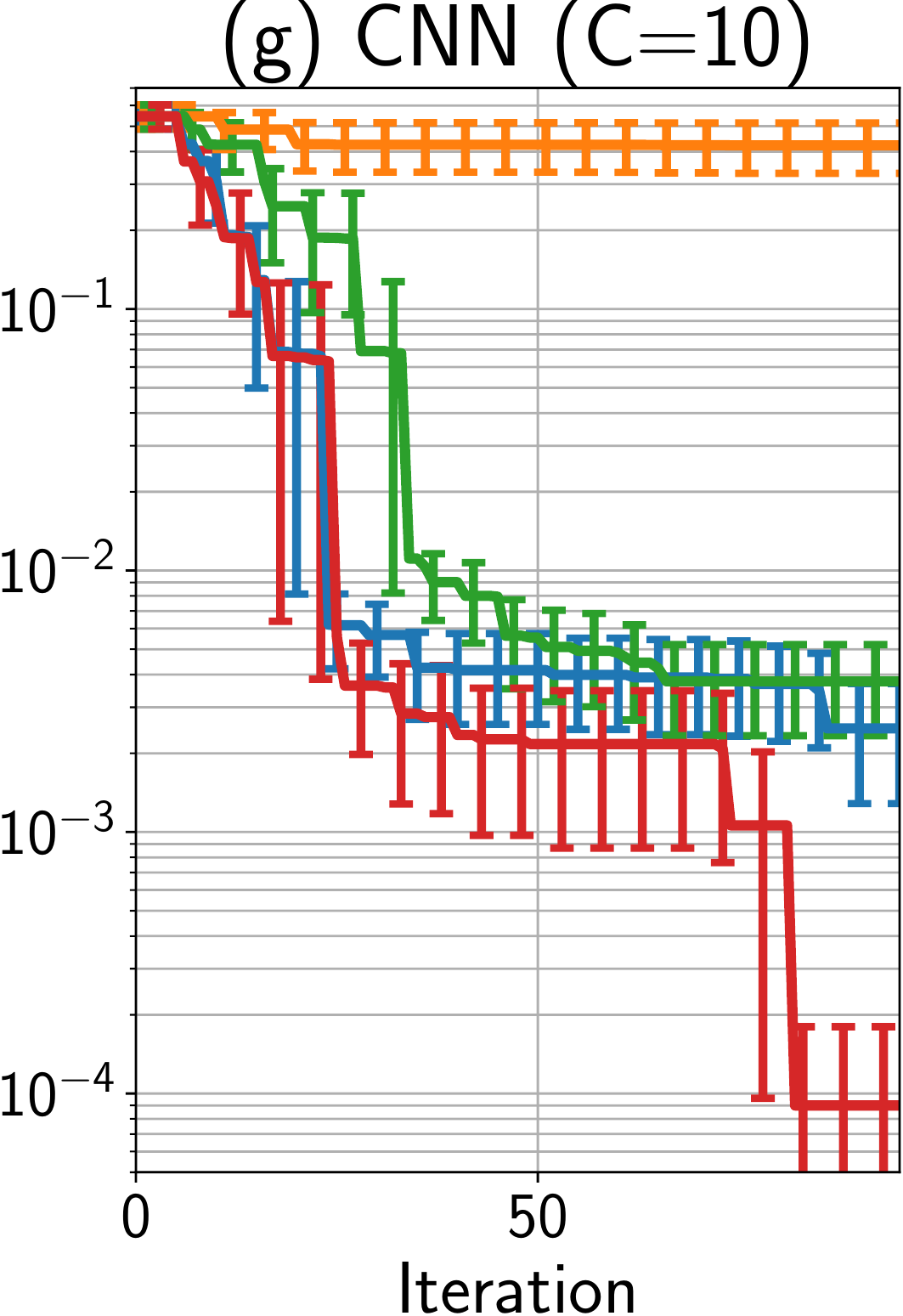}%

    \caption{
    Utility gap defined by the best observed value in sequential querying (average and standard error).
    %
 % \blue{The dashed lines in the synthetic function experiment represent the `single constraint' counterpart of each method that has the same color.}
    }
    \label{fig:sequential-best-observed}
\end{figure*}

Figure~\ref{fig:sequential-best-observed} shows the utility gap defined by the best observed value, which is $\min \{ f_* - f(\*x_i) \mid (\*x_i, y_i) \in \cD_{t-1}, g_c(\*x_i) \geq z_c, \forall c  \}$ when feasible solution is already obtained, otherwise $f_* - \min_{\*x \in \cX} f(\*x)$.
We see that all the plots show the similar behavior as Fig.~\ref{fig:sequential}.
This indicates that the recommendation $\hat{\*x}_t$ is mostly selected around the training data since $\hat{\*x}_t$ is chosen under constraints $\Pr(g_c(\*x) \geq z_c) \geq \sqrt[C]{0.95}$ for $\forall c$.
Note that PESC is not shown in Fig.~\ref{fig:sequential-best-observed} since Spearmint does not provide the best observed values.

%%%%%%%%%%%%%%%%%%%%%%%%%%%%%%%%%%%%%%%%%%%%%%%%%%%%%%%%%%%%%%%%%%%%%%%%%%%%%%%%%%
\subsection{Experiment for Correlated Synthetic Function}
\label{sec:experiment-correlated}

\begin{figure}[t]
    \centering
    \igr{.3}{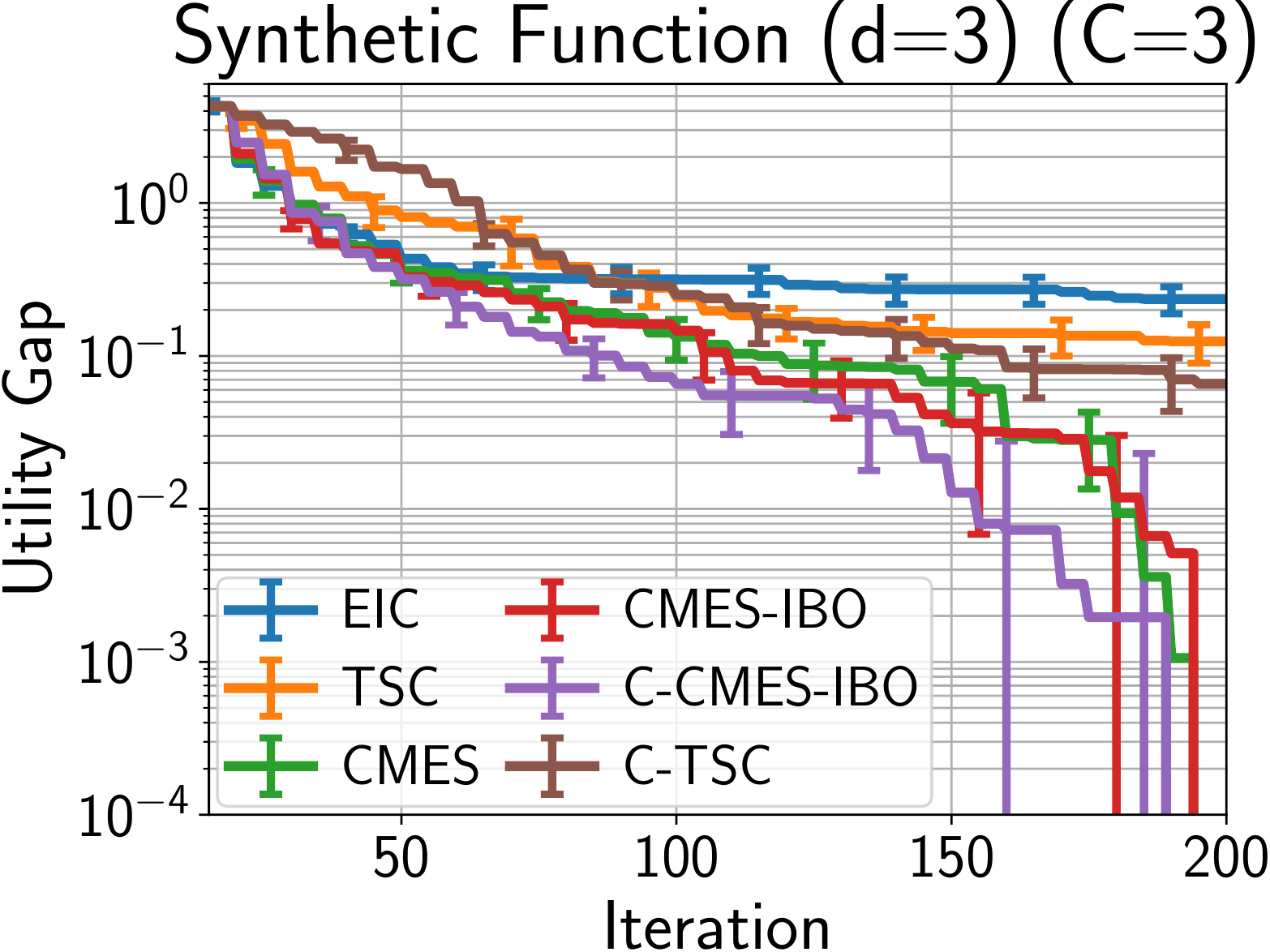}
    % \subfloat[Parallel Querying]{
    % \igr{.3}{Results_Inf_Syn_log_average_Q=3_d=5.pdf}
    % }
    \caption{Utility gap of correlated GP-derived synthetic functions (average and standard error).
    % The solid line and error bar express the average and the standard error, respectively.
    }
    \label{fig:synthetic-correlated}
\end{figure}

We evaluate the performance of CMES-IBO with correlated GPs on synthetic functions ($d = 3$ and $C = 3$).
Because of its high computational complexity, we did not employ CMES here (see Appendix~\ref{sec:computational_complexity}).
The objective and $C=3$ constraints were sampled from GPs with the RBF kernel, and the length scales of the kernels were $0.1$.
All the correlation coefficients among the objective and constraint functions were set as 0.5.
The same settings of GPs for the function generation were also used in each BO method.
The number of initial points is set as $5$.
The threshold for constraints were set as $z_c = 0$ for $\forall c$, and the input domain was $[0, 1]^3$.
We sampled $5$ sets of functions, and the experiment of each set ran $10$ times.
We report the mean and standard error of these $50$ trials.
Figure~\ref{fig:synthetic-correlated} shows the result.
C-CMES-IBO denotes the correlated extension of CMES-IBO.
We see that C-CMES-IBO improves CMES-IBO in this case and is superior to the other methods.
Although the correlated model can improve performance when the true function has a strong correlation, we mainly focus on the independent setting because the efficiency is often comparable to the correlated model, and the computations of independent CMES-IBO is much simpler than correlated case.

%%%%%%%%%%%%%%%%%%%%%%%%%%%%%%%%%%%%%%%%%%%%%%%%%%%%%%%%%%%%%%%%%%%%%%%%%%%%%%%%%%
\subsection{Experiment for Computational Time Evaluation}
\label{sec:experiment-time}

% In Fig.~\ref{tab:time-gramacy} and \ref{tab:time-g7}, we evaluate computational time of the acquisition function maximization.
% %
% Figure~\ref{tab:time-gramacy} is for the Gramacy function ($d = 2$ and $C = 1$).
In Table~\ref{tab:time-gramacy}, we evaluate computational time of the acquisition function maximization at $t = 30$.
EIC and TSC were relatively faster than the others.
CMES-IBO and CMES took longer times because of sampling of $f_*$ for which we need to solve constraint optimization problems $K$ times, while the acquisition function maximization itself was fast enough because of their simple closed forms of the acquisition functions.
Although it is not shown here, PESC has the same computational requirement for sampling $\*x_*$ because the same constraint optimization problems need to be solved.
It is worth noting that the $K$ times sampling of $f_*$ can be accelerated by computing in parallel because each one of the samplings is independent.

%
% Figure~\ref{tab:time-g7} is for the G7 function ($d = 10$ and $C  = 8$).
% %
% In this case, again, EIC and TSC have shorter total times.
%
% CMES-IBO and CMES took longer times because of sampling of $f_*$, but the acquisition function maximization was faster than EIC and TSC.

% \begin{table}
%     \caption{Computational time for the Gramacy function at $t=30$ (mean $\pm$ standard error)}
%     \begin{tabular}{c|c|c}
%          & sampling of $f_*$ & optimization of $\alpha(\*x)$ \\ \hline
%         EIC &  NA  &  $\pm$  \\ \hline
%         TSC &  NA  &  $\pm$  \\ \hline
%         CMES &  $\pm$  &  $\pm$  \\ \hline
%         CMES-IBO &  $\pm$  &  $\pm$
%     \end{tabular}
% \end{table}

\begin{table}
 \caption{Computational time (sec.) for the Gramacy function at $t=30$ (mean $\pm$ standard error)}
 \label{tab:time-gramacy}
    \centering
    \begin{tabular}{c|c|c}
         & sampling of $f_*$ & optimization of $\alpha(\*x)$ \\ \hline
        EIC &  NA  &  1.148 $\pm$ 0.287  \\ \hline
        TSC &  NA  & 2.670 $\pm$ 0.464 \\ \hline
        CMES & 20.728 $\pm$ 1.498 & 0.780 $\pm$ 1.317 \\ \hline
        CMES-IBO & 21.836 $\pm$ 0.933 & 1.163 $\pm$ 2.130
    \end{tabular}
\end{table}

% \begin{table}
%  \caption{\blue{Computational time for the G7 function at $t=50$ (mean $\pm$ standard error)}}
%  \label{tab:time-g7}
%  \centering
%     \begin{tabular}{c|c|c}
%          & sampling of $f_*$ & optimization of $\alpha(\*x)$ \\ \hline
%         EIC &  NA  & 32.463 $\pm$ 29.320  \\ \hline
%         TSC &  NA  & 33.939 $\pm$ 0.205  \\ \hline
%         CMES & 336.359 $\pm$ 2.697 & 0.155 $\pm$ 0.043 \\ \hline
%         CMES-IBO & 338.117 $\pm$ 3.778  & 0.231 $\pm$ 0.227
%     \end{tabular}
% \end{table}

\vfill
%%%%%%%%%%%%%%%%%%%%%%%%%%%%%%%%%%%%%%%%%%%%%%%%%%%%%%%%%%%%%%%%%%%%%%%%%%%%%%%%%%%%%%%%%%%%%%
\end{document}